\documentclass[review,10pt]{JMtemplate}

\usepackage{lastpage}
\usepackage[utf8]{inputenc}
\usepackage[T1]{fontenc}

\usepackage{amsmath}
\usepackage{amsfonts}
\usepackage{booktabs}
\usepackage{multirow}
\usepackage{graphicx}
\usepackage{nicefrac}
\usepackage{microtype}
\usepackage{xcolor}
\usepackage{placeins}
\usepackage{lipsum}
\usepackage{longtable}
\usepackage{booktabs}      
\usepackage{threeparttable}
\usepackage{multicol}
\usepackage{array}
\usepackage{subcaption}
\usepackage{changepage}
\usepackage{pdflscape}
\usepackage{makecell}
\usepackage{adjustbox}
\usepackage{changepage}
\usepackage{caption}
\usepackage[figuresright]{rotating}
\graphicspath{{media/}}
\usepackage{marvosym}

\newcommand*{\dif}{\mathop{}\!\mathrm{d}}

\usepackage[
    colorlinks=true,    
    linkcolor=blue,     
    urlcolor=blue,      
    citecolor=green,    
    filecolor=magenta   
]{hyperref}
\usepackage{textcomp}

\begin{document}
\begin{frontmatter}
\title{The Origins of Stochasticity: Comprehensive Investigations on Uncertainty Quantification for Large Language Models}

\author{\textbf{Xiang-Jun Ou}\textsuperscript{2,*} \quad
\textbf{Shuang Liang}\textsuperscript{\rm 1,2,*}  \quad
\textbf{Xin-Yu Hu}\textsuperscript{\rm 1,2}  \quad
\textbf{Rong-Hao Huang}\textsuperscript{\rm 2} \quad
\textbf{Jing Wang}\textsuperscript{\rm 2} \\[0.2em]
\textbf{Shao-Qun Zhang}\textsuperscript{\rm 1,2,\Letter} \\[0.3em]
\small \textsuperscript{1} National Key Laboratory for Novel Software Technology, Nanjing University, Nanjing 210063, China.\\
\small \textsuperscript{2} School of Intelligent Science and Technology, Nanjing University, Suzhou 215163, China.\\
\small \texttt{ zhangsq@lamda.nju.edu.cn }
}

\begin{abstract}
Recent advancements in Large Language Models (LLMs) have enabled sophisticated reasoning and content generation, yet their inherent stochasticity poses significant challenges for ensuring predictive credibility. While traditional uncertainty taxonomy paradigms, such as the dichotomy of aleatoric and epistemic uncertainties, provide conceptual foundations, they often fail to capture the multi-component and multi-stage nature of LLM generation and struggle to evaluate the effectiveness of various  Uncertainty Quantification (UQ) methods. In this paper, we propose a granular uncertainty taxonomy that systematically attributes LLM uncertainty into input-level, parameter-level, token-level, and decoding-process sources. Correspondingly, we categorize existing UQ methods into Bayesian, ensemble, consensus-based, and single-pass approaches. Furthermore, we introduce a comprehensive evaluation framework covering diverse generation settings and metrics. We empirically evaluate 21 typical UQ methods across three prominent LLM families, including Qwen3, Llama 3.2, and DeepSeek-V3, on benchmarks such as TriviaQA, GSM8K, and HumanEval. Our experimental results demonstrate that (i) the effectiveness of UQ methods is sensitive to task types and generation settings; (ii) consensus-based methods, typed Deg and EigV, consistently outperform other UQ approaches; and (iii) larger model scales correlate with lower uncertainty estimates, suggesting an empirical scaling law for LLM uncertainty. This work bridges the gap between theoretical origins and practical deployment, providing a versatile diagnostic tool for systematically quantifying uncertainty in LLM applications.

\end{abstract}
\end{frontmatter}

\section{Introduction}  \label{sec:intro}
Recent advances in Large Language Models (LLMs) have led to a hot wave in reasoning-augmented and open-ended content generation~\citep{wei2022CoT,achiam2023gpt}, including domain-specific knowledge retrieval and multi-step problem-solving. An omnipresent challenge in these tasks is to achieve a balance between creative generation and output credibility led by the stochasticity of LLMs~\citep{Lin2022Verbalized_Probability,farquhar2024SemanticEntropy}, particularly when navigating ambiguous prompts or nuanced semantic contexts. For instance, multi-step problem-solving tasks are susceptible to uncertain outputs, where identical prompts result in inconsistent responses; even in cases of outcome convergence, the model's reasoning process remains heterogeneous and prone to internal logical contradictions. If a model's capacity to withstand the influence of inside stochasticity is measured at all, it is usually evaluated heuristically or statistically against some uncertainty sources or tasks in which success is taken as an indicator of Uncertainty Quantification (UQ) in a colloquial sense. 

In the landscape of modern artificial intelligence research, addressing the triad of attributing uncertainty sources of a given LLM, quantifying the effects led by these uncertainty sources, and making uncertainty-based decisions is fundamental and paramount to deploying robust and trustworthy systems. These challenges represent a hierarchical pipeline. The first challenge lies in identifying the origins of uncertainty, which is usually achieved by decomposing the total predictive uncertainty into its constituent components. The typical attribution argues that uncertainty sources consist of data and model uncertainties~\citep{Hllermeier2021Aleatoric_and_epistemic_uncertainty}. Once the sources are identified, the objective shifts to estimating the effects led by these uncertainty sources using calibrated scores~\citep{zhang2026uncertainty}. The capacity of UQ techniques acts as an essential safeguard against overconfident yet erroneous predictions. Intuitively, once uncertainty is estimated, a primary decision-making mechanism is the rejection of outputs with high uncertainty scores.

\subsection{Related Studies}
Uncertainty often characterizes the variability in system outputs or decisions, frequently driven by inherent stochasticity~\citep{Kendall17what}. Given its nature as a descriptive construct rather than an intrinsic property, any comprehensive taxonomy of uncertainty must be structured according to its constitutive origins, that is, uncertainty sources.

\paragraph{Aleatoric and Epistemic Uncertainties.} Uncertainty can be broadly categorized into two classes, named the aleatoric and epistemic uncertainties~\citep{Hllermeier2021Aleatoric_and_epistemic_uncertainty}, which correspond to the data and model uncertainties, respectively. Aleatoric uncertainty characterizes the output variability of a model resulting from the inherent randomness, noise, or natural variability present in a process or dataset. In the case of LLMs, the aleatoric uncertainty arises when there is stochasticity in the ground-truth responses. For instance, in open-ended generative tasks such as dialogue modeling or creative storytelling, a single input prompt may be associated with a distribution of equally valid, yet divergent, ground-truth responses due to the inherent ambiguity of human language. Since this uncertainty is inherently rooted in the data distribution, it is typically irreducible, even through further training or model refinement~\citep{Hllermeier2021Aleatoric_and_epistemic_uncertainty}. Epistemic uncertainty characterizes the variability or doubt regarding a prediction stemming from the limited capacity of a model against the underlying knowledge. In the context of LLMs, epistemic uncertainty arises when the model encounters prompts that lie outside its training distribution or involve domains where the model has insufficient exposure. For example, when a model is queried about highly specialized, niche technical knowledge or recent events that occurred post-training, it may exhibit high epistemic uncertainty, leading to hallucinations or overconfident yet inaccurate predictions. Unlike its aleatoric counterpart, this form of uncertainty is reducible; it can be effectively mitigated by expanding the training dataset~\citep{Brown2020Language}, employing retrieval-augmented generation~\citep{Lewis2020RAG}, or fine-tuning the model on target-specific corpora~\citep{Lin2022Verbalized_Probability}.

It is obvious that the traditional uncertainty attribution paradigm of aleatoric and epistemic uncertainties is too crude, although it may provide an important conceptual foundation for uncertainty sources of LLMs. Specifically, this paradigm often conflates distinct operational mechanisms within the transformer architecture, obscuring whether predictive uncertainty arises from input-level noise or deficiencies in representation learning. Importantly, developers may fail to attribute and quantify the effects of each uncertainty source during the LLM generation process according to this uncertainty attribution paradigm. It is necessary to develop a granular and source-based taxonomy that allows for a more precise decomposition of these uncertainty effects, effectively mapping high-level uncertainty back to the specific components of the generation process.

\begin{figure*}[t]
    \centering
    \includegraphics[width=\textwidth]{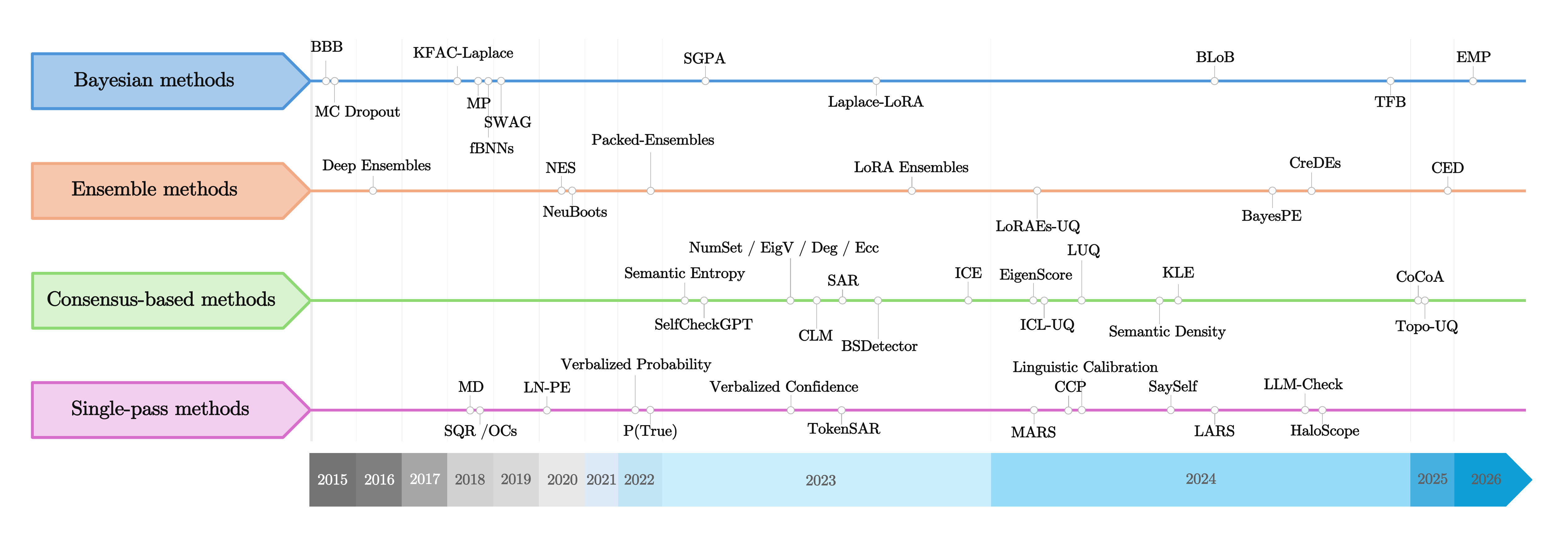}
    \caption{Timeline of remarkable uncertainty quantification methods, including Bayesian, ensemble, consensus-based, and single-pass ones, grouped by the uncertainty taxonomy.}
    \label{fig:uq_timeline}
\end{figure*}

\paragraph{Uncertainty Quantification.} Past decades have witnessed significant efforts on quantifying the uncertainty of Deep Neural Networks (DNNs)~\citep{Blundell2015BayesbyBackprop,Gal2016MCDropout}, such as Bayesian and ensemble UQ ones~\citep{lakshminarayanan2017DeepEnsemble,jospin2022BNNs}. Though undoubtedly helpful, such statistics and heuristics may suffer from the curse of massive amounts of data and parameters in LLMs. Besides, modern LLMs are used more extensively and involve more complex scenarios, including but not limited to traditional prediction and new-derived reasoning-augmented generation~\citep{wei2022CoT}. There have been various approaches to quantify the uncertainty of LLMs. An intuitive way is to prompt the LLM to judge its own reasoning uncertainty~\citep{Kadavath2022Ptrue/PIK,Lin2022Verbalized_Probability}. However, this approach evidently suffers from the integration of the epistemic uncertainty of LLM generation and the aleatoric uncertainty arising from prompting stochasticity, thereby amplifying uncertainty in LLM reasoning or decision-making. Deep ensembles are widely used in UQ methods for LLMs~\citep{lakshminarayanan2017DeepEnsemble}. For example, studies based on low-rank adaptation~\citep{wang2023LoRAEnsemble} use the outputs of a set of independent models or of the same model to quantify the model's uncertainty about a given cue. Due to the massive amounts of data and parameters that amplify the cost of conducting statistical trials on LLMs, most subsequent UQ studies for LLMs have eschewed traditional statistical UQ methods and instead introduced approximate quantification methods with lower computational costs. For example, LLM-check~\citep{sriramanan2024llm-check} directly uses the unique architecture of a specific LLM to estimate its uncertainty. Figure~\ref{fig:uq_timeline} illustrates the timeline of representative UQ methods for DNNs and LLMs.

There have been a few efforts to survey UQ methods for LLMs. \citet{Liu2025survey} reviewed the existing methods relative to UQ and confidence calibration in LLMs according to computational efficiency and uncertainty dimensions. \citet{Shorinwa2026survey} conducted a considerably comprehensive survey of UQ for LLMs, including a taxonomy of UQ methods, relevant benchmarks, practical applications, and open challenges for future research. Despite the analytical progress, many aspects of attributing and quantifying the uncertainty of LLMs are far from clear. The first one is that the typical paradigm that attributes the uncertainty source of LLMs to the aleatoric and epistemic uncertainties usually fails to deconstruct the granular stages of LLM generation and reasoning, thereby obscuring how uncertainty propagates across distinct architectural layers and logical steps. The second one is that there are no comprehensive experiments or empirical investigations to compare the actual deployment effects of various UQ methods for LLMs. Existing studies predominantly focus on the feasibility design of UQ to LLMs, leaving a critical gap in understanding and comparing how different UQ techniques perform in practical deployment across diverse architectures, tasks, and real-world scenarios. Developers are forced to rely on trial-and-error or heuristic methods to make choices, rather than decisions based on sufficient information and evidence.

\subsection{Our Contributions}
In this paper, we provide comprehensive investigations on attributing and quantifying the uncertainty of LLMs. The main contribution is the development of the uncertainty taxonomy, as shown in Table~\ref{tab:uq_methods}, based on which the uncertainty sources of LLMs are grouped into the input, model parameter, token-level distribution, and decoding process, while the quantification methods are categorized as the Bayesian, ensemble, consensus-based, and single-pass ones. Correspondingly, we also specify the model accessibility, training necessity, consumption, and evaluation. The proposed uncertainty taxonomy establishes a unified and structured framework that transcends specific model architectures or downstream tasks, making it highly applicable across the diverse landscape of modern LLM deployments. 

Another contribution of this work is the comprehensive evaluation framework of LLM UQ methods, which consists of a metric aspect covering 7 evaluation metrics and a generation-setting aspect that includes answer-only generation and reasoning-augmented generation. Based on this evaluation framework, we empirically evaluate the 21 typical UQ methods of five LLMs, including Qwen3 family, Llama 3.2 family, and DeepSeek-V3, on three datasets that contain TriviaQA, GSM8K, and HumanEval, which are evaluated by 7 metrics from the perspectives of ranking, calibration, and coverage. The experimental results show that (1) the effectiveness of LLM UQ methods is explicitly influenced by task types and generation settings; (2) Deg and EigV that belong to the consensus-based UQ methods exhibit the best overall performance among all evaluated methods across LLMs, datasets, tasks, and evaluation objectives; and (3) LLMs with larger scales lead to lower uncertainty estimates among all evaluated methods and white-box LLMs, revealing an empirical scaling law of LLM uncertainty. Overall, by bridging the gap between theoretical origins and practical quantification of LLM stochasticity and capacity, this work provides a versatile tool for researchers and practitioners to systematically diagnose and quantify the uncertainty in various applications of LLMs.

The rest of this paper is organized as follows. Section~\ref{sec:LLM_UQ} categorizes existing uncertainty sources and quantification methods for LLMs. Section~\ref{sec:experiments} conducts comparative experiments to evaluate the remarkable uncertainty quantification methods of LLMs. Section~\ref{sec:discussions} concludes this work with in-depth discussions and prospects.

\section{Uncertainty Quantification for LLMs}  \label{sec:LLM_UQ}
In this section, we categorize existing UQ methods for LLMs under a six-dimensional taxonomy according to the origin pipeline from the uncertainty source and quantification to the practical conditions under which a UQ method can be applied and evaluated. 
We analyze the uncertainty sources in Subsection~\ref{subsec:uncertainty_sources} and introduce the types of UQ methods in Subsection~\ref{subsec:UQ_types}. 
We further compare these UQ methods in terms of model accessibility in Subsection~\ref{subsec:access}, training necessity in Subsection~\ref{subsec:training_necessity}, consumption in Subsection~\ref{subsec:consumption}, and evaluation in Subsection~\ref{subsec:evaluation}. 
Table~\ref{tab:uq_methods} summarizes the UQ taxonomy for LLMs.

\subsection{Uncertainty Sources in LLM Generation}  \label{subsec:uncertainty_sources}
\begin{figure}[!htb]
    \centering
    \includegraphics[width=\textwidth]{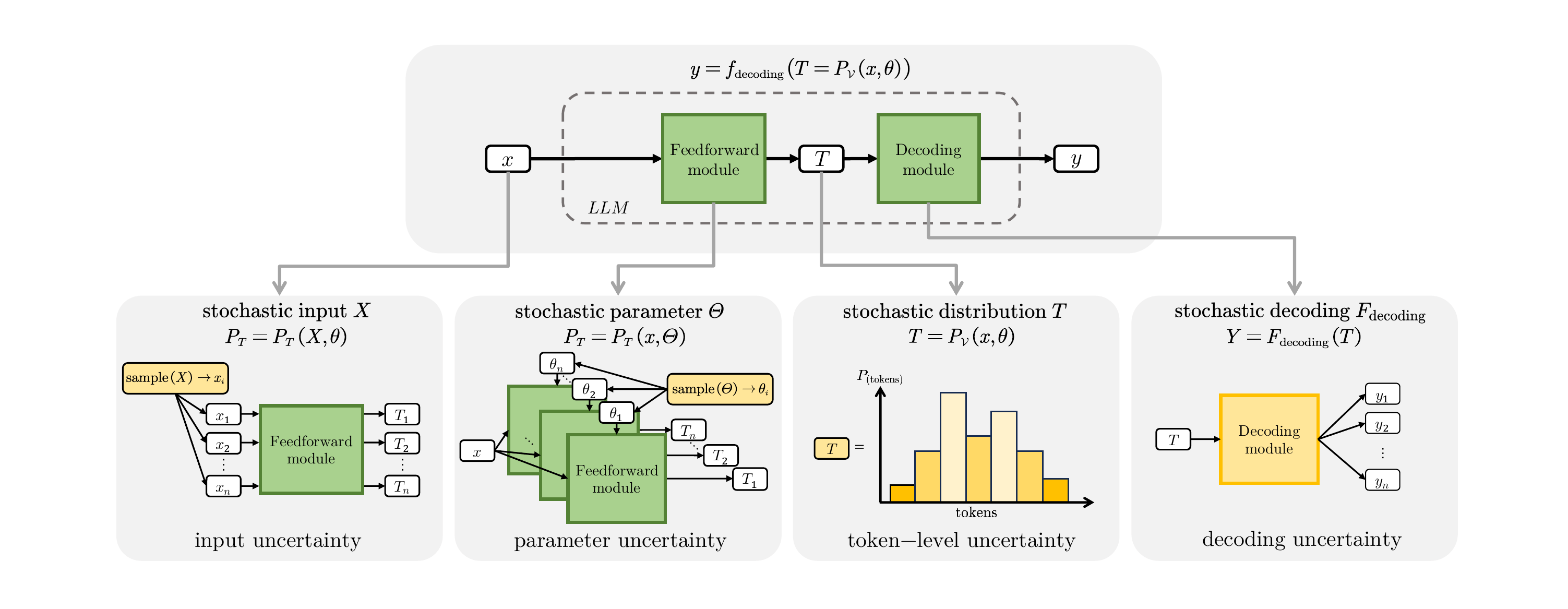}
\caption{Illustration of four uncertainty sources that originate from the stochasticity of inputs, model parameters, token-level distributions, and decoding process in LLMs.}
    \label{fig:uncertainty_sources}
\end{figure}

This subsection provides an in-depth analysis of the uncertainty sources in LLM generation. We start this analysis with a formal introduction to the LLM generation process. 
Let $X$ and $\Theta$ denote the random variables corresponding to the input and the parameters, respectively. 
Given a realization $(x,\theta)$, the LLM induces a sequence of token-level distributions over the vocabulary $\mathcal{V}$ during autoregressive generation. 
Specifically, at each generation step, the model predicts the next-token distribution conditioned on the input and the previously generated context. 
For simplicity, we formulate this sequence of token-level distributions as
\begin{equation}  \label{eq:T}
T = P_{\mathcal{V}}(x,\theta) \ .
\end{equation}
Let $f_{\text{decoding}}$ denote the overall generation mapping from token-level distributions to the final output, which encompasses both the decoding procedure and contextual accumulation in autoregressive generation. Hence, the final output of LLM generation can be written as
\begin{equation}  \label{eq:y}
y=f_{\text{decoding}}\left(T=P_{\mathcal{V}}(x,\theta)\right) \ .
\end{equation}
Based on Eqs.~(\ref{eq:T}-\ref{eq:y}), we can position the sources of uncertainty, i.e., the output variability led by the input uncertainty, parameter uncertainty, token-level uncertainty, and decoding uncertainty, which correspond to the stochasticity injected into the input $x$, parameters $\theta$, token-level distributions $T$, and decoding process $f_{\text{decoding}}(\cdot)$, respectively. Figure~\ref{fig:uncertainty_sources} illustrates the four categories of uncertainty sources in LLM generation. 

Notice that there is a significant causal relationship between input and parameter uncertainties and token-level and decoding uncertainties, where input and parameter uncertainties collectively influence both token-level and decoding uncertainties, with the former also having a direct impact on the latter. Besides, both input and parameter uncertainties originate from their inherent stochasticity; in contrast, both token-level and decoding uncertainties are not only influenced by their inherent stochasticity and the upstream uncertainties, but also modulated by model capacity and decoding strategies, respectively.

\paragraph{Input Uncertainty.} Input uncertainty arises from stochasticity, such as inherent ambiguity, incompleteness, or perturbation of the input $x$. Formally, we formalize the input with stochasticity as a random variable $X$, which is propagated through a fixed model parameterized by $\theta$, thereby inducing a stochastic token-level distribution $P_\mathcal{V}(X,\theta)$ over the vocabulary $\mathcal{V}$. LLMs often exhibit sensitivity to subtle variations in prompt phrasing~\citep{Min2020AmbigQA,Sclar2024PromptSensitivity}; for instance, paraphrased prompts or noisy contexts can perturb the model’s token-level distribution, resulting in divergent generation outcomes despite the application of identical model weights and decoding strategies. Consequently, the input uncertainty quantifies the extent to which ambiguity or perturbation within the input space propagates to the token-level distribution in LLM-based generation.

\clearpage
\begin{table}
\caption{The comprehensive survey of uncertainty quantification methods for LLMs.}
\label{tab:uq_methods}
\centering
\resizebox{\textwidth}{!}{
\begin{tabular}{llccccc}
\toprule
\textbf{Methods} & \textbf{Type} & \textbf{Uncertainty Sources} & \textbf{Model Accessibility} & \textbf{Training Necessity} & \textbf{Consumption} & \textbf{Evaluation} \\
\midrule
BBB~\citep{Blundell2015BayesbyBackprop} & Bayesian & Parameter & White-box & Full training & High & Calibration \\
MC Dropout~\citep{Gal2016MCDropout} & Bayesian & Parameter & White-box & Full training & High & Calibration \\
fBNNs~\citep{Sun2019fBNNs} & Bayesian & Parameter & White-box & Full training & High & Calibration \\
SWAG~\citep{Maddox2019SWAG} & Bayesian & Parameter & White-box & Full training & High & Calibration/Ranking \\
SGPA~\citep{Chen2023SGPA} & Bayesian & Parameter & White-box & Full training & High & Calibration/Ranking \\
BLoB~\citep{Wang2024BLoB} & Bayesian & Parameter & White-box & Fine-tuning & High & Calibration \\
Laplace-LoRA~\citep{yang2023laplace-lora} & Bayesian & Parameter & White-box & Fine-tuning & High & Calibration \\
KFAC-Laplace~\citep{Ritter2018KFAC-Laplace} & Bayesian & Parameter & White-box & Post-hoc & Medium & Calibration \\
TFB~\citep{Shi2024TFB} & Bayesian & Parameter & White-box & Post-hoc & Medium & Calibration \\
MP~\citep{Wu2019MP} & Bayesian & Parameter & White-box & Post-hoc & Medium & Calibration/Ranking \\
EMP~\citep{Liang2026EMP} & Bayesian & Parameter & White-box & Post-hoc & Medium & Calibration/Ranking \\
\midrule
\addlinespace[0.1em]
\midrule
BayesPE~\citep{Tonolini2024BayesPE} & Ensemble & Input & White-box & Post-hoc & Low & Calibration\\
Deep Ensembles~\citep{lakshminarayanan2017DeepEnsemble} & Ensemble & Parameter & White-box & Full training & High & Calibration/Ranking \\
NeuBoots~\citep{Shin2021NeuralBootstrapper} & Ensemble & Parameter & White-box & Full training & High & Calibration/Ranking \\
NES~\citep{Zaidi2021NES} & Ensemble & Parameter & White-box & Full training & High & Calibration/Ranking \\
Packed-Ensembles~\citep{Laurent2023Packed-Ensembles} & Ensemble & Parameter & White-box & Full training & High & Calibration/Ranking \\
DEN~\citep{Kharbanda2024DEN} & Ensemble & Parameter & White-box & Full training & High & Calibration/Ranking \\
CreDEs~\citep{Wang2024CreDEs} & Ensemble & Parameter & White-box & Full training & High & Calibration/Ranking \\
CED~\citep{Wang2026CED} & Ensemble & Parameter & White-box & Full training & High & Calibration/Ranking \\
LoRA Ensembles~\citep{wang2023LoRAEnsemble} & Ensemble & Parameter & White-box & Fine-tuning & High & Calibration \\
LoRAEs-UQ~\citep{Balabanov2024LoRAEs-UQ} & Ensemble & Parameter & White-box & Fine-tuning & High & Calibration \\
RMI~\citep{malinin2020RMI} & Ensemble & Parameter/Token-level & White-box & Post-hoc & Medium & Ranking \\
\midrule
\addlinespace[0.1em]
\midrule
Self-Detection~\citep{zhao2024Self-Detection} & Consensus-based & Input/Decoding & Black-box & Post-hoc & Low & Ranking \\
ICL-UQ~\citep{ling2024UQ_ICL} & Consensus-based & Input/Decoding & Black-box & Post-hoc & Low & Calibration \\
ICE~\citep{Hou2024ICE} & Consensus-based & Input/Decoding & Black-box & Post-hoc & Low & Calibration \\
SPUQ~\citep{Gao2024SPUQ} & Consensus-based & Input/Decoding & Black-box & Post-hoc & Low & Calibration \\
SAC$^3$~\citep{zhang2023sac3} & Consensus-based & Input/Decoding & Black-box & Post-hoc & Medium & Ranking \\
EigV-J~\citep{lin2024numset/eigv/deg/ecc} & Consensus-based & Decoding & Black-box & Post-hoc & Low & Ranking \\
Deg-J~\citep{lin2024numset/eigv/deg/ecc} & Consensus-based & Decoding & Black-box & Post-hoc & Low & Ranking \\
Ecc-J~\citep{lin2024numset/eigv/deg/ecc} & Consensus-based & Decoding & Black-box & Post-hoc & Low & Ranking \\
MI Estimator~\citep{yadkori2024MI_Estimator} & Consensus-based & Decoding & Black-box & Post-hoc & Low & Ranking \\
CLM~\citep{quach2024CLM} & Consensus-based & Decoding & Black-box & Post-hoc & Low & Coverage/Ranking \\
ConU~\citep{wang2024ConU} & Consensus-based & Decoding & Black-box & Post-hoc & Low & Coverage/Ranking \\
LexSim~\citep{fomicheva2020PPL} & Consensus-based & Decoding & Black-box & Post-hoc & Low & Ranking \\
SelfCheckGPT~\citep{manakul2023selfcheckgpt} & Consensus-based & Decoding & Black-box & Post-hoc & Medium & Ranking \\
NumSet~\citep{lin2024numset/eigv/deg/ecc} & Consensus-based & Decoding & Black-box & Post-hoc & Medium & Ranking \\
EigV~\citep{lin2024numset/eigv/deg/ecc} & Consensus-based & Decoding & Black-box & Post-hoc & Medium & Ranking \\
Deg~\citep{lin2024numset/eigv/deg/ecc} & Consensus-based & Decoding & Black-box & Post-hoc & Medium & Ranking \\
Ecc~\citep{lin2024numset/eigv/deg/ecc} & Consensus-based & Decoding & Black-box & Post-hoc & Medium & Ranking \\
BSDetector~\citep{chen2024BSDETECTOR} & Consensus-based & Decoding & Black-box & Post-hoc & Medium & Ranking \\
D-UE~\citep{Da2024D-UE} & Consensus-based & Decoding & Black-box & Post-hoc & Medium & Ranking \\
CSS~\citep{Ao2024CSS} & Consensus-based & Decoding & Black-box & Post-hoc & Medium & Ranking \\
LUQ~\citep{zhang2024luq} & Consensus-based & Decoding & Black-box & Post-hoc & Medium & Ranking \\
KLE~\citep{nikitin2024KLE} & Consensus-based & Decoding & Black-box & Post-hoc & Medium & Ranking \\
MD-UQ~\citep{chen2025MD-UQ} & Consensus-based & Decoding & Black-box & Post-hoc & Medium & Ranking \\
LofreeCP~\citep{Su2024LofreeCP} & Consensus-based & Decoding & Black-box & Post-hoc & Medium & Coverage/Ranking \\
EigenScore~\citep{Chen2024INSIDE} & Consensus-based & Decoding & White-box & Post-hoc & Low & Ranking \\
CoCoA~\citep{vashurin2025CoCoA} & Consensus-based & Decoding/Token-level & White-box & Post-hoc & Low & Ranking \\
\bottomrule
\end{tabular} }
\end{table}
\FloatBarrier

\begin{table}
    \centering
    \resizebox{\textwidth}{!}{
\begin{tabular}{llccccc}
\toprule
\textbf{Methods} & \textbf{Type} & \textbf{Uncertainty Sources} & \textbf{Model Accessibility} & \textbf{Training Necessity} & \textbf{Consumption} & \textbf{Evaluation} \\
\midrule
Semantic Entropy~\citep{farquhar2024SemanticEntropy} & Consensus-based & Decoding/Token-level & White-box & Post-hoc & Medium & Ranking \\
Semantic Density~\citep{Qiu2024SemanticDensity} & Consensus-based & Decoding/Token-level & White-box & Post-hoc & Medium & Ranking \\
SentSAR~\citep{Duan2024SAR} & Consensus-based & Decoding/Token-level & White-box & Post-hoc & Medium & Ranking \\
SAR~\citep{Duan2024SAR} & Consensus-based & Decoding/Token-level & White-box & Post-hoc & Medium & Ranking \\
Cycles of Thought~\citep{becker2024Cycles_of_Thought} & Consensus-based & Reasoning & Black-box & Post-hoc & Medium & Ranking \\
Topo-UQ~\citep{Da2025Topo-uq} & Consensus-based & Reasoning & Black-box & Post-hoc & Medium & Ranking \\
CoT Entropy~\citep{Ye2025CoTEntropy} & Consensus-based & Reasoning & White-box & Post-hoc & Low & Ranking \\
TouT~\citep{mo2024TouT} & Consensus-based & Reasoning & White-box & Post-hoc & Medium & Ranking \\
\midrule
\addlinespace[0.1em]
\midrule
SQR~\citep{tagasovska2019SQR/OCs} & Single-pass & Input & White-box & Full training & High & Coverage/Calibration \\
MD~\citep{lee2018MD} & Single-pass & Input & White-box & Post-hoc & Low & Ranking \\
RainProof~\citep{Darrin2023RainProof} & Single-pass & Input & White-box & Post-hoc & Low & Ranking/Coverage \\
HUQ~\citep{vazhentsev2023HUQ} & Single-pass & Input/Parameter & White-box & Full training & High & Ranking/Calibration \\
Feature-Gaps~\citep{bakman2025Feature-Gaps} & Single-pass & Input/Decoding & White-box & Fine-tuning & High & Ranking \\
RMD~\citep{ren2022RMD} & Single-pass & Input/Decoding & White-box & Post-hoc & Low & Ranking \\
OCs~\citep{tagasovska2019SQR/OCs} & Single-pass & Parameter & White-box & Full training & High & Ranking \\
SATRMD~\citep{vazhentsev2025SATRMD} & Single-pass & Token-level & White-box & Fine-tuning & High & Ranking \\
LARS~\citep{yaldiz2025LARS} & Single-pass & Token-level & White-box & Fine-tuning & High & Ranking/Calibration \\
MaxSeqProb~\citep{Vashurin2025LM-Polygraph} & Single-pass & Token-level & White-box & Post-hoc & Low & Ranking \\
MeanTokEnt~\citep{fomicheva2020PPL} & Single-pass & Token-level & White-box & Post-hoc & Low & Ranking \\
MaxTokProb~\citep{fomicheva2020PPL} & Single-pass & Token-level & White-box & Post-hoc & Low & Ranking \\
PPL~\citep{fomicheva2020PPL} & Single-pass & Token-level & White-box & Post-hoc & Low & Ranking \\
Focus~\citep{zhang2023focus} & Single-pass & Token-level & White-box & Post-hoc & Low & Ranking \\
CSL~\citep{lin2024CSL} & Single-pass & Token-level & White-box & Post-hoc & Low & Ranking \\
BoostedProb~\citep{dinh2025generative} & Single-pass & Token-level & White-box & Post-hoc & Low & Ranking \\
CP-MCQA~\citep{Kumar2023Conformal_Prediction} & Single-pass & Token-level & White-box & Post-hoc & Low & Coverage/Ranking \\
PMI~\citep{Junya2019PMI} & Single-pass & Token-level & White-box & Post-hoc & Medium & Ranking \\
Conditional PMI~\citep{Poel2022Conditional_PMI} & Single-pass & Token-level & White-box & Post-hoc & Medium & Ranking \\
MARS~\citep{Bakman2024MARS} & Single-pass & Token-level & White-box & Post-hoc & Medium & Ranking \\
CCP~\citep{Fadeeva2024CCP} & Single-pass & Token-level & White-box & Post-hoc & Medium & Ranking \\
TokenSAR~\citep{Duan2024SAR} & Single-pass & Token-level & White-box & Post-hoc & Medium & Ranking \\
RAUQ~\citep{vazhentsev2025RAUQ} & Single-pass & Token-level/Decoding & White-box & Post-hoc & Low & Ranking \\
FrequencyScoring~\citep{cherian2024conditional_boosting} & Single-pass & Decoding & Black-box & Post-hoc & Low & Coverage \\
Conditional Boosting~\citep{cherian2024conditional_boosting} & Single-pass & Decoding & Black-box & Post-hoc & Low & Coverage/Calibration \\
Level-adaptive CP~\citep{cherian2024conditional_boosting} & Single-pass & Decoding & Black-box & Post-hoc & Low & Coverage/Calibration \\
HaloScope~\citep{du2024haloscope} & Single-pass & Decoding & White-box & Fine-tuning & High & Ranking \\
SEPs~\citep{Kossen2024Semantic_Entropy_Probes} & Single-pass & Decoding & White-box & Fine-tuning & High & Ranking \\
BLL~\citep{Dakhmouche2025BLL} & Single-pass & Decoding & White-box & Fine-tuning & High & Ranking/Calibration \\
UaIT~\citep{liu2024UaIT} & Single-pass & Decoding & White-box & Fine-tuning & High & Calibration \\
Supervised UE~\citep{liu2024Supervised_estimation} & Single-pass & Decoding & White-box & Fine-tuning & High & Calibration/Ranking \\
LACIE~\citep{Stengel-Eskin2024LACIE} & Single-pass & Decoding & White-box & Fine-tuning & High & Calibration/Ranking \\
Factoscope~\citep{he2023Factoscope} & Single-pass & Decoding/Token-level & White-box & Fine-tuning & High & Ranking \\
LLM-Check~\citep{sriramanan2024llm-check} & Single-pass & Decoding/Token-level & White-box & Post-hoc & Low & Ranking \\
SaySelf~\citep{Xu2024SaySelf} & Single-pass & Reasoning & White-box & Fine-tuning & High & Calibration \\
Linguistic Calibration~\citep{Band2024Linguistic_Calibration} & Single-pass & Reasoning & White-box & Fine-tuning & High & Calibration \\
UQAC~\citep{li2025UQAC} & Single-pass & Reasoning/Token-level & White-box & Post-hoc & Low & Calibration/Ranking \\
CoT-UQ-AP~\citep{Zhang2025CoT-UQ} & Single-pass & Reasoning/Token-level & White-box & Post-hoc & Medium & Ranking \\
Verbalized Confidence~\citep{tian2023Verbalized_Confidence} & Single-pass & - & Black-box & Post-hoc & Low & Calibration \\
CoT-UQ-SE~\citep{Zhang2025CoT-UQ} & Single-pass & - & Black-box & Post-hoc & Medium & Ranking \\
Verbalized Probability~\citep{Lin2022Verbalized_Probability} & Single-pass & - & White-box & Fine-tuning & High & Calibration \\
$P(\text{IK})$~\citep{Kadavath2022Ptrue/PIK} & Single-pass & - & White-box & Fine-tuning & High & Calibration/Ranking \\
$P(\text{True})$~\citep{Kadavath2022Ptrue/PIK} & Single-pass & - & White-box & Post-hoc & Low & Calibration/Ranking \\
\bottomrule
\end{tabular} }
\end{table}
\FloatBarrier

\clearpage

\paragraph{Parameter Uncertainty.}
Parameter uncertainty arises from the inherent stochasticity associated with the model parameters $\theta$. 
While deployed LLMs typically operate with a fixed set of model weights, they can be interpreted through a probabilistic lens as a single realization of a random variable $\Theta$. 
Upon feeding an input instance $x$ into the model, distinct but plausible parameter configurations may induce varying token-level distributions $P_\mathcal{V}(x, \Theta)$ over the vocabulary $\mathcal{V}$. Specifically, even with a constant input and deterministic decoding strategy, uncertainty within the parameter space can shift the probability mass across candidate tokens, thereby causing fluctuations in the induced token-level distribution and ultimately the generated output. Consequently, parameter uncertainty quantifies the propagation of parametric stochasticity to the token-level distributions in LLM-based generation.

\paragraph{Token-level Uncertainty.}
Token-level uncertainty arises from the inherent stochasticity encoded within the token-level distributions $P_\mathcal{V}(x, \theta)$ over the vocabulary $\mathcal{V}$. During the LLM autoregressive generation process, the model yields such a distribution at each generation step, the topology of which reflects how certain the model is about the next token. Specifically, a peaked distribution that probability mass is concentrated on a restricted set of candidate tokens typically indicates lower uncertainty; conversely, a flatter distribution implies that the model assigns comparable probabilities across multiple tokens, indicating diminished certainty in the next generation step.

\paragraph{Decoding Uncertainty.}
Decoding uncertainty arises from the inherent stochasticity led by the decoding process, which maps token-level distributions to final output sequences. Even when the input instance $x$, the model parameters $\theta$, and the resulting token-level distributions $T=P_\mathcal{V}(x, \theta)$ are held constant, divergent decoding strategies $F_\text{decoding}(\cdot)$ can yield heterogeneous output sequences, i.e., $Y = F_\text{decoding}(T)$. Deterministic decoding approaches, such as greedy search, operate under rigid rules to minimize randomness during the selection process. In contrast, generative decoding strategies, including ancestral sampling, temperature scaling, top-$k$ sampling, and nucleus sampling~\citep{Guo2017ECE,Lewis2018topk,Holtzman2020topp}, sample tokens from modified or truncated token-level distributions, thereby injecting additional variability into the generation process. Consequently, the decoding uncertainty quantifies the variability introduced during the transformation of token-level distributions into concrete generated text.

\paragraph{Discussions on Uncertainty Sources.} The aforementioned four uncertainty sources arise from different levels of the LLM generation process, thus having different effects. From the perspective of the LLM generation process, input and parameter uncertainties shape the induced token-level distributions, while token-level uncertainty further affects the variability available to the decoding process, where uncertainty may propagate and accumulate across autoregressive generation steps. From the perspective of influence relations, both input and parameter uncertainties originate from their inherent stochasticity; there is no significant causal relationship between these two uncertainty sources. Input and parameter uncertainties collectively influence both token-level and decoding uncertainties, with the former also having a direct impact on the latter. Notably, both token-level and decoding uncertainties are also subject to their inherent stochasticity; for instance, the logits generated by LLMs are actual distributions of some latent random variables. Beyond these stochastic factors, token-level and decoding uncertainties are further modulated by model capacity and decoding strategies, respectively. For example, LLMs with a smaller size or less capacity tend to exhibit a higher propensity for hallucination.

Significantly, the traditional uncertainty attribution paradigm, in which aleatoric uncertainty captures the inherent stochasticity in data and epistemic uncertainty reflects model uncertainty caused by limited knowledge or insufficient data, provides an important conceptual foundation but may fail to attribute and quantify the effects of each uncertainty source during the LLM generation process. Our proposed uncertainty taxonomy, grounded in the structural formulation of the LLM generation process, fixes this issue by relating the proposed source-based taxonomy to the classical decomposition of uncertainty into aleatoric and epistemic components~\citep{Hllermeier2021Aleatoric_and_epistemic_uncertainty}. For example, decoding uncertainty arises from stochastic generation procedures, while token-level uncertainty reflects the concentration or dispersion of predictive distributions over the vocabulary; these factors are not directly operationalized in the classical aleatoric-epistemic framework. Moreover, in autoregressive language generation, data-related and model-related uncertainty may interact through the token-level distribution and propagate across generation steps, making their boundary difficult to disentangle within the standard inference pipeline. Therefore, by explicitly associating uncertainty sources with distinct and computable components of the generation pipeline, namely the input, model parameters, token-level distribution, and decoding strategy, our proposed uncertainty taxonomy provides a more mechanistic, fine-grained, and process-aware view of uncertainty in LLMs. 

Reasoning uncertainty that has received increasing attention indicates another uncertainty that emerges during multi-step LLM generation and reasoning-intensive tasks. In contrast to the input uncertainty, parameter uncertainty, token-level uncertainty, and decoding uncertainty that correspond to a primitive source tied to specific components of the generation pipeline, the reasoning uncertainty characterizes the variability of intermediate reasoning trajectories, rationales, or step-wise conclusions produced during the LLM autoregressive generation. Specifically, when LLMs explicitly generate reasoning traces, such as chain-of-thought rationales~\citep{Zhang2025CoT-UQ}, intermediate derivations~\citep{Xu2024SaySelf}, or step-wise explanations~\citep{Da2025Topo-uq}, distinct but plausible reasoning paths may culminate in divergent final answers or arrive at identical conclusions through inconsistent rationales. Under the proposed uncertainty taxonomy, reasoning uncertainty is a higher-order manifestation of LLM uncertainty and is a consequence induced by the proposed four uncertainty sources; it can be conceptualized as the propagation and accumulation of token-level and decoding uncertainties across successive reasoning steps, frequently modulated by upstream input and parameter uncertainties.

\subsection{Types of LLM Uncertainty Quantification Methods}  \label{subsec:UQ_types}
In this subsection, we categorize existing UQ methods for LLMs into four major types, that is, Bayesian, ensemble, consensus-based, and single-pass methods, according to their estimation mechanisms. Table~\ref{tab:uq_methods} lists the existing UQ methods grouped into these four types.

\paragraph{Bayesian Methods.}
Bayesian methods quantify uncertainty by modeling the stochastic source as a random variable and estimating its posterior distribution. 
Bayesian methods often emerge from the parameter space that incorporates the parameter uncertainty into the token-level distributions. Hence, the Bayesian token-level distributions can be written as
\begin{equation} \label{eq:bayesian}
 P_{\mathcal V}^{\mathrm{Bayes}}(x,\Theta)
= \!\int_{\Omega} P_{\mathcal V}(x, \theta) \, p(\theta \mid\! D) \, \dif \theta   \ ,
\end{equation}
where $p(\theta \mid\! D)$ denotes the posterior distribution of the model parameters given training data $D$ and $\theta \in \Omega$. Eq.~\eqref{eq:bayesian} provides a typical UQ paradigm that primarily captures parameter uncertainty by integrating all plausible parameters $\theta \in \Omega$ over the vocabulary. It is observed from Eq.~\eqref{eq:bayesian} that the key of Bayesian UQ methods is to obtain the collection of all plausible parameters $\Omega$ and estimate the posterior distribution for any $\theta \in \Omega$. 

However, exact posterior inference is often intractable for LLMs in practice; inspired by conventional paradigms, many approximation inference methods such as variational inference, Monte Carlo approximation, Laplace approximation and dropout-based sampling~\citep{jospin2022BNNs} are typically employed. 
Laplace-LoRA~\citep{yang2023laplace-lora} is the representative work that applies a Laplace approximation to low-rank adapter parameters after fine-tuning the LLMs, thus introducing a tractable posterior over LoRA updates while keeping the backbone model fixed. 
BLoB~\citep{Wang2024BLoB} further formulates LoRA parameters probabilistically and optimizes their posterior distribution during fine-tuning, enabling Bayesian low-rank adaptation by backpropagation. 
Overall, Bayesian UQ methods provide probabilistically grounded estimates of parameter uncertainty in LLMs through characterizing posterior distributions over parameters.

\paragraph{Ensemble Methods.} 
Ensemble methods primarily characterize parameter uncertainty by constructing an empirical posterior distribution over models to replace the exact or approximate estimation of the posterior distribution in Bayesian methods. The empirical posterior distribution is led by multiple model realizations, such as different plausible parameters or even different architectures. The key idea is that if a set of models trained or conducted on the same dataset manifests different functional mappings, their predictive diversity can be regarded as a manifestation of uncertainty. Accordingly, the token-level distributions with respect to ensemble methods can be written as
\begin{equation}  \label{eq:ensemble}
    P_{\mathcal V}^{\mathrm{ens}}(x, \Theta) = \frac{1}{K} \sum_{k=1}^{K} \alpha_k \,P_{\mathcal V}(x, \theta_k) \ ,
\end{equation}
where $\{ \theta_k \}_{k=1}^K$ denotes the collection of plausible model realizations, $P_{\mathcal V}(x, \theta_k)$ denotes the predictive distributions of the $k$-th model given input $x$, and $\alpha_k$ denotes the scaling factor that corresponds to the joint distribution coefficient relative to $P_{\mathcal V}(x, \theta_k)$ of Bayesian methods and usually performs as the voting coefficient in conventional ensemble learning~\citep{zhou2025ensemble}. 

It is observed that ensemble methods avoid explicit posterior modeling but rely heavily on the construction of model realizations, leveraging stochasticity in training (e.g., random initialization or optimization trajectories). 
Ensemble methods primarily characterize the parameter uncertainty, as different parameter realizations $\theta_k$ correspond to different predictive functions, while token-level uncertainty is reflected in the averaged predictive distributions. 
Representative studies of ensemble UQ methods for LLMs contain Deep Ensembles~\citep{lakshminarayanan2017DeepEnsemble} that train multiple independently initialized models and average their predictive distributions to approximate uncertainty, and LoRA-based ensembles~\citep{wang2023LoRAEnsemble} that adapts to LLMs by attaching multiple low-rank adapters to a shared backbone, enabling parameter-efficient ensemble construction within standard fine-tuning workflows.

\paragraph{Consensus-based Methods.} Consensus-based methods primarily characterize decoding uncertainty and input uncertainty through behavioral patterns in the generation space. Provided a pre-trained LLM with fixed model parameters, consensus-based methods quantify uncertainty by measuring the consistency or inconsistency among multiple generated outputs that are obtained via repeated sampling or multiple inference runs. The key idea is that consistent outputs across multiple runs indicate higher reliability, while significant divergence in semantic, syntactic, or logical structure reflects unstable model behavior, that is, higher uncertainty. This variability may be caused by perturbed inputs that introduce variability in the generation process under fixed model parameters and further reflected in the token-level distributions, thereby giving rise to token-level uncertainty. Representative approaches include Semantic Entropy~\citep{farquhar2024SemanticEntropy}, which first groups multiple sampled generations into semantically equivalent clusters and then computes entropy over the cluster distribution constructed from their associated likelihoods. This approach primarily characterizes decoding uncertainty, while the use of generation likelihoods reflects token-level uncertainty. Another line of work is Self-Prompting Uncertainty Quantification~\citep{Gao2024SPUQ}, which conducts prompt perturbations and aggregates predictions across multiple prompts.

\paragraph{Single-pass Methods.}
Single-pass methods primarily characterize token-level uncertainty, as the token-level distributions explicitly assign probabilities to candidate tokens, which directly reflect the model's confidence over possible outcomes while also capturing decoding uncertainty through the variability reflected in the generation process. 
The key idea of single-pass methods lies in leveraging the token-level distributions formed during a single generation process, or alternatively, eliciting confidence signals through model self-evaluation, without requiring repeated sampling or model modification. Unlike consensus-based methods that uncertainty estimates are inherently derived from the disagreement or consistency among multiple generated responses, single-pass methods are able to produce a valid uncertainty score from one generation or one forward pass alone. It is worth noting that a single-pass method can also be applied repeatedly across multiple independent generations and then averaged to obtain a more stable uncertainty estimate. However, this repeated use is optional rather than required. 

Representative approaches include LACIE~\citep{Stengel-Eskin2024LACIE}, which calibrates confidence by aligning model outputs with preference signals derived from a speaker–listener framework. Some approaches may involve additional training, such as calibration or probing, but the resulting uncertainty estimates are still obtained from a single forward pass at inference time. In particular, some single-pass methods obtain uncertainty estimates through self-prompting. Instead of deriving uncertainty from the sources explicitly characterized in our framework, such as input uncertainty, parameter uncertainty, token-level uncertainty, or decoding uncertainty, these methods directly ask the LLM to evaluate its own response. A typical example is $P(\text{True})$~\citep{Kadavath2022Ptrue/PIK}, which elicits self-reported confidence by prompting the model to assess the correctness of its own predictions. The resulting uncertainty estimate comes from the model's self-assessment ability, rather than from uncertainty signals explicitly exposed during the generation process. However, this approach evidently accumulates the uncertainty that an LLM generates predictions and the other one that an LLM produces judgments.

\subsection{Model Accessibility}  \label{subsec:access}
Model accessibility describes the extent to which a UQ method relies on model parameters or modifiable components of LLMs. 
According to the required access level, existing UQ methods can be broadly divided into black-box and white-box methods. 
Black-box methods estimate uncertainty only from observable model outputs via API, without access to internal states, token-level probabilities, gradients, or model parameters, and typically do not allow additional training or fine-tuning of the target model. 
In contrast, white-box methods can extract information, such as token-level probabilities, logits, hidden states, gradients, or parameters, from model internals; sometimes, one may quantify and further reduce the LLM uncertainty by re-training or fine-tuning. Overall, model accessibility reflects a practical trade-off between applicability and informativeness. 
Black-box methods are easier to deploy across closed-source systems, whereas white-box methods can obtain more precise uncertainty estimates by exploiting richer internal signals.

\paragraph{Black-box.}
Black-box methods operate under restricted-access settings, where only the generated responses are observable, and do not require access to internal model states, token-level distributions, gradients, or model parameters, and generally cannot modify the target model through additional training or fine-tuning. Thereby, only uncertainties from output-level properties, such as semantic consistency, response diversity, reasoning structure, or agreement among multiple generated answers, can be quantified. For example, Topo-UQ~\citep{Da2025Topo-uq} constructs reasoning topologies as graphs from multiple generated explanations and quantifies uncertainty by measuring semantic-aware graph differences across these structures. Larger discrepancies among reasoning graphs indicate higher variability in the reasoning process and thus higher uncertainty. Black-box methods take flexible and model-agnostic advantages that are particularly suitable for closed-source LLMs and API-based applications. However, their estimates are usually less fine-grained, as they can only infer uncertainty indirectly from observable outputs.

\paragraph{White-box.}
It is necessary for white-box methods to access internal information of LLMs, such as token-level distributions, logits, hidden states, gradients, or model parameters. By leveraging these signals, they estimate uncertainty through the structure of token-level distributions, intermediate representations, or parameter-level variability. For example, MARS~\citep{Bakman2024MARS} utilizes token-level probabilities of generated sequences and constructs a response score by weighting tokens according to their semantic contribution, replacing length-normalized likelihood scoring. This approach reflects the model's confidence over generated content, where lower scores indicate lower confidence and thus higher uncertainty. Compared with black-box methods, white-box methods can provide more fine-grained and distribution-aware uncertainty estimates, but they are less applicable when only API access is available or when internal model information is hidden.

\subsection{Training Necessity}  \label{subsec:training_necessity}
Training necessity describes the extent to which a UQ method requires parameter updates. From this perspective, existing methods can be broadly characterized at three levels: post-hoc estimation without parameter updates, fine-tuning with limited parameter modifications, and full training with substantial updates to model parameters. Post-hoc methods quantify uncertainty directly from existing outputs, token-level distributions, or internal signals without modifying the target model. Fine-tuning methods introduce and update a limited number of trainable parameters, such as LoRA modules, adapters, or calibration heads, while keeping most of the backbone model fixed. In contrast, full-training methods require training the entire model, a substantial portion of its parameters, or multiple full model instances, resulting in the highest training cost and strongest dependence on additional optimization.

\paragraph{Post-hoc.}
At the post-hoc level, the uncertainty is estimated without any parameter updates to the target model. 
These methods compute uncertainty scores from existing information, such as generated responses, token-level distributions, logits, hidden states, or semantic consistency among outputs. 
As a result, they are computationally efficient and can be readily applied to off-the-shelf LLMs, provided that the required output-level or internal signals are accessible. 
A representative example is SelfCheckGPT~\citep{manakul2023selfcheckgpt}, which requires no additional training of the target model. 
It generates a target answer together with multiple sampled answers, evaluates the consistency between each sample and the target, and aggregates the resulting scores into a uncertainty estimate. 
Lower consistency indicates that the model's generations are less stable and therefore more uncertain.

\paragraph{Fine-tuning.}
Some methods require limited parameter updates to improve or enable uncertainty estimation.
Instead of retraining the entire LLM, these methods typically update lightweight modules, task-specific heads, or selected model parameters while keeping most of the backbone model fixed. 
The additional trainable parameters can be optimized using supervised labels, calibration objectives, preference data, or reward signals related to uncertainty estimation. 
A representative example is SaySelf~\citep{Xu2024SaySelf}, which teaches LLMs to express fine-grained confidence estimates together with self-reflective rationales. 
SaySelf first constructs model-specific training data by summarizing uncertainty from inconsistencies among multiple sampled reasoning chains, and then uses supervised fine-tuning to train the model to generate confidence estimates and uncertainty-aware rationales. 
It further applies reinforcement learning with a reward function designed to encourage accurate high-confidence predictions and penalize overconfidence on incorrect responses. 
In this way, fine-tuning methods can directly improve the model's ability to verbalize and calibrate uncertainty, but they require additional optimization and training data beyond post-hoc estimation.

\paragraph{Full training.}
Methods under full training require updating the entire model, a substantial portion of its parameters, or modifying and training core model components to explicitly support uncertainty estimation. 
These methods often involve training uncertainty-aware architectures, parameter distributions, or modules that are integrated into the model's forward computation. 
For example, Sparse Gaussian Process Attention (SGPA)~\citep{Chen2023SGPA} replaces the scaled dot-product attention in Transformers with a kernel-based attention mechanism and applies sparse Gaussian process techniques to approximate posterior uncertainty in the output space of multi-head attention blocks. 
By incorporating Bayesian inference directly into the attention mechanism, SGPA improves calibration and uncertainty estimation, but requires architectural modification and additional training of the Transformer model. 
Such methods can provide strong uncertainty modeling capabilities, but they incur significantly higher computational cost and are less practical for large-scale LLMs unless substantial training resources are available.
\subsection{Consumption}  \label{subsec:consumption}

In this subsection, we categorize LLM UQ methods into three types, that is, low, medium, and high, according to the additional computational cost required to obtain uncertainty estimates. Low-consumption methods compute uncertainty through mathematical statistics over available outputs or require only a single model call. Medium-consumption methods require multiple model calls, such as repeated sampling, multiple self-evaluation queries, or additional calls to external evaluators. High-consumption methods require training, fine-tuning, or maintaining multiple model instances.

\paragraph{Low Consumption.}
Low-consumption methods either compute uncertainty scores directly from already available information, such as generated tokens, token-level probabilities, sequence likelihoods, and predictive distributions, or obtain uncertainty estimates through a single model call, thus inducing minimal additional computational overhead. These methods typically rely on simple mathematical statistics or direct self-assessment from the model; many single-pass methods fall into this category. For example, ConU~\citep{wang2024ConU} applies conformal prediction to transform heuristic uncertainty scores into calibrated prediction sets with correctness coverage guarantees, relying on mathematical calibration rather than model training or repeated external evaluation.

\paragraph{Medium Consumption.}
Medium-consumption methods do not update the parameters of the target model but usually rely on generating multiple responses, querying the model several times, or calling external evaluators such as Natural Language Inference (NLI) models~\citep{liu2019roberta}, verifiers, or embedding models. Hence, many consensus-based methods fall into this category. For example, Semantic Entropy~\citep{farquhar2024SemanticEntropy} operates at this level by sampling multiple responses and performing semantic clustering and consistency evaluation using an external NLI model, specifically a DeBERTa-large model~\citep{he2021deberta} fine-tuned on the MNLI dataset~\citep{williams2018MultiNLI}.

\paragraph{High Consumption.}
High-consumption methods obtain the uncertainty estimates by training uncertainty-aware components or constructing multiple predictors, thus requiring substantial computational resources. Instead of direct mathematical scoring or repeated inference, they usually involve training, fine-tuning, or maintaining multiple model instances. Representative examples include BLoB~\citep{Wang2024BLoB} and SaySelf~\citep{Xu2024SaySelf}.

\subsection{Evaluation}  \label{subsec:evaluation}
In this paper, we consider three major evaluation objectives, involving ranking, calibration, and coverage, for measuring the performance of different LLM UQ methods and assessing whether an uncertainty estimate is useful in downstream applications. Ranking focuses on whether uncertainty scores can distinguish reliable outputs from unreliable ones; calibration evaluates whether confidence or probability estimates match empirical correctness; and coverage concerns whether a method can provide formal guarantees for prediction sets or abstention decisions. The evaluation methods above provide a principled basis for selecting appropriate UQ approaches in empirical comparison.

From another perspective, different UQ methods naturally emphasize different evaluation objectives. Consensus-based and single-pass methods are often well-suited for ranking because they provide relative uncertainty signals. UQ Methods that require training or fine-tuning are frequently designed to improve calibration by aligning confidence estimates with ground-truth correctness through supervised learning or reinforcement learning. In contrast, UQ methods relative to conformal prediction are primarily associated with coverage guarantees. 

\paragraph{Ranking.}
Ranking measures how well uncertainty scores separate correct from incorrect outputs~\citep{Kadavath2022Ptrue/PIK,farquhar2024SemanticEntropy,lin2024numset/eigv/deg/ecc}, thus supporting risk-aware selection and being good at distinguishing high-risk outputs from low-risk ones of LLMs, especially suffering from hallucination detection. In general, an uncertainty score does not necessarily need to be perfectly calibrated as a probability; instead, it should assign higher uncertainty to incorrect, hallucinated, or unreliable responses. Common ranking-oriented metrics include Area Under the Receiver Operating Characteristic (AUROC), Area Under the Precision-Recall Curve (AUPRC), and Area Under the Accuracy-Rejection Curve (AUARC).

\paragraph{Calibration.}
Calibration evaluates whether the confidence or probability estimates produced by a UQ method align with empirical correctness frequencies~\citep{brier1950score,Guo2017ECE}. A well-calibrated method should assign confidence scores that correspond to the actual likelihood of being correct. Calibration is especially important for model deployment, reliability assessment, and human-AI decision-making, where users may rely on confidence estimates to decide whether to trust, verify, or reject a model output. Typical calibration metrics include Expected Calibration Error (ECE) and Maximum Calibration Error (MCE).

\paragraph{Coverage.}
Coverage aims at evaluating whether an uncertainty method can provide reliable guarantees over prediction sets, confidence sets, or abstention decisions. It works by constructing a set of possible answers or accept/reject decisions that satisfy a desired coverage level under appropriate assumptions~\citep{vovk2005CP,Geifman2019AURC}; thus, its objective is commonly associated with conformal prediction. In contrast to ranking and calibration that mainly assess the quality of scalar uncertainty or confidence scores, coverage focuses on whether the uncertainty method can support formal reliability guarantees. Coverage is particularly relevant in high-stakes scenarios where controlling the probability of erroneous acceptance is more important than simply ranking outputs.

\section{Experiments}  \label{sec:experiments}
In this section, we empirically evaluate representative UQ methods of LLMs on real-world tasks. The experiments are performed to answer the following questions.
\begin{itemize}
    \item[Q1.] In general, the uncertainty scores of LLMs are typically unsupervised, that is, lacking ground-truth labels of LLM UQ methods. How do developers empirically evaluate the effectiveness of LLM UQ methods?
    \item[Q2.] How do representative UQ methods of LLMs compare across real-world LLM tasks involving question answering, reasoning, and code generation tasks?
    \item[Q3.] Whether and to what extent does the model complexity of LLMs affect their uncertainties, or equally, the effectiveness of the investigated UQ methods?
\end{itemize}

\subsection{Configurations}
All experiments are conducted on one NVIDIA RTX 6000 GPU. Here, we collect 21 typical UQ methods from Table~\ref{tab:uq_methods} based on two criteria of whether they apply to LLMs and whether their codes are open-source or reproducible. The conducted LLM UQ methods are listed in Table~\ref{qwen3_1_7b_answer_only_results}. We conduct experiments of both open-source and closed-source LLMs; the former includes the Qwen3 family~\citep{Qwen_Team2025qwen3} at scales of 1.7B and 4B and the Llama 3.2 family~\citep{Llama_Team2024llama} at scales of 1B and 3B, while the latter contains DeepSeek-V3~\citep{DeepSeek-AI2024deepseek} accessed through the DeepSeek API. The evaluated benchmarks include question-answering tasks such as TriviaQA, as well as reasoning tasks such as mathematical reasoning on GSM8K and code generation on HumanEval. TriviaQA is a question-answering benchmark containing 95K question-answer pairs and over 650K question-answer-evidence triples~\citep{Joshi2017triviaqa}, where each example requires an LLM to generate a short textual answer. GSM8K is a grade-school mathematical reasoning benchmark containing 8.5K math word problems~\citep{Cobbe2021gsm8k}, where each example requires an LLM to solve a multi-step arithmetic problem and generate the final numerical answer. HumanEval is a code generation benchmark containing 164 hand-written Python programming problems~\citep{Chen2021humaneval}, where each example provides a function signature and a natural-language docstring, and the LLM-generated code is evaluated by unit tests.

Now, we introduce the evaluation framework of LLM UQ methods. The dimensions of evaluation metrics in seminal studies can be roughly divided into three categories: ranking, calibration, and coverage, which have been introduced in Subsection~\ref{subsec:evaluation}. Specifically, ranking focuses on whether uncertainty scores can distinguish correct responses from incorrect ones, including AUROC~\citep{Davis2006AUROC}, AUPRC~\citep{Davis2006AUROC}, and AUARC~\citep{nadeem2009AUARC,lin2024numset/eigv/deg/ecc}, where a larger ranking value indicates a stronger ability of LLM UQ methods to distinguish correct responses from incorrect ones. Calibration evaluates whether confidence or probability estimates match empirical correctness, including ECE~\citep{Guo2017ECE}, MCE~\citep{Guo2017ECE}, and Brier Score (BS)~\citep{brier1950score}, where a lower calibration value indicates that the produced probabilities relative to LLM UQ methods are more consistent with the observed correctness frequencies. Coverage concerns whether a method can provide formal guarantees for prediction sets or abstention decisions, including Area Under the Risk-Coverage Curve (AURC)~\citep{Geifman2019AURC}, where a lower coverage value indicates a better trade-off between precise prediction and uncertain ones retained by LLM UQ methods.

\begin{table}[th]
\caption{The comprehensive evaluation framework for LLM UQ methods proposed in this work, where $\checkmark$ indicates that the corresponding metric is adopted.}
\label{tab:uq_evaluation_framework}
\begin{tabular}{ccccccccc}
\toprule
 & \multirow{2}{*}{\textbf{\begin{tabular}[c]{@{}c@{}}Evaluation \\ Metrics\end{tabular}}} & \multicolumn{3}{c}{Ranking} & \multicolumn{3}{c}{Calibration} & Coverage \\
\cmidrule(lr){3-5} \cmidrule(lr){6-8} \cmidrule(lr){9-9}
 & & AUROC & AUPRC & AUARC & ECE & MCE & BS & AURC \\
\midrule
 & Answer-only & \checkmark & \checkmark & \checkmark & \checkmark & \checkmark & \checkmark & \checkmark \\
\cmidrule(lr){2-9}
\multirow{-3}{*}{\textbf{\begin{tabular}[c]{@{}c@{}}Generation \\ Settings\end{tabular}}} & \begin{tabular}[c]{@{}c@{}}Reasoning-\\ augmented\end{tabular} & \checkmark & \checkmark & \checkmark & \checkmark & \checkmark & \checkmark & \checkmark \\
\bottomrule
\end{tabular}
\end{table}

As LLM generation has shifted from answer-only generation, also known as input-output generation, to reasoning-augmented generation in recent years~\citep{wei2022CoT,yao2023ToT,chen2025CoT_survey}, uncertainty sources are extended from the final output alone to both the final output and the reasoning process that leads to it. Therefore, a comprehensive evaluation framework requires judging the performance of various LLMs in not only the answer-only generation where the LLM directly generates the final answer without explicitly generating intermediate reasoning steps, but also the reasoning-augmented generation that the LLM is prompted to first generate intermediate reasoning traces, such as explanations~\citep{Band2024Linguistic_Calibration,Xu2024SaySelf}, derivations~\citep{Da2025Topo-uq}, or chain-of-thought rationales~\citep{Zhang2025CoT-UQ,Ye2025CoTEntropy}, before generating the final answer. Table~\ref{tab:uq_evaluation_framework} summarizes the comprehensive evaluation framework for demonstrating the effectiveness of the conducted LLM UQ methods, which consists of two aspects, namely a metric aspect covering three evaluation dimensions with seven corresponding metrics and a generation-setting aspect including answer-only generation and reasoning-augmented generation. \textbf{The proposed evaluation framework answers Q1.}

Building upon the proposed evaluation framework for LLM UQ methods, we can empirically compare the performance of LLM UQ methods under the answer-only and reasoning-augmented settings and further answer Q2 by calculating the seven metrics across the three evaluation dimensions under two generation settings accordingly. The following subsections show the comparative experiments and empirical results.

\subsection{Performance Comparison of LLM UQ Methods in Answer-Only Generation}
\label{subsec:answer_only_uq}
This subsection empirically investigates the effects of UQ methods under the answer-only generation setting. Here, we conduct 16 UQ methods, including single-pass methods that contain MaxSeqProb~\citep{Vashurin2025LM-Polygraph}, MeanTokEnt~\citep{fomicheva2020PPL}, MaxTokEnt~\citep{fomicheva2020PPL}, PPL~\citep{fomicheva2020PPL}, Verbalized Probability~\citep{Lin2022Verbalized_Probability}, and $P(\text{True})$~\citep{Kadavath2022Ptrue/PIK}, as well as consensus-based methods that contain BSDetector~\citep{chen2024BSDETECTOR}, SelfCheckGPT~\citep{manakul2023selfcheckgpt}, NumSet~\citep{lin2024numset/eigv/deg/ecc}, EigV~\citep{lin2024numset/eigv/deg/ecc}, Deg~\citep{lin2024numset/eigv/deg/ecc}, Ecc~\citep{lin2024numset/eigv/deg/ecc}, ConU~\citep{wang2024ConU}, Semantic Entropy~\citep{farquhar2024SemanticEntropy}, Semantic Density~\citep{Qiu2024SemanticDensity}, and MARS~\citep{Bakman2024MARS}.

Tables~\ref{qwen3_1_7b_answer_only_results}, \ref{qwen3_4b_answer_only_results}, \ref{llama-3.2-1B_answer_only_results}, \ref{llama-3.2-3B_answer_only_results}, and \ref{deepseek-v3_answer_only_results} list the performance of 16 LLM UQ methods under the answer-only generation setting on Qwen3-1.7B, Qwen3-4B, Llama-3.2-1B, Llama-3.2-3B, and DeepSeek-V3, respectively. There are four primary observations. First, almost all UQ methods yield lower AUPRC and AUARC scores on GSM8K than those on other datasets for both the Qwen3 and Llama 3.2 families, implying that it is more challenging to quantify the uncertainty of LLMs in such mathematical reasoning tasks with the answer-only generation setting. Second, the average performance of all UQ methods for DeepSeek-V3 surpasses those for the Qwen3 and Llama 3.2 families on TriviaQA and GSM8K but falls short on HumanEval, revealing that black-box LLMs may gain a stronger ability to distinguish correct from incorrect generations on QA and mathematical reasoning tasks than the white-box LLMs. Third, many UQ methods yield better performance on larger-scale LLMs than on smaller ones across both the Qwen3 and Llama 3.2 families, implying that the LLM with greater model complexity may produce more precise answers with lower uncertainty scores. We conduct an in-depth investigation on Subsection~\ref{subsec:scaling_law}. Fourth, for all LLMs, Verbalized Probability, $P(\text{True})$, and BSDetector show lower variance in AUROC, AUPRC, and AUARC across datasets, indicating that these methods tend to provide more credible UQ performance. It is noted that these three methods rely on self-evaluation. Hence, we conjecture that the additional judgment induced by self-evaluation may provide more robust uncertainty estimates rather than probability-based ones in answer-only generation scenarios. Besides, these three methods perform better on the Qwen3 family than those on the Llama 3.2 family by 16.21\%, 11.57\%, and 4.33\%, respectively. Within each family, larger-scale LLMs also outperform smaller-scale LLMs by 10.29\%, 9.94\%, and 6.60\%, respectively. This observation suggests that the effectiveness of self-evaluation-based methods depends heavily on the LLM's ability to reflect on its outputs and estimate its confidence through self-prompting. Typically, the Qwen3 family exhibits stronger such abilities than the Llama 3.2 family, and these abilities improve with larger model scales within each family.

\begin{figure*}[!htb]
    \centering
    \includegraphics[width=\textwidth]{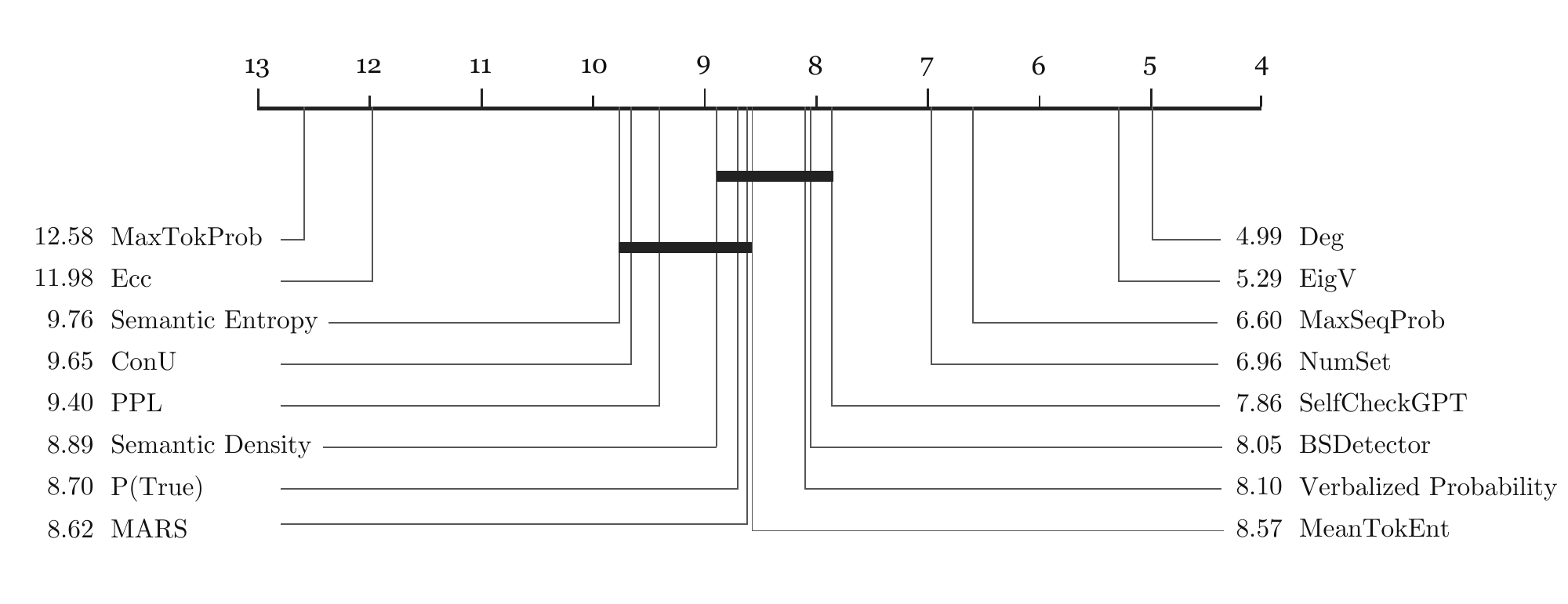}
    \caption{Overall evaluation ranking of the investigated LLM UQ methods in the answer-only generation setting.}
    \label{fig:answer_only_ranking}
\end{figure*}

Further, Figure~\ref{fig:answer_only_ranking} visualizes the overall evaluation ranks of the investigated LLM UQ methods in the answer-only generation setting, where the ranking scores are based on the average performance over 7 evaluation metrics across the Qwen3 and Llama 3.2 families. A lower ranking score indicates better overall performance of a certain UQ method. There are three primary observations. First, Deg and EigV achieve the best ranks among all evaluated methods. Second, MaxSeqProb achieves the second-best rank among all evaluated methods and the best average rank within the single-pass category. Third, UQ methods that explicitly elicit an additional judgment from the model, such as Verbalized Probability, $P(\text{True})$, and BSDetector, achieve the third tier of ranks in UQ performance among all evaluated methods. Figure~\ref{fig:answer_only_ranking_combined} in Appendix~\ref{app:ranking_answer_only} further visualizes the overall ranking plots for the ranking, calibration, and coverage evaluations in the answer-only setting. 

\begin{table*}[!htb]
    \centering
    \caption{Performance Comparison of the investigated LLM UQ methods of Qwen3-1.7B in answer-only generation across three evaluated benchmarks.}
    \label{qwen3_1_7b_answer_only_results}
    \resizebox{0.9\textwidth}{!}{
    \begin{tabular}{llccccccc}
        \toprule
        \multirow{2}{*}{\textbf{Datasets}}
        & \multirow{2}{*}{\textbf{Methods}}
        & \multicolumn{3}{c}{\textbf{Ranking}}
        & \multicolumn{3}{c}{\textbf{Calibration}}
        & \textbf{Coverage} \\
        \cmidrule(lr){3-5} \cmidrule(lr){6-8} \cmidrule(lr){9-9}
        & & \textbf{AUROC} & \textbf{AUPRC} & \textbf{AUARC}
        & \textbf{ECE} & \textbf{MCE} & \textbf{BS}
        & \textbf{AURC} \\
        \midrule
        \multirow{16}{*}{\textbf{TriviaQA}}
        & MaxSeqProb~\citep{Vashurin2025LM-Polygraph}
        & 67.62$_{\pm 0.65}$ & 29.90$_{\pm 1.26}$ & 22.51$_{\pm 0.82}$
        & 52.38$_{\pm 0.47}$ & 66.22$_{\pm 1.54}$ & 36.90$_{\pm 0.37}$
        & 77.48$_{\pm 0.82}$ \\

        & MeanTokEnt~\citep{fomicheva2020PPL}
        & 69.44$_{\pm 0.75}$ & 33.69$_{\pm 1.15}$ & 29.05$_{\pm 0.79}$
        & 56.67$_{\pm 0.46}$ & 66.00$_{\pm 0.89}$ & 47.80$_{\pm 0.36}$
        & 70.94$_{\pm 0.79}$ \\

        & MaxTokProb~\citep{fomicheva2020PPL}
        & 47.56$_{\pm 0.76}$ & 16.61$_{\pm 0.48}$ & 15.81$_{\pm 0.55}$
        & 74.43$_{\pm 0.49}$ & 79.54$_{\pm 0.50}$ & 72.61$_{\pm 0.46}$
        & 84.18$_{\pm 0.55}$ \\

        & PPL~\citep{fomicheva2020PPL}
        & 71.00$_{\pm 0.75}$ & 35.06$_{\pm 1.17}$ & 29.82$_{\pm 0.79}$
        & 73.90$_{\pm 0.45}$ & 79.50$_{\pm 1.98}$ & 69.39$_{\pm 0.42}$
        & 70.17$_{\pm 0.79}$ \\
        
        & BSDetector~\citep{chen2024BSDETECTOR}
        & 72.31$_{\pm 0.61}$ & 29.01$_{\pm 0.77}$ & 26.09$_{\pm 0.67}$
        & 57.36$_{\pm 0.47}$ & 66.02$_{\pm 1.19}$ & 47.99$_{\pm 0.42}$
        & 73.90$_{\pm 0.67}$ \\
        
        & SelfCheckGPT~\citep{manakul2023selfcheckgpt}
        & 64.96$_{\pm 0.64}$ & 23.17$_{\pm 0.61}$ & 21.43$_{\pm 0.57}$
        & 58.67$_{\pm 0.48}$ & 72.20$_{\pm 1.19}$ & 51.69$_{\pm 0.45}$
        & 78.56$_{\pm 0.57}$ \\
        
        & NumSet~\citep{lin2024numset/eigv/deg/ecc}
        & 70.10$_{\pm 0.56}$ & 28.75$_{\pm 0.72}$ & 26.96$_{\pm 0.73}$
        & 67.26$_{\pm 0.46}$ & 76.50$_{\pm 0.87}$ & 60.51$_{\pm 0.44}$
        & 73.03$_{\pm 0.73}$ \\
        
        & EigV~\citep{lin2024numset/eigv/deg/ecc}
        & 69.87$_{\pm 0.59}$ & 29.05$_{\pm 0.74}$ & 26.92$_{\pm 0.71}$
        & 50.25$_{\pm 0.51}$ & 67.41$_{\pm 0.96}$ & 43.96$_{\pm 0.47}$
        & 73.07$_{\pm 0.71}$ \\
        
        & Deg~\citep{lin2024numset/eigv/deg/ecc}
        & 69.88$_{\pm 0.59}$ & 28.99$_{\pm 0.74}$ & 26.92$_{\pm 0.73}$
        & 50.25$_{\pm 0.51}$ & 67.49$_{\pm 0.95}$ & 45.34$_{\pm 0.49}$
        & 73.07$_{\pm 0.73}$ \\
        
        & Ecc~\citep{lin2024numset/eigv/deg/ecc}
        & 36.59$_{\pm 0.43}$ & 17.80$_{\pm 0.45}$ & 11.30$_{\pm 0.41}$
        & 40.57$_{\pm 0.49}$ & 76.37$_{\pm 1.94}$ & 33.15$_{\pm 0.13}$
        & 88.68$_{\pm 0.41}$ \\
        
        & ConU~\citep{wang2024ConU}
        & 69.61$_{\pm 0.64}$ & 28.86$_{\pm 0.74}$ & 25.53$_{\pm 0.71}$
        & 65.09$_{\pm 0.46}$ & 76.16$_{\pm 1.08}$ & 58.88$_{\pm 0.45}$
        & 74.46$_{\pm 0.71}$ \\
        
        & Semantic Entropy~\citep{farquhar2024SemanticEntropy}
        & 69.92$_{\pm 0.62}$ & 28.93$_{\pm 0.74}$ & 25.58$_{\pm 0.71}$
        & 66.16$_{\pm 0.45}$ & 75.18$_{\pm 0.95}$ & 59.14$_{\pm 0.44}$
        & 74.41$_{\pm 0.71}$ \\
        
        & Semantic Density~\citep{Qiu2024SemanticDensity}
        & 69.73$_{\pm 0.63}$ & 28.88$_{\pm 0.74}$ & 25.54$_{\pm 0.71}$
        & 60.72$_{\pm 0.47}$ & 75.78$_{\pm 1.25}$ & 54.29$_{\pm 0.47}$
        & 74.45$_{\pm 0.71}$ \\
        
        & MARS~\citep{Bakman2024MARS}
        & 71.68$_{\pm 0.74}$ & 35.58$_{\pm 1.17}$ & 30.11$_{\pm 0.79}$
        & 65.86$_{\pm 0.45}$ & 72.17$_{\pm 0.89}$ & 58.32$_{\pm 0.39}$
        & 69.88$_{\pm 0.79}$ \\
        
        & Verbalized Probability~\citep{Lin2022Verbalized_Probability}
        & 57.65$_{\pm 0.63}$ & 29.93$_{\pm 0.64}$ & 30.68$_{\pm 0.71}$
        & 63.09$_{\pm 0.46}$ & 67.02$_{\pm 1.82}$ & 58.89$_{\pm 0.37}$
        & 69.31$_{\pm 0.71}$ \\
        
        & $P(\text{True})$~\citep{Kadavath2022Ptrue/PIK}
        & 64.96$_{\pm 0.76}$ & 31.11$_{\pm 1.11}$ & 27.50$_{\pm 0.79}$
        & 50.69$_{\pm 0.45}$ & 56.24$_{\pm 2.94}$ & 40.78$_{\pm 0.26}$
        & 72.49$_{\pm 0.79}$ \\
        \midrule
        \midrule
        \multirow{16}{*}{\textbf{GSM8K}}
        & MaxSeqProb~\citep{Vashurin2025LM-Polygraph}
        & 47.34$_{\pm 5.03}$ & 2.96$_{\pm 1.25}$ & 2.21$_{\pm 0.61}$
        & 56.19$_{\pm 0.84}$ & 94.42$_{\pm 0.98}$ & 41.26$_{\pm 0.86}$
        & 97.72$_{\pm 0.61}$ \\

        & MeanTokEnt~\citep{fomicheva2020PPL}
        & 37.55$_{\pm 5.22}$ & 2.27$_{\pm 0.75}$ & 1.71$_{\pm 0.51}$
        & 84.12$_{\pm 0.60}$ & 93.79$_{\pm 0.47}$ & 75.86$_{\pm 0.64}$
        & 98.21$_{\pm 0.51}$ \\

        & MaxTokProb~\citep{fomicheva2020PPL}
        & 37.25$_{\pm 4.53}$ & 1.89$_{\pm 0.35}$ & 1.74$_{\pm 0.49}$
        & 95.15$_{\pm 0.51}$ & 97.65$_{\pm 0.39}$ & 93.93$_{\pm 0.51}$
        & 98.19$_{\pm 0.49}$ \\

        & PPL~\citep{fomicheva2020PPL}
        & 38.67$_{\pm 5.50}$ & 2.34$_{\pm 0.78}$ & 1.79$_{\pm 0.53}$
        & 92.39$_{\pm 0.48}$ & 95.40$_{\pm 0.39}$ & 88.57$_{\pm 0.48}$
        & 98.13$_{\pm 0.53}$ \\

        & BSDetector~\citep{chen2024BSDETECTOR}
        & 34.68$_{\pm 3.26}$ & 3.49$_{\pm 0.74}$ & 2.54$_{\pm 0.56}$
        & 39.15$_{\pm 1.01}$ & 96.53$_{\pm 3.34}$ & 28.07$_{\pm 0.73}$
        & 97.38$_{\pm 0.56}$ \\

        & SelfCheckGPT~\citep{manakul2023selfcheckgpt}
        & 43.84$_{\pm 4.85}$ & 2.32$_{\pm 0.49}$ & 1.92$_{\pm 0.46}$
        & 85.19$_{\pm 0.62}$ & 96.04$_{\pm 1.50}$ & 78.76$_{\pm 0.71}$
        & 98.01$_{\pm 0.46}$ \\

        & NumSet~\citep{lin2024numset/eigv/deg/ecc}
        & 45.19$_{\pm 4.10}$ & 2.19$_{\pm 0.40}$ & 2.11$_{\pm 0.52}$
        & 93.29$_{\pm 0.48}$ & 97.90$_{\pm 0.43}$ & 90.39$_{\pm 0.51}$
        & 97.81$_{\pm 0.52}$ \\

        & EigV~\citep{lin2024numset/eigv/deg/ecc}
        & 45.71$_{\pm 3.98}$ & 2.20$_{\pm 0.41}$ & 2.12$_{\pm 0.51}$
        & 83.32$_{\pm 0.69}$ & 94.12$_{\pm 0.67}$ & 76.90$_{\pm 0.77}$
        & 97.80$_{\pm 0.51}$ \\

        & Deg~\citep{lin2024numset/eigv/deg/ecc}
        & 45.65$_{\pm 3.99}$ & 2.19$_{\pm 0.40}$ & 2.12$_{\pm 0.51}$
        & 83.32$_{\pm 0.69}$ & 95.14$_{\pm 0.60}$ & 79.76$_{\pm 0.81}$
        & 97.80$_{\pm 0.51}$ \\

        & Ecc~\citep{lin2024numset/eigv/deg/ecc}
        & 54.02$_{\pm 3.15}$ & 2.91$_{\pm 0.72}$ & 2.74$_{\pm 0.65}$
        & 42.64$_{\pm 0.47}$ & 99.85$_{\pm 2.12}$ & 2.85$_{\pm 0.44}$
        & 97.19$_{\pm 0.65}$ \\

        & ConU~\citep{wang2024ConU}
        & 46.24$_{\pm 4.37}$ & 2.45$_{\pm 0.58}$ & 2.13$_{\pm 0.49}$
        & 90.94$_{\pm 0.54}$ & 97.27$_{\pm 0.43}$ & 87.34$_{\pm 0.64}$
        & 97.79$_{\pm 0.49}$ \\

        & Semantic Entropy~\citep{farquhar2024SemanticEntropy}
        & 45.87$_{\pm 4.45}$ & 2.44$_{\pm 0.58}$ & 2.13$_{\pm 0.49}$
        & 91.59$_{\pm 0.51}$ & 97.40$_{\pm 0.41}$ & 87.91$_{\pm 0.58}$
        & 97.80$_{\pm 0.49}$ \\

        & Semantic Density~\citep{Qiu2024SemanticDensity}
        & 46.13$_{\pm 4.39}$ & 2.45$_{\pm 0.58}$ & 2.13$_{\pm 0.49}$
        & 88.50$_{\pm 0.61}$ & 97.62$_{\pm 0.41}$ & 84.25$_{\pm 0.71}$
        & 97.80$_{\pm 0.49}$ \\

        & MARS~\citep{Bakman2024MARS}
        & 39.37$_{\pm 5.31}$ & 2.33$_{\pm 0.75}$ & 1.80$_{\pm 0.52}$
        & 88.67$_{\pm 0.52}$ & 94.70$_{\pm 0.41}$ & 82.55$_{\pm 0.52}$
        & 98.12$_{\pm 0.52}$ \\

        & Verbalized Probability~\citep{Lin2022Verbalized_Probability}
        & 54.71$_{\pm 1.85}$ & 17.79$_{\pm 1.21}$ & 12.79$_{\pm 1.15}$
        & 54.20$_{\pm 0.73}$ & 90.27$_{\pm 3.25}$ & 37.15$_{\pm 0.75}$
        & 87.13$_{\pm 1.15}$ \\

        & $P(\text{True})$~\citep{Kadavath2022Ptrue/PIK}
        & 37.48$_{\pm 2.63}$ & 14.18$_{\pm 1.73}$ & 12.58$_{\pm 1.45}$
        & 57.83$_{\pm 1.05}$ & 90.86$_{\pm 11.46}$ & 47.44$_{\pm 0.55}$
        & 87.34$_{\pm 1.45}$ \\
        \midrule
        \midrule
        \multirow{16}{*}{\textbf{HumanEval}}
        & MaxSeqProb~\citep{Vashurin2025LM-Polygraph}
        & 56.10$_{\pm 4.46}$ & 62.60$_{\pm 5.46}$ & 60.33$_{\pm 5.40}$
        & 31.83$_{\pm 3.79}$ & 56.53$_{\pm 9.93}$ & 32.83$_{\pm 2.44}$
        & 39.06$_{\pm 5.40}$ \\

        & MeanTokEnt~\citep{fomicheva2020PPL}
        & 55.21$_{\pm 4.40}$ & 59.28$_{\pm 5.39}$ & 66.36$_{\pm 5.11}$
        & 20.52$_{\pm 3.22}$ & 46.78$_{\pm 17.80}$ & 26.03$_{\pm 2.31}$
        & 33.03$_{\pm 5.11}$ \\

        & MaxTokProb~\citep{fomicheva2020PPL}
        & 50.69$_{\pm 0.67}$ & 57.54$_{\pm 3.89}$ & 57.22$_{\pm 5.31}$
        & 37.33$_{\pm 3.66}$ & 45.26$_{\pm 2.43}$ & 35.07$_{\pm 2.64}$
        & 42.17$_{\pm 5.31}$ \\

        & PPL~\citep{fomicheva2020PPL}
        & 63.82$_{\pm 4.44}$ & 68.01$_{\pm 5.40}$ & 65.27$_{\pm 5.16}$
        & 27.88$_{\pm 3.62}$ & 50.54$_{\pm 6.90}$ & 30.10$_{\pm 2.68}$
        & 34.12$_{\pm 5.16}$ \\

        & BSDetector~\citep{chen2024BSDETECTOR}
        & 62.69$_{\pm 4.43}$ & 69.04$_{\pm 5.19}$ & 66.20$_{\pm 5.00}$
        & 18.87$_{\pm 3.40}$ & 57.23$_{\pm 18.45}$ & 26.00$_{\pm 2.18}$
        & 33.19$_{\pm 5.00}$ \\

        & SelfCheckGPT~\citep{manakul2023selfcheckgpt}
        & 51.30$_{\pm 4.28}$ & 61.42$_{\pm 5.07}$ & 59.23$_{\pm 5.04}$
        & 31.92$_{\pm 3.86}$ & 80.90$_{\pm 23.95}$ & 35.40$_{\pm 3.03}$
        & 40.16$_{\pm 5.04}$ \\

        & NumSet~\citep{lin2024numset/eigv/deg/ecc}
        & 55.05$_{\pm 4.24}$ & 60.25$_{\pm 4.48}$ & 59.71$_{\pm 5.09}$
        & 31.27$_{\pm 3.71}$ & 46.72$_{\pm 8.99}$ & 33.56$_{\pm 3.08}$
        & 39.68$_{\pm 5.09}$ \\

        & EigV~\citep{lin2024numset/eigv/deg/ecc}
        & 54.63$_{\pm 4.25}$ & 60.30$_{\pm 4.62}$ & 59.79$_{\pm 5.18}$
        & 30.94$_{\pm 3.61}$ & 51.60$_{\pm 8.82}$ & 33.95$_{\pm 2.71}$
        & 39.60$_{\pm 5.18}$ \\

        & Deg~\citep{lin2024numset/eigv/deg/ecc}
        & 54.61$_{\pm 4.25}$ & 60.25$_{\pm 4.59}$ & 59.77$_{\pm 5.16}$
        & 30.94$_{\pm 3.61}$ & 51.60$_{\pm 8.82}$ & 33.95$_{\pm 2.71}$
        & 39.62$_{\pm 5.16}$ \\

        & Ecc~\citep{lin2024numset/eigv/deg/ecc}
        & 45.66$_{\pm 3.59}$ & 54.68$_{\pm 4.11}$ & 50.15$_{\pm 5.07}$
        & 54.98$_{\pm 3.77}$ & 88.93$_{\pm 16.91}$ & 54.41$_{\pm 3.54}$
        & 49.24$_{\pm 5.07}$ \\

        & ConU~\citep{wang2024ConU}
        & 53.89$_{\pm 4.35}$ & 60.03$_{\pm 4.63}$ & 59.41$_{\pm 5.10}$
        & 32.18$_{\pm 3.72}$ & 56.23$_{\pm 15.15}$ & 34.34$_{\pm 2.97}$
        & 39.98$_{\pm 5.10}$ \\

        & Semantic Entropy~\citep{farquhar2024SemanticEntropy}
        & 53.12$_{\pm 4.42}$ & 59.75$_{\pm 4.65}$ & 59.20$_{\pm 5.18}$
        & 37.12$_{\pm 3.71}$ & 62.68$_{\pm 9.97}$ & 37.34$_{\pm 3.08}$
        & 40.19$_{\pm 5.18}$ \\

        & Semantic Density~\citep{Qiu2024SemanticDensity}
        & 53.81$_{\pm 4.35}$ & 60.00$_{\pm 4.63}$ & 59.41$_{\pm 5.18}$
        & 31.11$_{\pm 3.60}$ & 63.97$_{\pm 15.83}$ & 34.55$_{\pm 2.88}$
        & 39.98$_{\pm 5.18}$ \\

        & MARS~\citep{Bakman2024MARS}
        & 55.40$_{\pm 4.48}$ & 64.10$_{\pm 5.38}$ & 61.78$_{\pm 5.26}$
        & 32.42$_{\pm 3.70}$ & 74.72$_{\pm 15.58}$ & 34.64$_{\pm 3.13}$
        & 37.61$_{\pm 5.26}$ \\

        & Verbalized Probability~\citep{Lin2022Verbalized_Probability}
        & 65.99$_{\pm 4.40}$ & 67.71$_{\pm 4.88}$ & 66.69$_{\pm 5.02}$
        & 24.29$_{\pm 3.69}$ & 53.59$_{\pm 12.57}$ & 28.35$_{\pm 2.70}$
        & 32.70$_{\pm 5.02}$ \\

        & $P(\text{True})$~\citep{Kadavath2022Ptrue/PIK}
        & 70.30$_{\pm 4.31}$ & 75.04$_{\pm 4.81}$ & 71.07$_{\pm 4.53}$
        & 32.91$_{\pm 3.50}$ & 61.16$_{\pm 10.41}$ & 32.23$_{\pm 3.36}$
        & 28.32$_{\pm 4.53}$ \\
        \bottomrule
    \end{tabular}
    }
\end{table*}

\clearpage
\begin{table*}[t]
    \centering
    \caption{Performance Comparison of the investigated LLM UQ methods of Qwen3-4B in answer-only generation across three evaluated benchmarks.}
    \label{qwen3_4b_answer_only_results}
    \resizebox{0.9\textwidth}{!}{
    \begin{tabular}{llccccccc}
        \toprule
        \multirow{2}{*}{\textbf{Datasets}}
        & \multirow{2}{*}{\textbf{Methods}}
        & \multicolumn{3}{c}{\textbf{Ranking}}
        & \multicolumn{3}{c}{\textbf{Calibration}}
        & \textbf{Coverage} \\
        \cmidrule(lr){3-5} \cmidrule(lr){6-8} \cmidrule(lr){9-9}
        & & \textbf{AUROC} & \textbf{AUPRC} & \textbf{AUARC}
        & \textbf{ECE} & \textbf{MCE} & \textbf{BS}
        & \textbf{AURC} \\
        \midrule
        \multirow{16}{*}{\textbf{TriviaQA}}
        & MaxSeqProb~\citep{Vashurin2025LM-Polygraph}
        & 84.66$_{\pm 0.45}$ & 72.76$_{\pm 0.99}$ & 59.91$_{\pm 0.74}$
        & 21.71$_{\pm 0.45}$ & 35.52$_{\pm 1.87}$ & 20.61$_{\pm 0.32}$
        & 40.08$_{\pm 0.74}$ \\

        & MeanTokEnt~\citep{fomicheva2020PPL}
        & 82.50$_{\pm 0.47}$ & 70.82$_{\pm 1.02}$ & 58.82$_{\pm 0.75}$
        & 42.76$_{\pm 0.48}$ & 58.01$_{\pm 1.16}$ & 35.97$_{\pm 0.37}$
        & 41.17$_{\pm 0.75}$ \\

        & MaxTokProb~\citep{fomicheva2020PPL}
        & 59.01$_{\pm 0.62}$ & 37.94$_{\pm 0.75}$ & 37.52$_{\pm 0.79}$
        & 60.27$_{\pm 0.54}$ & 69.74$_{\pm 2.06}$ & 58.63$_{\pm 0.52}$
        & 62.47$_{\pm 0.79}$ \\

        & PPL~\citep{fomicheva2020PPL}
        & 82.13$_{\pm 0.47}$ & 70.52$_{\pm 1.01}$ & 58.64$_{\pm 0.74}$
        & 60.25$_{\pm 0.52}$ & 81.46$_{\pm 0.71}$ & 57.12$_{\pm 0.49}$
        & 41.35$_{\pm 0.74}$ \\
        
        & BSDetector~\citep{chen2024BSDETECTOR}
        & 75.98$_{\pm 0.51}$ & 51.37$_{\pm 0.88}$ & 47.55$_{\pm 0.78}$
        & 45.54$_{\pm 0.48}$ & 58.84$_{\pm 1.30}$ & 39.26$_{\pm 0.42}$
        & 52.44$_{\pm 0.78}$ \\

        & SelfCheckGPT~\citep{manakul2023selfcheckgpt}
        & 73.58$_{\pm 0.55}$ & 47.35$_{\pm 0.82}$ & 44.31$_{\pm 0.76}$
        & 44.38$_{\pm 0.49}$ & 65.90$_{\pm 1.66}$ & 38.83$_{\pm 0.44}$
        & 55.68$_{\pm 0.76}$ \\

        & NumSet~\citep{lin2024numset/eigv/deg/ecc}
        & 75.70$_{\pm 0.46}$ & 53.17$_{\pm 0.75}$ & 49.74$_{\pm 0.76}$
        & 44.45$_{\pm 0.49}$ & 84.00$_{\pm 18.07}$ & 38.98$_{\pm 0.37}$
        & 50.25$_{\pm 0.76}$ \\

        & EigV~\citep{lin2024numset/eigv/deg/ecc}
        & 75.74$_{\pm 0.46}$ & 53.69$_{\pm 0.75}$ & 49.87$_{\pm 0.77}$
        & 37.83$_{\pm 0.49}$ & 44.93$_{\pm 1.18}$ & 33.35$_{\pm 0.45}$
        & 50.12$_{\pm 0.77}$ \\

        & Deg~\citep{lin2024numset/eigv/deg/ecc}
        & 75.74$_{\pm 0.46}$ & 53.63$_{\pm 0.75}$ & 49.87$_{\pm 0.76}$
        & 37.83$_{\pm 0.49}$ & 45.93$_{\pm 2.98}$ & 34.33$_{\pm 0.47}$
        & 50.11$_{\pm 0.76}$ \\

        & Ecc~\citep{lin2024numset/eigv/deg/ecc}
        & 33.76$_{\pm 0.42}$ & 33.72$_{\pm 0.54}$ & 20.70$_{\pm 0.50}$
        & 22.22$_{\pm 0.57}$ & 71.79$_{\pm 0.67}$ & 29.63$_{\pm 0.11}$
        & 79.28$_{\pm 0.50}$ \\

        & ConU~\citep{wang2024ConU}
        & 75.01$_{\pm 0.48}$ & 52.71$_{\pm 0.78}$ & 47.89$_{\pm 0.80}$
        & 51.77$_{\pm 0.49}$ & 71.56$_{\pm 1.40}$ & 46.44$_{\pm 0.46}$
        & 52.10$_{\pm 0.80}$ \\

        & Semantic Entropy~\citep{farquhar2024SemanticEntropy}
        & 75.26$_{\pm 0.48}$ & 52.79$_{\pm 0.78}$ & 47.96$_{\pm 0.81}$
        & 51.23$_{\pm 0.48}$ & 70.94$_{\pm 1.54}$ & 45.30$_{\pm 0.45}$
        & 52.02$_{\pm 0.81}$ \\

        & Semantic Density~\citep{Qiu2024SemanticDensity}
        & 75.11$_{\pm 0.48}$ & 52.73$_{\pm 0.78}$ & 47.92$_{\pm 0.80}$
        & 46.89$_{\pm 0.48}$ & 72.66$_{\pm 1.59}$ & 41.66$_{\pm 0.46}$
        & 52.07$_{\pm 0.80}$ \\

        & MARS~\citep{Bakman2024MARS}
        & 82.43$_{\pm 0.47}$ & 71.07$_{\pm 1.02}$ & 58.90$_{\pm 0.75}$
        & 53.34$_{\pm 0.50}$ & 70.63$_{\pm 0.96}$ & 47.59$_{\pm 0.45}$
        & 41.09$_{\pm 0.75}$ \\

        & Verbalized Probability~\citep{Lin2022Verbalized_Probability}
        & 73.14$_{\pm 0.51}$ & 60.77$_{\pm 0.77}$ & 59.64$_{\pm 0.77}$
        & 20.05$_{\pm 0.53}$ & 58.26$_{\pm 10.83}$ & 24.55$_{\pm 0.20}$
        & 40.35$_{\pm 0.77}$ \\

        & $P(\text{True})$~\citep{Kadavath2022Ptrue/PIK}
        & 85.82$_{\pm 0.40}$ & 74.23$_{\pm 0.96}$ & 60.66$_{\pm 0.76}$
        & 53.47$_{\pm 0.50}$ & 80.04$_{\pm 1.07}$ & 48.47$_{\pm 0.49}$
        & 39.32$_{\pm 0.76}$ \\
        \midrule
        \midrule
        \multirow{16}{*}{\textbf{GSM8K}}
        & MaxSeqProb~\citep{Vashurin2025LM-Polygraph}
        & 60.99$_{\pm 3.83}$ & 8.04$_{\pm 1.84}$ & 6.14$_{\pm 1.02}$
        & 49.09$_{\pm 1.02}$ & 89.43$_{\pm 1.64}$ & 36.05$_{\pm 0.96}$
        & 93.78$_{\pm 1.02}$ \\

        & MeanTokEnt~\citep{fomicheva2020PPL}
        & 43.30$_{\pm 4.20}$ & 5.89$_{\pm 1.69}$ & 4.39$_{\pm 0.88}$
        & 83.38$_{\pm 0.74}$ & 92.73$_{\pm 0.60}$ & 76.62$_{\pm 0.74}$
        & 95.53$_{\pm 0.88}$ \\

        & MaxTokProb~\citep{fomicheva2020PPL}
        & 36.71$_{\pm 3.33}$ & 3.33$_{\pm 0.50}$ & 3.01$_{\pm 0.62}$
        & 91.81$_{\pm 0.67}$ & 95.84$_{\pm 0.59}$ & 90.35$_{\pm 0.70}$
        & 96.91$_{\pm 0.62}$ \\

        & PPL~\citep{fomicheva2020PPL}
        & 45.44$_{\pm 4.34}$ & 6.16$_{\pm 1.71}$ & 4.65$_{\pm 0.90}$
        & 91.77$_{\pm 0.61}$ & 94.40$_{\pm 0.58}$ & 88.97$_{\pm 0.62}$
        & 95.28$_{\pm 0.90}$ \\
        
        & BSDetector~\citep{chen2024BSDETECTOR}
        & 37.16$_{\pm 3.39}$ & 3.37$_{\pm 0.50}$ & 2.74$_{\pm 0.51}$
        & 44.27$_{\pm 0.95}$ & 99.56$_{\pm 0.33}$ & 30.06$_{\pm 0.71}$
        & 97.18$_{\pm 0.51}$ \\

        & SelfCheckGPT~\citep{manakul2023selfcheckgpt}
        & 54.52$_{\pm 4.07}$ & 5.72$_{\pm 1.14}$ & 5.04$_{\pm 0.88}$
        & 77.96$_{\pm 0.84}$ & 93.20$_{\pm 0.72}$ & 70.19$_{\pm 0.94}$
        & 94.88$_{\pm 0.88}$ \\

        & NumSet~\citep{lin2024numset/eigv/deg/ecc}
        & 51.59$_{\pm 3.23}$ & 4.51$_{\pm 0.66}$ & 4.42$_{\pm 0.81}$
        & 85.68$_{\pm 0.79}$ & 95.55$_{\pm 0.72}$ & 80.77$_{\pm 0.86}$
        & 95.50$_{\pm 0.81}$ \\

        & EigV~\citep{lin2024numset/eigv/deg/ecc}
        & 51.95$_{\pm 3.26}$ & 4.54$_{\pm 0.67}$ & 4.46$_{\pm 0.81}$
        & 72.52$_{\pm 1.01}$ & 91.31$_{\pm 1.69}$ & 65.03$_{\pm 1.10}$
        & 95.46$_{\pm 0.81}$ \\

        & Deg~\citep{lin2024numset/eigv/deg/ecc}
        & 51.79$_{\pm 3.28}$ & 4.53$_{\pm 0.67}$ & 4.45$_{\pm 0.82}$
        & 72.52$_{\pm 1.01}$ & 91.69$_{\pm 1.69}$ & 67.32$_{\pm 1.15}$
        & 95.48$_{\pm 0.82}$ \\

        & Ecc~\citep{lin2024numset/eigv/deg/ecc}
        & 47.32$_{\pm 2.15}$ & 4.20$_{\pm 0.54}$ & 3.66$_{\pm 0.59}$
        & 7.68$_{\pm 0.72}$ & 51.45$_{\pm 6.58}$ & 7.17$_{\pm 0.67}$
        & 96.26$_{\pm 0.59}$ \\

        & ConU~\citep{wang2024ConU}
        & 51.26$_{\pm 3.39}$ & 4.77$_{\pm 0.73}$ & 4.00$_{\pm 0.68}$
        & 83.70$_{\pm 0.82}$ & 94.86$_{\pm 0.68}$ & 78.60$_{\pm 0.95}$
        & 95.93$_{\pm 0.68}$ \\

        & Semantic Entropy~\citep{farquhar2024SemanticEntropy}
        & 51.14$_{\pm 3.41}$ & 4.76$_{\pm 0.73}$ & 3.98$_{\pm 0.65}$
        & 84.11$_{\pm 0.81}$ & 94.85$_{\pm 0.71}$ & 78.69$_{\pm 0.92}$
        & 95.95$_{\pm 0.65}$ \\

        & Semantic Density~\citep{Qiu2024SemanticDensity}
        & 51.10$_{\pm 3.42}$ & 4.76$_{\pm 0.73}$ & 3.98$_{\pm 0.67}$
        & 79.78$_{\pm 0.93}$ & 95.03$_{\pm 0.72}$ & 74.14$_{\pm 1.06}$
        & 95.94$_{\pm 0.67}$ \\

        & MARS~\citep{Bakman2024MARS}
        & 46.44$_{\pm 4.17}$ & 5.98$_{\pm 1.58}$ & 4.65$_{\pm 0.89}$
        & 87.25$_{\pm 0.67}$ & 92.78$_{\pm 0.66}$ & 81.85$_{\pm 0.69}$
        & 95.27$_{\pm 0.89}$ \\

        & Verbalized Probability~\citep{Lin2022Verbalized_Probability}
        & 59.27$_{\pm 1.97}$ & 32.47$_{\pm 2.25}$ & 31.00$_{\pm 2.05}$
        & 39.88$_{\pm 1.26}$ & 69.45$_{\pm 2.44}$ & 35.35$_{\pm 0.97}$
        & 68.93$_{\pm 2.05}$ \\

        & $P(\text{True})$~\citep{Kadavath2022Ptrue/PIK}
        & 64.04$_{\pm 2.15}$ & 51.18$_{\pm 2.80}$ & 38.43$_{\pm 1.88}$
        & 44.65$_{\pm 1.02}$ & 83.74$_{\pm 20.65}$ & 36.20$_{\pm 0.52}$
        & 61.50$_{\pm 1.88}$ \\
        \midrule
        \midrule
        \multirow{16}{*}{\textbf{HumanEval}}
        & MaxSeqProb~\citep{Vashurin2025LM-Polygraph}
        & 47.53$_{\pm 4.88}$ & 66.07$_{\pm 5.11}$ & 64.04$_{\pm 5.47}$
        & 28.86$_{\pm 3.54}$ & 65.84$_{\pm 11.47}$ & 31.59$_{\pm 2.49}$
        & 35.35$_{\pm 5.47}$ \\

        & MeanTokEnt~\citep{fomicheva2020PPL}
        & 59.60$_{\pm 5.23}$ & 71.92$_{\pm 5.09}$ & 69.92$_{\pm 5.26}$
        & 20.24$_{\pm 3.50}$ & 56.01$_{\pm 18.61}$ & 24.91$_{\pm 2.78}$
        & 29.47$_{\pm 5.26}$ \\

        & MaxTokProb~\citep{fomicheva2020PPL}
        & 50.00$_{\pm 0.00}$ & 67.04$_{\pm 3.60}$ & 66.85$_{\pm 5.15}$
        & 17.04$_{\pm 3.60}$ & 17.04$_{\pm 3.60}$ & 25.00$_{\pm 0.00}$
        & 32.54$_{\pm 5.15}$ \\

        & PPL~\citep{fomicheva2020PPL}
        & 55.15$_{\pm 5.28}$ & 69.68$_{\pm 5.19}$ & 67.76$_{\pm 5.42}$
        & 24.21$_{\pm 3.56}$ & 50.27$_{\pm 8.61}$ & 27.07$_{\pm 3.07}$
        & 31.63$_{\pm 5.42}$ \\

        & BSDetector~\citep{chen2024BSDETECTOR}
        & 75.71$_{\pm 4.10}$ & 86.91$_{\pm 2.92}$ & 83.20$_{\pm 2.97}$
        & 18.40$_{\pm 2.91}$ & 67.64$_{\pm 21.34}$ & 20.12$_{\pm 2.14}$
        & 16.19$_{\pm 2.97}$ \\

        & SelfCheckGPT~\citep{manakul2023selfcheckgpt}
        & 45.75$_{\pm 4.85}$ & 65.60$_{\pm 4.98}$ & 63.58$_{\pm 5.32}$
        & 30.62$_{\pm 3.66}$ & 86.99$_{\pm 9.27}$ & 31.77$_{\pm 3.26}$
        & 35.81$_{\pm 5.32}$ \\

        & NumSet~\citep{lin2024numset/eigv/deg/ecc}
        & 48.24$_{\pm 4.12}$ & 66.17$_{\pm 4.42}$ & 65.24$_{\pm 5.53}$
        & 27.35$_{\pm 3.80}$ & 41.13$_{\pm 8.13}$ & 29.75$_{\pm 3.37}$
        & 34.15$_{\pm 5.53}$ \\

        & EigV~\citep{lin2024numset/eigv/deg/ecc}
        & 48.33$_{\pm 4.15}$ & 66.09$_{\pm 4.52}$ & 65.25$_{\pm 5.48}$
        & 34.80$_{\pm 3.64}$ & 67.66$_{\pm 8.08}$ & 33.00$_{\pm 3.23}$
        & 34.14$_{\pm 5.48}$ \\

        & Deg~\citep{lin2024numset/eigv/deg/ecc}
        & 48.27$_{\pm 4.14}$ & 66.11$_{\pm 4.49}$ & 65.26$_{\pm 5.48}$
        & 34.80$_{\pm 3.64}$ & 67.66$_{\pm 8.08}$ & 33.00$_{\pm 3.23}$
        & 34.13$_{\pm 5.48}$ \\

        & Ecc~\citep{lin2024numset/eigv/deg/ecc}
        & 52.31$_{\pm 2.56}$ & 67.79$_{\pm 3.59}$ & 66.87$_{\pm 4.76}$
        & 65.78$_{\pm 3.55}$ & 93.18$_{\pm 10.77}$ & 64.77$_{\pm 3.51}$
        & 32.52$_{\pm 4.76}$ \\

        & ConU~\citep{wang2024ConU}
        & 47.67$_{\pm 4.04}$ & 65.92$_{\pm 4.48}$ & 65.02$_{\pm 5.52}$
        & 33.68$_{\pm 3.48}$ & 70.96$_{\pm 11.97}$ & 33.06$_{\pm 3.22}$
        & 34.37$_{\pm 5.52}$ \\

        & Semantic Entropy~\citep{farquhar2024SemanticEntropy}
        & 47.12$_{\pm 3.96}$ & 65.78$_{\pm 4.48}$ & 64.96$_{\pm 5.46}$
        & 39.09$_{\pm 3.67}$ & 81.91$_{\pm 7.26}$ & 37.48$_{\pm 3.38}$
        & 34.43$_{\pm 5.46}$ \\

        & Semantic Density~\citep{Qiu2024SemanticDensity}
        & 47.67$_{\pm 4.06}$ & 65.92$_{\pm 4.48}$ & 65.08$_{\pm 5.49}$
        & 35.30$_{\pm 3.53}$ & 73.05$_{\pm 11.17}$ & 34.50$_{\pm 3.24}$
        & 34.31$_{\pm 5.49}$ \\

        & MARS~\citep{Bakman2024MARS}
        & 55.57$_{\pm 4.87}$ & 71.53$_{\pm 4.84}$ & 69.59$_{\pm 4.87}$
        & 27.24$_{\pm 3.63}$ & 77.14$_{\pm 15.88}$ & 29.00$_{\pm 3.07}$
        & 29.80$_{\pm 4.87}$ \\

        & Verbalized Probability~\citep{Lin2022Verbalized_Probability}
        & 65.81$_{\pm 4.83}$ & 75.41$_{\pm 3.80}$ & 71.06$_{\pm 5.01}$
        & 18.30$_{\pm 3.18}$ & 50.90$_{\pm 6.73}$ & 21.56$_{\pm 2.61}$
        & 28.33$_{\pm 5.01}$ \\

        & $P(\text{True})$~\citep{Kadavath2022Ptrue/PIK}
        & 63.64$_{\pm 4.83}$ & 69.72$_{\pm 4.44}$ & 66.21$_{\pm 4.89}$
        & 40.36$_{\pm 3.54}$ & 71.69$_{\pm 9.45}$ & 39.59$_{\pm 3.47}$
        & 33.18$_{\pm 4.89}$ \\
        \bottomrule
    \end{tabular}
    }
\end{table*}

\begin{table*}[t]
    \centering
    \caption{Performance Comparison of the investigated LLM UQ methods of Llama-3.2-1B in answer-only generation across three evaluated benchmarks.}
    \label{llama-3.2-1B_answer_only_results}
    \resizebox{0.9\textwidth}{!}{
    \begin{tabular}{llccccccc}
        \toprule
        \multirow{2}{*}{\textbf{Datasets}}
        & \multirow{2}{*}{\textbf{Methods}}
        & \multicolumn{3}{c}{\textbf{Ranking}}
        & \multicolumn{3}{c}{\textbf{Calibration}}
        & \textbf{Coverage} \\
        \cmidrule(lr){3-5} \cmidrule(lr){6-8} \cmidrule(lr){9-9}
        & & \textbf{AUROC} & \textbf{AUPRC} & \textbf{AUARC}
        & \textbf{ECE} & \textbf{MCE} & \textbf{BS}
        & \textbf{AURC} \\
        \midrule
        \multirow{16}{*}{\textbf{TriviaQA}}
        & MaxSeqProb~\citep{Vashurin2025LM-Polygraph}
        & 61.75$_{\pm 0.70}$ & 36.67$_{\pm 0.83}$ & 35.21$_{\pm 0.74}$
        & 28.98$_{\pm 0.51}$ & 59.22$_{\pm 4.81}$ & 29.02$_{\pm 0.50}$
        & 64.78$_{\pm 0.74}$ \\

        & MeanTokEnt~\citep{fomicheva2020PPL}
        & 58.13$_{\pm 0.63}$ & 32.34$_{\pm 0.70}$ & 31.37$_{\pm 0.68}$
        & 24.67$_{\pm 0.52}$ & 91.38$_{\pm 1.39}$ & 27.28$_{\pm 0.16}$
        & 68.62$_{\pm 0.68}$ \\

        & MaxTokProb~\citep{fomicheva2020PPL}
        & 58.00$_{\pm 0.67}$ & 32.59$_{\pm 0.77}$ & 32.25$_{\pm 0.74}$
        & 69.23$_{\pm 0.51}$ & 81.02$_{\pm 2.45}$ & 68.51$_{\pm 0.51}$
        & 67.73$_{\pm 0.74}$ \\

        & PPL~\citep{fomicheva2020PPL}
        & 62.78$_{\pm 0.67}$ & 37.64$_{\pm 0.85}$ & 35.95$_{\pm 0.74}$
        & 64.24$_{\pm 0.50}$ & 71.32$_{\pm 1.09}$ & 61.45$_{\pm 0.44}$
        & 64.04$_{\pm 0.74}$ \\
        
        & BSDetector~\citep{chen2024BSDETECTOR}
        & 45.00$_{\pm 0.64}$ & 33.29$_{\pm 0.65}$ & 32.90$_{\pm 0.71}$
        & 15.69$_{\pm 0.55}$ & 85.51$_{\pm 6.29}$ & 27.12$_{\pm 0.17}$
        & 67.09$_{\pm 0.71}$ \\

        & SelfCheckGPT~\citep{manakul2023selfcheckgpt}
        & 79.61$_{\pm 0.63}$ & 62.15$_{\pm 1.16}$ & 50.19$_{\pm 0.78}$
        & 29.29$_{\pm 0.48}$ & 42.17$_{\pm 1.28}$ & 25.13$_{\pm 0.31}$
        & 49.80$_{\pm 0.78}$ \\

        & NumSet~\citep{lin2024numset/eigv/deg/ecc}
        & 80.94$_{\pm 0.57}$ & 60.44$_{\pm 1.01}$ & 50.79$_{\pm 0.78}$
        & 36.17$_{\pm 0.47}$ & 47.71$_{\pm 1.04}$ & 29.44$_{\pm 0.34}$
        & 49.19$_{\pm 0.78}$ \\

        & EigV~\citep{lin2024numset/eigv/deg/ecc}
        & 81.47$_{\pm 0.59}$ & 63.69$_{\pm 1.04}$ & 51.29$_{\pm 0.76}$
        & 35.84$_{\pm 0.47}$ & 47.71$_{\pm 1.04}$ & 17.18$_{\pm 0.32}$
        & 48.70$_{\pm 0.76}$ \\

        & Deg~\citep{lin2024numset/eigv/deg/ecc}
        & 81.40$_{\pm 0.59}$ & 63.16$_{\pm 1.04}$ & 51.25$_{\pm 0.75}$
        & 21.46$_{\pm 0.45}$ & 27.54$_{\pm 2.21}$ & 17.18$_{\pm 0.32}$
        & 48.74$_{\pm 0.75}$ \\

        & Ecc~\citep{lin2024numset/eigv/deg/ecc}
        & 25.66$_{\pm 0.57}$ & 22.99$_{\pm 0.41}$ & 14.22$_{\pm 0.44}$
        & 47.85$_{\pm 0.53}$ & 91.93$_{\pm 0.77}$ & 46.51$_{\pm 0.30}$
        & 85.77$_{\pm 0.44}$ \\

        & ConU~\citep{wang2024ConU}
        & 75.60$_{\pm 0.64}$ & 57.51$_{\pm 1.11}$ & 46.48$_{\pm 0.76}$
        & 44.54$_{\pm 0.50}$ & 60.23$_{\pm 1.41}$ & 38.82$_{\pm 0.40}$
        & 53.51$_{\pm 0.76}$ \\

        & Semantic Entropy~\citep{farquhar2024SemanticEntropy}
        & 76.87$_{\pm 0.62}$ & 58.13$_{\pm 1.11}$ & 46.91$_{\pm 0.76}$
        & 45.33$_{\pm 0.48}$ & 57.95$_{\pm 1.49}$ & 37.98$_{\pm 0.37}$
        & 53.08$_{\pm 0.76}$ \\

        & Semantic Density~\citep{Qiu2024SemanticDensity}
        & 76.03$_{\pm 0.63}$ & 57.66$_{\pm 1.11}$ & 46.60$_{\pm 0.76}$
        & 37.25$_{\pm 0.50}$ & 60.81$_{\pm 1.74}$ & 32.87$_{\pm 0.38}$
        & 53.39$_{\pm 0.76}$ \\

        & MARS~\citep{Bakman2024MARS}
        & 62.56$_{\pm 0.70}$ & 36.90$_{\pm 0.83}$ & 35.66$_{\pm 0.74}$
        & 44.75$_{\pm 0.50}$ & 56.99$_{\pm 2.28}$ & 40.16$_{\pm 0.31}$
        & 64.33$_{\pm 0.74}$ \\

        & Verbalized Probability~\citep{Lin2022Verbalized_Probability}
        & 60.38$_{\pm 0.72}$ & 18.81$_{\pm 0.59}$ & 18.51$_{\pm 0.61}$
        & 57.33$_{\pm 0.49}$ & 80.61$_{\pm 0.66}$ & 51.54$_{\pm 0.48}$
        & 81.48$_{\pm 0.61}$ \\

        & $P(\text{True})$~\citep{Kadavath2022Ptrue/PIK}
        & 52.30$_{\pm 0.96}$ & 31.42$_{\pm 0.99}$ & 31.18$_{\pm 0.96}$
        & 38.57$_{\pm 0.68}$ & 85.23$_{\pm 19.23}$ & 36.37$_{\pm 0.30}$
        & 68.80$_{\pm 0.96}$ \\
        \midrule
        \midrule
        \multirow{16}{*}{\textbf{GSM8K}}
        & MaxSeqProb~\citep{Vashurin2025LM-Polygraph}
        & 62.10$_{\pm 4.16}$ & 7.59$_{\pm 2.55}$ & 4.82$_{\pm 0.90}$
        & 3.56$_{\pm 0.49}$ & 73.40$_{\pm 13.98}$ & 3.53$_{\pm 0.49}$
        & 95.11$_{\pm 0.90}$ \\

        & MeanTokEnt~\citep{fomicheva2020PPL}
        & 53.25$_{\pm 4.65}$ & 4.32$_{\pm 0.84}$ & 3.73$_{\pm 0.69}$
        & 44.27$_{\pm 0.76}$ & 90.95$_{\pm 2.24}$ & 27.63$_{\pm 0.65}$
        & 96.19$_{\pm 0.69}$ \\

        & MaxTokProb~\citep{fomicheva2020PPL}
        & 59.63$_{\pm 4.07}$ & 5.38$_{\pm 1.12}$ & 4.44$_{\pm 0.77}$
        & 91.91$_{\pm 0.54}$ & 94.03$_{\pm 0.55}$ & 88.37$_{\pm 0.57}$
        & 95.48$_{\pm 0.77}$ \\

        & PPL~\citep{fomicheva2020PPL}
        & 61.30$_{\pm 4.10}$ & 5.22$_{\pm 1.00}$ & 4.38$_{\pm 0.76}$
        & 74.83$_{\pm 0.57}$ & 86.73$_{\pm 1.29}$ & 60.82$_{\pm 0.57}$
        & 95.54$_{\pm 0.76}$ \\
        
        & BSDetector~\citep{chen2024BSDETECTOR}
        & 44.41$_{\pm 4.46}$ & 3.55$_{\pm 0.95}$ & 2.94$_{\pm 0.67}$
        & 33.93$_{\pm 0.70}$ & 94.68$_{\pm 2.32}$ & 18.13$_{\pm 0.44}$
        & 96.98$_{\pm 0.67}$ \\

        & SelfCheckGPT~\citep{manakul2023selfcheckgpt}
        & 58.75$_{\pm 4.04}$ & 5.15$_{\pm 1.09}$ & 4.30$_{\pm 0.77}$
        & 22.64$_{\pm 0.76}$ & 99.16$_{\pm 0.13}$ & 12.31$_{\pm 0.52}$
        & 95.62$_{\pm 0.77}$ \\

        & NumSet~\citep{lin2024numset/eigv/deg/ecc}
        & 62.99$_{\pm 4.02}$ & 5.77$_{\pm 1.21}$ & 4.92$_{\pm 0.85}$
        & 38.36$_{\pm 0.76}$ & 83.21$_{\pm 2.13}$ & 21.54$_{\pm 0.57}$
        & 95.01$_{\pm 0.85}$ \\

        & EigV~\citep{lin2024numset/eigv/deg/ecc}
        & 61.75$_{\pm 4.07}$ & 5.54$_{\pm 1.18}$ & 4.64$_{\pm 0.81}$
        & 38.09$_{\pm 0.78}$ & 75.32$_{\pm 4.12}$ & 6.98$_{\pm 0.46}$
        & 95.29$_{\pm 0.81}$ \\

        & Deg~\citep{lin2024numset/eigv/deg/ecc}
        & 61.65$_{\pm 4.06}$ & 5.51$_{\pm 1.19}$ & 4.64$_{\pm 0.82}$
        & 22.07$_{\pm 0.65}$ & 49.83$_{\pm 3.75}$ & 6.98$_{\pm 0.46}$
        & 95.28$_{\pm 0.82}$ \\

        & Ecc~\citep{lin2024numset/eigv/deg/ecc}
        & 36.31$_{\pm 3.96}$ & 2.97$_{\pm 0.59}$ & 2.47$_{\pm 0.59}$
        & 37.85$_{\pm 0.84}$ & 91.98$_{\pm 4.34}$ & 24.06$_{\pm 0.67}$
        & 97.46$_{\pm 0.59}$ \\

        & ConU~\citep{wang2024ConU}
        & 52.54$_{\pm 4.42}$ & 4.40$_{\pm 0.98}$ & 3.72$_{\pm 0.73}$
        & 55.00$_{\pm 0.93}$ & 92.41$_{\pm 1.23}$ & 40.76$_{\pm 0.90}$
        & 96.20$_{\pm 0.73}$ \\

        & Semantic Entropy~\citep{farquhar2024SemanticEntropy}
        & 54.00$_{\pm 4.09}$ & 4.46$_{\pm 0.98}$ & 3.78$_{\pm 0.73}$
        & 56.99$_{\pm 0.82}$ & 91.63$_{\pm 1.76}$ & 40.35$_{\pm 0.79}$
        & 96.14$_{\pm 0.73}$ \\

        & Semantic Density~\citep{Qiu2024SemanticDensity}
        & 53.50$_{\pm 4.24}$ & 4.44$_{\pm 0.98}$ & 3.76$_{\pm 0.73}$
        & 46.65$_{\pm 0.95}$ & 91.53$_{\pm 1.78}$ & 32.40$_{\pm 0.87}$
        & 96.17$_{\pm 0.73}$ \\

        & MARS~\citep{Bakman2024MARS}
        & 57.89$_{\pm 4.18}$ & 4.52$_{\pm 0.87}$ & 3.93$_{\pm 0.72}$
        & 63.25$_{\pm 0.67}$ & 91.75$_{\pm 2.23}$ & 45.19$_{\pm 0.60}$
        & 95.99$_{\pm 0.72}$ \\

        & Verbalized Probability~\citep{Lin2022Verbalized_Probability}
        & 47.35$_{\pm 3.88}$ & 3.23$_{\pm 0.55}$ & 2.94$_{\pm 0.61}$
        & 33.96$_{\pm 1.12}$ & 97.63$_{\pm 0.89}$ & 30.01$_{\pm 1.12}$
        & 96.98$_{\pm 0.61}$ \\

        & $P(\text{True})$~\citep{Kadavath2022Ptrue/PIK}
        & 48.10$_{\pm 4.80}$ & 3.38$_{\pm 1.15}$ & 2.67$_{\pm 0.66}$
        & 62.78$_{\pm 0.60}$ & 86.40$_{\pm 4.87}$ & 43.64$_{\pm 0.46}$
        & 97.25$_{\pm 0.66}$ \\
        \midrule
        \midrule
        \multirow{16}{*}{\textbf{HumanEval}}
        & MaxSeqProb~\citep{Vashurin2025LM-Polygraph}
        & 58.52$_{\pm 5.35}$ & 41.06$_{\pm 6.33}$ & 36.04$_{\pm 5.33}$
        & 28.61$_{\pm 3.51}$ & 55.66$_{\pm 12.02}$ & 27.78$_{\pm 3.45}$
        & 63.35$_{\pm 5.33}$ \\

        & MeanTokEnt~\citep{fomicheva2020PPL}
        & 50.57$_{\pm 4.55}$ & 29.60$_{\pm 4.21}$ & 26.33$_{\pm 4.19}$
        & 37.10$_{\pm 3.14}$ & 85.67$_{\pm 8.33}$ & 34.24$_{\pm 2.04}$
        & 73.06$_{\pm 4.19}$ \\

        & MaxTokProb~\citep{fomicheva2020PPL}
        & 47.81$_{\pm 4.77}$ & 29.66$_{\pm 4.09}$ & 26.88$_{\pm 4.10}$
        & 59.03$_{\pm 3.33}$ & 72.56$_{\pm 4.32}$ & 57.05$_{\pm 3.26}$
        & 72.51$_{\pm 4.10}$ \\

        & PPL~\citep{fomicheva2020PPL}
        & 52.00$_{\pm 4.65}$ & 30.44$_{\pm 4.31}$ & 27.36$_{\pm 4.22}$
        & 58.40$_{\pm 3.34}$ & 70.21$_{\pm 3.45}$ & 55.30$_{\pm 2.82}$
        & 72.03$_{\pm 4.22}$ \\

        & BSDetector~\citep{chen2024BSDETECTOR}
        & 52.71$_{\pm 4.89}$ & 34.97$_{\pm 5.63}$ & 31.35$_{\pm 5.06}$
        & 28.57$_{\pm 3.23}$ & 69.58$_{\pm 9.01}$ & 30.74$_{\pm 1.78}$
        & 68.04$_{\pm 5.06}$ \\

        & SelfCheckGPT~\citep{manakul2023selfcheckgpt}
        & 51.19$_{\pm 5.82}$ & 39.11$_{\pm 6.23}$ & 34.04$_{\pm 5.13}$
        & 16.75$_{\pm 3.23}$ & 65.21$_{\pm 14.08}$ & 23.91$_{\pm 2.14}$
        & 65.35$_{\pm 5.13}$ \\

        & NumSet~\citep{lin2024numset/eigv/deg/ecc}
        & 58.40$_{\pm 5.29}$ & 43.19$_{\pm 6.03}$ & 38.55$_{\pm 5.12}$
        & 17.50$_{\pm 3.37}$ & 76.69$_{\pm 19.08}$ & 23.02$_{\pm 2.69}$
        & 60.84$_{\pm 5.12}$ \\

        & EigV~\citep{lin2024numset/eigv/deg/ecc}
        & 60.39$_{\pm 5.36}$ & 43.78$_{\pm 6.33}$ & 38.82$_{\pm 5.16}$
        & 24.71$_{\pm 3.52}$ & 68.01$_{\pm 4.54}$ & 25.42$_{\pm 3.20}$
        & 60.57$_{\pm 5.16}$ \\

        & Deg~\citep{lin2024numset/eigv/deg/ecc}
        & 60.59$_{\pm 5.35}$ & 44.66$_{\pm 6.41}$ & 39.27$_{\pm 5.14}$
        & 24.78$_{\pm 3.45}$ & 67.45$_{\pm 3.56}$ & 25.13$_{\pm 3.20}$
        & 60.12$_{\pm 5.14}$ \\

        & Ecc~\citep{lin2024numset/eigv/deg/ecc}
        & 40.67$_{\pm 5.34}$ & 27.37$_{\pm 4.36}$ & 26.58$_{\pm 4.95}$
        & 54.12$_{\pm 3.72}$ & 78.04$_{\pm 9.78}$ & 51.57$_{\pm 2.94}$
        & 72.82$_{\pm 4.95}$ \\

        & ConU~\citep{wang2024ConU}
        & 54.08$_{\pm 4.91}$ & 35.64$_{\pm 5.66}$ & 32.33$_{\pm 4.97}$
        & 35.92$_{\pm 3.31}$ & 70.14$_{\pm 5.34}$ & 37.03$_{\pm 2.53}$
        & 67.06$_{\pm 4.97}$ \\

        & Semantic Entropy~\citep{farquhar2024SemanticEntropy}
        & 52.25$_{\pm 5.01}$ & 34.88$_{\pm 5.58}$ & 31.60$_{\pm 4.94}$
        & 39.02$_{\pm 3.62}$ & 79.20$_{\pm 12.20}$ & 37.90$_{\pm 2.19}$
        & 67.79$_{\pm 4.94}$ \\

        & Semantic Density~\citep{Qiu2024SemanticDensity}
        & 54.76$_{\pm 4.79}$ & 35.82$_{\pm 5.63}$ & 32.48$_{\pm 4.94}$
        & 30.44$_{\pm 3.23}$ & 69.04$_{\pm 7.48}$ & 33.34$_{\pm 2.35}$
        & 66.91$_{\pm 4.94}$ \\

        & MARS~\citep{Bakman2024MARS}
        & 58.39$_{\pm 4.78}$ & 39.68$_{\pm 6.02}$ & 35.28$_{\pm 5.15}$
        & 42.89$_{\pm 3.44}$ & 77.47$_{\pm 12.39}$ & 39.97$_{\pm 2.26}$
        & 64.11$_{\pm 5.15}$ \\

        & Verbalized Probability~\citep{Lin2022Verbalized_Probability}
        & 41.00$_{\pm 5.10}$ & 27.04$_{\pm 4.10}$ & 23.88$_{\pm 4.27}$
        & 21.56$_{\pm 3.57}$ & 92.83$_{\pm 6.92}$ & 27.40$_{\pm 1.98}$
        & 75.51$_{\pm 4.27}$ \\

        & $P(\text{True})$~\citep{Kadavath2022Ptrue/PIK}
        & 35.71$_{\pm 5.06}$ & 24.57$_{\pm 3.77}$ & 20.73$_{\pm 4.17}$
        & 31.78$_{\pm 3.76}$ & 97.44$_{\pm 4.17}$ & 33.27$_{\pm 1.73}$
        & 78.66$_{\pm 4.17}$ \\
        \bottomrule
    \end{tabular}
    }
\end{table*}

\begin{table*}[t]
    \centering
    \caption{Performance Comparison of the investigated LLM UQ methods of Llama-3.2-3B in answer-only generation across three evaluated benchmarks.}
    \label{llama-3.2-3B_answer_only_results}
    \resizebox{0.9\textwidth}{!}{
    \begin{tabular}{llccccccc}
        \toprule
        \multirow{2}{*}{\textbf{Datasets}}
        & \multirow{2}{*}{\textbf{Methods}}
        & \multicolumn{3}{c}{\textbf{Ranking}}
        & \multicolumn{3}{c}{\textbf{Calibration}}
        & \textbf{Coverage} \\
        \cmidrule(lr){3-5} \cmidrule(lr){6-8} \cmidrule(lr){9-9}
        & & \textbf{AUROC} & \textbf{AUPRC} & \textbf{AUARC}
        & \textbf{ECE} & \textbf{MCE} & \textbf{BS}
        & \textbf{AURC} \\
        \midrule
        \multirow{16}{*}{\textbf{TriviaQA}}
        & MaxSeqProb~\citep{Vashurin2025LM-Polygraph}
        & 59.73$_{\pm 0.57}$ & 59.36$_{\pm 0.81}$ & 58.54$_{\pm 0.78}$
        & 51.76$_{\pm 0.63}$ & 58.33$_{\pm 2.91}$ & 51.76$_{\pm 0.62}$
        & 41.45$_{\pm 0.78}$ \\

        & MeanTokEnt~\citep{fomicheva2020PPL}
        & 59.02$_{\pm 0.57}$ & 56.43$_{\pm 0.81}$ & 55.69$_{\pm 0.81}$
        & 5.52$_{\pm 0.58}$ & 94.66$_{\pm 0.47}$ & 24.84$_{\pm 0.16}$
        & 44.30$_{\pm 0.81}$ \\

        & MaxTokProb~\citep{fomicheva2020PPL}
        & 53.22$_{\pm 0.67}$ & 53.70$_{\pm 0.86}$ & 53.58$_{\pm 0.86}$
        & 46.86$_{\pm 0.63}$ & 86.71$_{\pm 2.62}$ & 46.72$_{\pm 0.63}$
        & 46.41$_{\pm 0.86}$ \\

        & PPL~\citep{fomicheva2020PPL}
        & 60.61$_{\pm 0.59}$ & 60.43$_{\pm 0.84}$ & 59.36$_{\pm 0.80}$
        & 37.87$_{\pm 0.62}$ & 44.89$_{\pm 4.16}$ & 38.68$_{\pm 0.51}$
        & 40.63$_{\pm 0.80}$ \\
        
        & BSDetector~\citep{chen2024BSDETECTOR}
        & 52.84$_{\pm 0.68}$ & 60.62$_{\pm 0.85}$ & 60.40$_{\pm 0.86}$
        & 12.33$_{\pm 0.57}$ & 78.94$_{\pm 11.15}$ & 25.02$_{\pm 0.18}$
        & 39.59$_{\pm 0.86}$ \\

        & SelfCheckGPT~\citep{manakul2023selfcheckgpt}
        & 83.33$_{\pm 0.50}$ & 81.12$_{\pm 0.77}$ & 74.90$_{\pm 0.68}$
        & 20.45$_{\pm 0.48}$ & 35.67$_{\pm 1.70}$ & 21.22$_{\pm 0.33}$
        & 25.09$_{\pm 0.68}$ \\

        & NumSet~\citep{lin2024numset/eigv/deg/ecc}
        & 83.05$_{\pm 0.50}$ & 79.58$_{\pm 0.67}$ & 74.84$_{\pm 0.68}$
        & 27.70$_{\pm 0.49}$ & 43.90$_{\pm 1.42}$ & 25.26$_{\pm 0.38}$
        & 25.15$_{\pm 0.68}$ \\

        & EigV~\citep{lin2024numset/eigv/deg/ecc}
        & 83.61$_{\pm 0.49}$ & 81.47$_{\pm 0.65}$ & 75.28$_{\pm 0.65}$
        & 26.91$_{\pm 0.50}$ & 43.45$_{\pm 1.44}$ & 17.18$_{\pm 0.32}$
        & 24.71$_{\pm 0.65}$ \\

        & Deg~\citep{lin2024numset/eigv/deg/ecc}
        & 83.59$_{\pm 0.49}$ & 81.22$_{\pm 0.65}$ & 75.26$_{\pm 0.66}$
        & 18.96$_{\pm 0.43}$ & 14.94$_{\pm 1.86}$ & 17.18$_{\pm 0.32}$
        & 24.72$_{\pm 0.66}$ \\

        & Ecc~\citep{lin2024numset/eigv/deg/ecc}
        & 26.12$_{\pm 0.57}$ & 45.88$_{\pm 0.62}$ & 31.41$_{\pm 0.70}$
        & 24.95$_{\pm 0.57}$ & 80.58$_{\pm 1.46}$ & 31.10$_{\pm 0.30}$
        & 68.58$_{\pm 0.70}$ \\

        & ConU~\citep{wang2024ConU}
        & 80.61$_{\pm 0.51}$ & 78.17$_{\pm 0.75}$ & 71.85$_{\pm 0.69}$
        & 30.27$_{\pm 0.56}$ & 50.06$_{\pm 1.92}$ & 28.86$_{\pm 0.47}$
        & 28.14$_{\pm 0.69}$ \\

        & Semantic Entropy~\citep{farquhar2024SemanticEntropy}
        & 81.41$_{\pm 0.50}$ & 78.52$_{\pm 0.75}$ & 72.12$_{\pm 0.68}$
        & 30.60$_{\pm 0.55}$ & 44.16$_{\pm 1.72}$ & 28.10$_{\pm 0.44}$
        & 27.86$_{\pm 0.68}$ \\

        & Semantic Density~\citep{Qiu2024SemanticDensity}
        & 80.88$_{\pm 0.51}$ & 78.28$_{\pm 0.75}$ & 71.93$_{\pm 0.69}$
        & 24.93$_{\pm 0.54}$ & 44.34$_{\pm 2.02}$ & 25.30$_{\pm 0.43}$
        & 28.06$_{\pm 0.69}$ \\

        & MARS~\citep{Bakman2024MARS}
        & 58.37$_{\pm 0.58}$ & 58.68$_{\pm 0.85}$ & 58.05$_{\pm 0.81}$
        & 25.27$_{\pm 0.62}$ & 55.48$_{\pm 28.71}$ & 31.00$_{\pm 0.37}$
        & 41.94$_{\pm 0.81}$ \\

        & Verbalized Probability~\citep{Lin2022Verbalized_Probability}
        & 51.83$_{\pm 0.59}$ & 48.12$_{\pm 0.65}$ & 48.07$_{\pm 0.75}$
        & 43.59$_{\pm 0.56}$ & 51.43$_{\pm 0.65}$ & 45.17$_{\pm 0.48}$
        & 51.92$_{\pm 0.75}$ \\
        
        & $P(\text{True})$~\citep{Kadavath2022Ptrue/PIK}
        & 48.68$_{\pm 0.69}$ & 49.57$_{\pm 0.76}$ & 48.38$_{\pm 0.81}$
        & 21.03$_{\pm 0.64}$ & 70.79$_{\pm 2.91}$ & 30.60$_{\pm 0.32}$
        & 51.61$_{\pm 0.81}$ \\
        \midrule
        \midrule
        \multirow{16}{*}{\textbf{GSM8K}}
        & MaxSeqProb~\citep{Vashurin2025LM-Polygraph}
        & 58.33$_{\pm 2.89}$ & 10.75$_{\pm 1.82}$ & 9.11$_{\pm 1.13}$
        & 6.99$_{\pm 0.66}$ & 90.50$_{\pm 14.58}$ & 7.23$_{\pm 0.65}$
        & 90.81$_{\pm 1.13}$ \\

        & MeanTokEnt~\citep{fomicheva2020PPL}
        & 50.57$_{\pm 3.20}$ & 7.75$_{\pm 1.00}$ & 7.17$_{\pm 0.89}$
        & 23.84$_{\pm 0.88}$ & 92.50$_{\pm 6.64}$ & 16.93$_{\pm 0.60}$
        & 92.75$_{\pm 0.89}$ \\

        & MaxTokProb~\citep{fomicheva2020PPL}
        & 55.93$_{\pm 2.98}$ & 8.47$_{\pm 1.06}$ & 7.84$_{\pm 0.92}$
        & 89.09$_{\pm 0.67}$ & 96.43$_{\pm 4.98}$ & 86.04$_{\pm 0.65}$
        & 92.08$_{\pm 0.92}$ \\

        & PPL~\citep{fomicheva2020PPL}
        & 58.49$_{\pm 2.89}$ & 10.79$_{\pm 1.83}$ & 9.14$_{\pm 1.13}$
        & 62.00$_{\pm 0.75}$ & 85.93$_{\pm 1.79}$ & 47.17$_{\pm 0.63}$
        & 90.79$_{\pm 1.13}$ \\
        
        & BSDetector~\citep{chen2024BSDETECTOR}
        & 51.40$_{\pm 2.93}$ & 7.65$_{\pm 0.93}$ & 7.13$_{\pm 0.86}$
        & 35.97$_{\pm 0.71}$ & 96.11$_{\pm 3.03}$ & 21.57$_{\pm 0.39}$
        & 92.79$_{\pm 0.86}$ \\

        & SelfCheckGPT~\citep{manakul2023selfcheckgpt}
        & 66.02$_{\pm 2.98}$ & 16.50$_{\pm 2.87}$ & 11.72$_{\pm 1.31}$
        & 24.46$_{\pm 0.85}$ & 78.83$_{\pm 5.01}$ & 16.84$_{\pm 0.63}$
        & 88.20$_{\pm 1.31}$ \\

        & NumSet~\citep{lin2024numset/eigv/deg/ecc}
        & 68.01$_{\pm 3.08}$ & 15.05$_{\pm 2.41}$ & 12.05$_{\pm 1.35}$
        & 41.49$_{\pm 0.87}$ & 79.09$_{\pm 3.21}$ & 27.35$_{\pm 0.68}$
        & 87.87$_{\pm 1.35}$ \\

        & EigV~\citep{lin2024numset/eigv/deg/ecc}
        & 67.17$_{\pm 3.08}$ & 15.47$_{\pm 2.54}$ & 11.90$_{\pm 1.35}$
        & 41.44$_{\pm 0.86}$ & 74.48$_{\pm 3.36}$ & 11.74$_{\pm 0.62}$
        & 88.03$_{\pm 1.35}$ \\

        & Deg~\citep{lin2024numset/eigv/deg/ecc}
        & 67.13$_{\pm 3.09}$ & 15.33$_{\pm 2.52}$ & 11.88$_{\pm 1.36}$
        & 24.22$_{\pm 0.84}$ & 62.25$_{\pm 3.75}$ & 11.74$_{\pm 0.62}$
        & 88.04$_{\pm 1.36}$ \\

        & Ecc~\citep{lin2024numset/eigv/deg/ecc}
        & 32.62$_{\pm 3.02}$ & 5.43$_{\pm 0.55}$ & 4.15$_{\pm 0.66}$
        & 31.04$_{\pm 1.02}$ & 80.68$_{\pm 3.02}$ & 21.57$_{\pm 0.64}$
        & 95.78$_{\pm 0.66}$ \\

        & ConU~\citep{wang2024ConU}
        & 59.85$_{\pm 3.52}$ & 13.95$_{\pm 2.57}$ & 10.52$_{\pm 1.30}$
        & 58.61$_{\pm 0.96}$ & 87.02$_{\pm 1.57}$ & 47.91$_{\pm 0.96}$
        & 89.40$_{\pm 1.30}$ \\

        & Semantic Entropy~\citep{farquhar2024SemanticEntropy}
        & 60.44$_{\pm 3.58}$ & 14.04$_{\pm 2.58}$ & 10.59$_{\pm 1.31}$
        & 59.27$_{\pm 0.87}$ & 85.20$_{\pm 1.75}$ & 46.64$_{\pm 0.87}$
        & 89.33$_{\pm 1.31}$ \\

        & Semantic Density~\citep{Qiu2024SemanticDensity}
        & 59.97$_{\pm 3.57}$ & 13.97$_{\pm 2.57}$ & 10.53$_{\pm 1.31}$
        & 51.24$_{\pm 0.99}$ & 85.63$_{\pm 1.73}$ & 40.66$_{\pm 0.99}$
        & 89.39$_{\pm 1.31}$ \\

        & MARS~\citep{Bakman2024MARS}
        & 57.25$_{\pm 3.00}$ & 9.18$_{\pm 1.27}$ & 8.38$_{\pm 1.01}$
        & 37.86$_{\pm 0.79}$ & 93.84$_{\pm 3.16}$ & 23.74$_{\pm 0.51}$
        & 91.55$_{\pm 1.01}$ \\

        & Verbalized Probability~\citep{Lin2022Verbalized_Probability}
        & 46.58$_{\pm 2.44}$ & 6.70$_{\pm 0.72}$ & 6.63$_{\pm 0.89}$
        & 74.93$_{\pm 1.20}$ & 93.47$_{\pm 0.76}$ & 74.46$_{\pm 1.17}$
        & 93.29$_{\pm 0.89}$ \\

        & $P(\text{True})$~\citep{Kadavath2022Ptrue/PIK}
        & 63.16$_{\pm 4.19}$ & 6.64$_{\pm 1.59}$ & 5.23$_{\pm 0.93}$
        & 45.02$_{\pm 0.67}$ & 94.63$_{\pm 1.21}$ & 25.60$_{\pm 0.48}$
        & 94.69$_{\pm 0.93}$ \\
        \midrule
        \midrule
        \multirow{16}{*}{\textbf{HumanEval}}
        & MaxSeqProb~\citep{Vashurin2025LM-Polygraph}
        & 42.98$_{\pm 4.76}$ & 38.90$_{\pm 5.52}$ & 35.44$_{\pm 5.49}$
        & 36.12$_{\pm 3.77}$ & 43.71$_{\pm 13.16}$ & 36.25$_{\pm 3.74}$
        & 63.95$_{\pm 5.49}$ \\

        & MeanTokEnt~\citep{fomicheva2020PPL}
        & 45.96$_{\pm 4.43}$ & 37.37$_{\pm 4.68}$ & 33.86$_{\pm 4.91}$
        & 31.82$_{\pm 3.69}$ & 89.06$_{\pm 16.12}$ & 35.39$_{\pm 1.91}$
        & 65.53$_{\pm 4.91}$ \\

        & MaxTokProb~\citep{fomicheva2020PPL}
        & 58.23$_{\pm 4.75}$ & 46.59$_{\pm 6.07}$ & 43.16$_{\pm 5.42}$
        & 57.39$_{\pm 3.60}$ & 79.45$_{\pm 5.45}$ & 56.08$_{\pm 3.40}$
        & 56.23$_{\pm 5.42}$ \\

        & PPL~\citep{fomicheva2020PPL}
        & 44.37$_{\pm 4.53}$ & 36.39$_{\pm 4.61}$ & 32.62$_{\pm 4.93}$
        & 39.28$_{\pm 3.69}$ & 91.97$_{\pm 11.92}$ & 40.17$_{\pm 2.28}$
        & 66.77$_{\pm 4.93}$ \\

        & BSDetector~\citep{chen2024BSDETECTOR}
        & 47.41$_{\pm 4.89}$ & 41.64$_{\pm 5.69}$ & 37.95$_{\pm 5.50}$
        & 29.48$_{\pm 3.96}$ & 67.25$_{\pm 7.34}$ & 34.48$_{\pm 2.06}$
        & 61.44$_{\pm 5.50}$ \\

        & SelfCheckGPT~\citep{manakul2023selfcheckgpt}
        & 51.02$_{\pm 5.38}$ & 46.99$_{\pm 6.41}$ & 42.61$_{\pm 5.60}$
        & 15.65$_{\pm 3.35}$ & 93.07$_{\pm 12.31}$ & 25.82$_{\pm 1.85}$
        & 56.78$_{\pm 5.60}$ \\

        & NumSet~\citep{lin2024numset/eigv/deg/ecc}
        & 61.20$_{\pm 4.47}$ & 51.03$_{\pm 5.82}$ & 48.39$_{\pm 5.30}$
        & 24.83$_{\pm 3.44}$ & 45.19$_{\pm 14.04}$ & 27.98$_{\pm 3.01}$
        & 51.00$_{\pm 5.30}$ \\

        & EigV~\citep{lin2024numset/eigv/deg/ecc}
        & 61.37$_{\pm 4.49}$ & 51.98$_{\pm 6.01}$ & 48.50$_{\pm 5.34}$
        & 31.11$_{\pm 3.82}$ & 75.38$_{\pm 2.91}$ & 31.34$_{\pm 3.46}$
        & 50.89$_{\pm 5.34}$ \\

        & Deg~\citep{lin2024numset/eigv/deg/ecc}
        & 61.38$_{\pm 4.49}$ & 51.81$_{\pm 6.00}$ & 48.52$_{\pm 5.34}$
        & 31.11$_{\pm 3.82}$ & 75.38$_{\pm 2.91}$ & 31.34$_{\pm 3.46}$
        & 50.87$_{\pm 5.34}$ \\

        & Ecc~\citep{lin2024numset/eigv/deg/ecc}
        & 39.02$_{\pm 4.45}$ & 32.92$_{\pm 4.04}$ & 31.47$_{\pm 4.79}$
        & 54.05$_{\pm 3.91}$ & 87.51$_{\pm 9.33}$ & 53.73$_{\pm 3.40}$
        & 67.92$_{\pm 4.79}$ \\

        & ConU~\citep{wang2024ConU}
        & 57.55$_{\pm 4.27}$ & 48.20$_{\pm 5.62}$ & 44.16$_{\pm 5.02}$
        & 32.41$_{\pm 3.64}$ & 61.53$_{\pm 6.46}$ & 35.83$_{\pm 2.74}$
        & 55.23$_{\pm 5.02}$ \\

        & Semantic Entropy~\citep{farquhar2024SemanticEntropy}
        & 55.14$_{\pm 4.36}$ & 47.12$_{\pm 5.57}$ & 43.22$_{\pm 5.02}$
        & 30.88$_{\pm 3.83}$ & 61.73$_{\pm 8.51}$ & 35.52$_{\pm 2.47}$
        & 56.17$_{\pm 5.02}$ \\

        & Semantic Density~\citep{Qiu2024SemanticDensity}
        & 57.61$_{\pm 4.32}$ & 48.27$_{\pm 5.62}$ & 44.18$_{\pm 4.99}$
        & 29.01$_{\pm 3.66}$ & 62.30$_{\pm 7.62}$ & 33.97$_{\pm 2.63}$
        & 55.21$_{\pm 4.99}$ \\

        & MARS~\citep{Bakman2024MARS}
        & 53.19$_{\pm 4.66}$ & 44.46$_{\pm 5.61}$ & 41.12$_{\pm 5.15}$
        & 38.79$_{\pm 3.82}$ & 56.79$_{\pm 7.74}$ & 39.12$_{\pm 2.44}$
        & 58.27$_{\pm 5.15}$ \\

        & Verbalized Probability~\citep{Lin2022Verbalized_Probability}
        & 46.77$_{\pm 5.03}$ & 40.54$_{\pm 5.98}$ & 37.46$_{\pm 5.64}$
        & 32.50$_{\pm 3.72}$ & 61.73$_{\pm 9.47}$ & 34.66$_{\pm 1.93}$
        & 61.93$_{\pm 5.64}$ \\

        & $P(\text{True})$~\citep{Kadavath2022Ptrue/PIK}
        & 53.02$_{\pm 4.78}$ & 43.90$_{\pm 5.68}$ & 40.68$_{\pm 5.25}$
        & 31.79$_{\pm 3.64}$ & 90.03$_{\pm 14.75}$ & 33.11$_{\pm 1.72}$
        & 58.71$_{\pm 5.25}$ \\
        \bottomrule
    \end{tabular}
    }
\end{table*}

\clearpage
\begin{table*}[t]
    \centering
    \caption{Performance Comparison of the investigated LLM UQ methods of DeepSeek-V3 in answer-only generation across three evaluated benchmarks.}
    \label{deepseek-v3_answer_only_results}
    \resizebox{0.9\textwidth}{!}{
    \begin{tabular}{llccccccc}
        \toprule
        \multirow{2}{*}{\textbf{Datasets}}
        & \multirow{2}{*}{\textbf{Methods}}
        & \multicolumn{3}{c}{\textbf{Ranking}}
        & \multicolumn{3}{c}{\textbf{Calibration}}
        & \textbf{Coverage} \\
        \cmidrule(lr){3-5} \cmidrule(lr){6-8} \cmidrule(lr){9-9}
        & & \textbf{AUROC} & \textbf{AUPRC} & \textbf{AUARC}
        & \textbf{ECE} & \textbf{MCE} & \textbf{BS}
        & \textbf{AURC} \\
        \midrule
        \multirow{7}{*}{\textbf{TriviaQA}}
        & BSDetector~\citep{chen2024BSDETECTOR}
        & 58.23$_{\pm 0.97}$ & 75.05$_{\pm 0.73}$ & 74.21$_{\pm 0.80}$
        & 19.09$_{\pm 0.46}$ & 46.45$_{\pm 4.76}$ & 18.56$_{\pm 0.43}$
        & 25.78$_{\pm 0.80}$ \\

        & SelfCheckGPT~\citep{manakul2023selfcheckgpt}
        & 47.92$_{\pm 0.95}$ & 71.05$_{\pm 0.74}$ & 69.88$_{\pm 0.84}$
        & 19.77$_{\pm 0.50}$ & 46.96$_{\pm 2.83}$ & 20.98$_{\pm 0.46}$
        & 30.11$_{\pm 0.84}$ \\

        & NumSet~\citep{lin2024numset/eigv/deg/ecc}
        & 57.18$_{\pm 0.45}$ & 79.02$_{\pm 0.49}$ & 78.89$_{\pm 0.62}$
        & 5.98$_{\pm 0.50}$ & 75.76$_{\pm 27.45}$ & 17.60$_{\pm 0.32}$
        & 21.10$_{\pm 0.62}$ \\

        & EigV~\citep{lin2024numset/eigv/deg/ecc}
        & 57.33$_{\pm 0.45}$ & 79.09$_{\pm 0.49}$ & 79.00$_{\pm 0.62}$
        & 5.98$_{\pm 0.50}$ & 57.50$_{\pm 5.69}$ & 21.29$_{\pm 0.47}$
        & 20.99$_{\pm 0.62}$ \\

        & Deg~\citep{lin2024numset/eigv/deg/ecc}
        & 57.33$_{\pm 0.45}$ & 79.08$_{\pm 0.49}$ & 79.00$_{\pm 0.62}$
        & 19.32$_{\pm 0.49}$ & 21.22$_{\pm 0.49}$ & 21.29$_{\pm 0.47}$
        & 20.99$_{\pm 0.62}$ \\

        & Ecc~\citep{lin2024numset/eigv/deg/ecc}
        & 48.37$_{\pm 0.23}$ & 76.17$_{\pm 0.49}$ & 73.49$_{\pm 0.73}$
        & 2.70$_{\pm 0.51}$ & 59.72$_{\pm 8.55}$ & 18.19$_{\pm 0.25}$
        & 26.50$_{\pm 0.73}$ \\

        & ConU~\citep{wang2024ConU}
        & 57.14$_{\pm 0.43}$ & 77.85$_{\pm 0.51}$ & 77.78$_{\pm 0.62}$
        & 22.44$_{\pm 0.50}$ & 54.96$_{\pm 11.87}$ & 22.54$_{\pm 0.49}$
        & 22.21$_{\pm 0.62}$ \\
        \midrule
        \midrule
        \multirow{7}{*}{\textbf{GSM8K}}
        & BSDetector~\citep{chen2024BSDETECTOR}
        & 80.98$_{\pm 1.14}$ & 53.81$_{\pm 2.14}$ & 47.39$_{\pm 1.86}$
        & 14.55$_{\pm 1.02}$ & 86.03$_{\pm 5.83}$ & 16.81$_{\pm 0.40}$
        & 52.54$_{\pm 1.86}$ \\

        & SelfCheckGPT~\citep{manakul2023selfcheckgpt}
        & 67.75$_{\pm 1.51}$ & 43.36$_{\pm 2.33}$ & 41.16$_{\pm 2.05}$
        & 48.78$_{\pm 1.29}$ & 82.21$_{\pm 2.74}$ & 45.39$_{\pm 1.18}$
        & 58.77$_{\pm 2.05}$ \\

        & NumSet~\citep{lin2024numset/eigv/deg/ecc}
        & 68.97$_{\pm 1.14}$ & 44.06$_{\pm 1.78}$ & 42.52$_{\pm 2.05}$
        & 58.39$_{\pm 1.24}$ & 71.52$_{\pm 1.87}$ & 54.41$_{\pm 1.18}$
        & 57.40$_{\pm 2.05}$ \\

        & EigV~\citep{lin2024numset/eigv/deg/ecc}
        & 68.48$_{\pm 1.17}$ & 44.29$_{\pm 1.79}$ & 42.45$_{\pm 2.01}$
        & 45.28$_{\pm 1.26}$ & 53.70$_{\pm 2.06}$ & 43.13$_{\pm 1.20}$
        & 57.47$_{\pm 2.01}$ \\

        & Deg~\citep{lin2024numset/eigv/deg/ecc}
        & 68.47$_{\pm 1.17}$ & 44.22$_{\pm 1.79}$ & 42.39$_{\pm 2.01}$
        & 46.49$_{\pm 1.26}$ & 54.25$_{\pm 2.78}$ & 43.13$_{\pm 1.20}$
        & 57.53$_{\pm 2.01}$ \\

        & Ecc~\citep{lin2024numset/eigv/deg/ecc}
        & 38.75$_{\pm 0.78}$ & 31.80$_{\pm 1.27}$ & 21.68$_{\pm 1.31}$
        & 33.51$_{\pm 1.33}$ & 44.04$_{\pm 3.34}$ & 34.80$_{\pm 1.29}$
        & 78.24$_{\pm 1.31}$ \\

        & ConU~\citep{wang2024ConU}
        & 68.22$_{\pm 1.17}$ & 44.09$_{\pm 1.79}$ & 42.37$_{\pm 2.01}$
        & 50.94$_{\pm 1.26}$ & 81.35$_{\pm 3.60}$ & 46.90$_{\pm 1.19}$
        & 57.55$_{\pm 2.01}$ \\
        \midrule
        \midrule
        \multirow{7}{*}{\textbf{HumanEval}}
        & BSDetector~\citep{chen2024BSDETECTOR}
        & 44.77$_{\pm 4.61}$ & 34.88$_{\pm 4.72}$ & 31.02$_{\pm 4.78}$
        & 50.32$_{\pm 3.94}$ & 85.15$_{\pm 12.51}$ & 50.39$_{\pm 3.29}$
        & 68.37$_{\pm 4.78}$ \\

        & SelfCheckGPT~\citep{manakul2023selfcheckgpt}
        & 41.24$_{\pm 4.58}$ & 31.44$_{\pm 4.43}$ & 28.13$_{\pm 4.51}$
        & 49.86$_{\pm 4.06}$ & 72.19$_{\pm 4.93}$ & 51.19$_{\pm 3.13}$
        & 71.26$_{\pm 4.51}$ \\

        & NumSet~\citep{lin2024numset/eigv/deg/ecc}
        & 41.16$_{\pm 4.77}$ & 30.47$_{\pm 3.93}$ & 29.35$_{\pm 4.94}$
        & 55.67$_{\pm 4.15}$ & 89.51$_{\pm 14.09}$ & 56.17$_{\pm 3.40}$
        & 70.04$_{\pm 4.94}$ \\

        & EigV~\citep{lin2024numset/eigv/deg/ecc}
        & 41.38$_{\pm 4.89}$ & 30.19$_{\pm 4.09}$ & 29.38$_{\pm 4.80}$
        & 47.04$_{\pm 4.07}$ & 79.03$_{\pm 12.91}$ & 49.82$_{\pm 3.15}$
        & 70.01$_{\pm 4.80}$ \\

        & Deg~\citep{lin2024numset/eigv/deg/ecc}
        & 41.48$_{\pm 4.89}$ & 30.35$_{\pm 4.08}$ & 29.35$_{\pm 4.81}$
        & 47.04$_{\pm 4.07}$ & 79.03$_{\pm 12.91}$ & 49.82$_{\pm 3.15}$
        & 70.04$_{\pm 4.81}$ \\

        & Ecc~\citep{lin2024numset/eigv/deg/ecc}
        & 55.44$_{\pm 4.33}$ & 39.38$_{\pm 4.90}$ & 38.23$_{\pm 5.20}$
        & 30.22$_{\pm 3.54}$ & 65.81$_{\pm 19.28}$ & 30.93$_{\pm 3.38}$
        & 61.16$_{\pm 5.20}$ \\

        & ConU~\citep{wang2024ConU}
        & 42.01$_{\pm 4.88}$ & 30.57$_{\pm 4.07}$ & 29.47$_{\pm 4.76}$
        & 49.27$_{\pm 4.18}$ & 90.58$_{\pm 12.76}$ & 51.65$_{\pm 3.23}$
        & 69.92$_{\pm 4.76}$ \\
        \bottomrule
    \end{tabular}
    }
\end{table*}

\subsection{Performance Comparison of LLM UQ Methods in Reasoning-Augmented Generation}
\label{subsec:reasoning_augmented_uq}
This subsection presents the performance comparison of LLM UQ methods under the reasoning-augmented generation setting, where the LLM is prompted in a chain-of-thought template~\citep{wei2022CoT} to generate an intermediate reasoning chain before generating the final answer. The evaluated UQ methods include those introduced in Subsection~\ref{subsec:answer_only_uq} and two additional UQ methods designed specifically for the reasoning-augmented generation setting, namely CoT-UQ~\citep{Zhang2025CoT-UQ} and Topo-UQ~\citep{Da2025Topo-uq}. Following the configurations recommended by~\citet{Zhang2025CoT-UQ}, we here consider four implementations of CoT-UQ, including CoT-UQ-ME, CoT-UQ-MI, and CoT-UQ-SAR as white-box variants and CoT-UQ-SP as the black-box variant.

Tables~\ref{qwen3_1_7b_answer_reasoning_results}, \ref{qwen3_4b_answer_reasoning_results}, \ref{Llama-3.2-1B_answer_reasoning_results}, and \ref{Llama-3.2-3B_answer_reasoning_results} list the performance of the investigated LLM UQ methods under the reasoning-augmented generation setting for Qwen3-1.7B, Qwen3-4B, Llama-3.2-1B, and Llama-3.2-3B. We have three primary observations. First, compared with the answer-only generation, the averaged performance of the reasoning-augmented generation, across LLMs and UQ methods, improves by 4.51\%, 32.99\% in ranking, 13.50\%, 30.00\% in calibration, and 6.78\%, 48.01\% in coverage on TriviaQA and GSM8K, respectively. We conjecture that these improvements profit from informative estimates within the reasoning chain, such as the stability of intermediate rationales~\citep{Da2025Topo-uq} and the consistency between reasoning steps and the final answer~\citep{Zhang2025CoT-UQ}, which contribute to distinguishing correct from incorrect generations in QA and mathematical reasoning tasks. In contrast, compared with the answer-only generation, the averaged performance of the reasoning-augmented generation, across LLMs and UQ methods, declines by 2.12\%, 0.96\%, and 1.87\% in ranking, calibration, and coverage on HumanEval, respectively. This decline contrasts with the improvements on TriviaQA and GSM8K, indicating additional estimates from the reasoning chain may degrade UQ performance in code generation tasks. Summing up the first group of observations, it suggests that the reasoning-augmented generation setting may be beneficial to UQ methods in QA and mathematical reasoning tasks, whereas the simpler answer-only setting may be preferable for code generation tasks.

Second, compared to the answer-only generation setting, all consensus-based methods obtain improvements of 16.26\%, 16.01\%, and 19.60\% in ranking, calibration, and coverage, respectively, whereas single-pass methods achieve much smaller improvements of only 9.54\%, 11.13\%, and 14.41\% correspondingly. A possible reason is that responses with reasoning chains contain richer semantic and structural information than answer-only ones, such as intermediate reasoning paths, explanations, and derivations. Consensus-based methods can leverage this additional information provided by reasoning-chain responses to estimate semantic consistency, reasoning diversity, and graph-based relations across different reasoning chains led by multiple sampled outputs, thereby capturing more informative uncertainty estimates, whereas single-pass methods typically rely on information from a single generation, such as token probabilities or self-evaluated confidence, limiting their ability to exploit inter-response disagreement signals. 

Third, compared to the answer-only setting, UQ methods for the Qwen3 family achieve improvements in the average of 23.64\%, 20.93\%, and 31.04\% in ranking, calibration, and coverage, respectively. By contrast, the corresponding improvements for the Llama 3.2 family are only 3.84\%, 7.43\%, and 4.27\%. We conjecture that LLMs in the Qwen3 family generate more informative, coherent, and answer-aligned reasoning chains, thereby providing more reliable evidence for uncertainty quantification.

\clearpage
\begin{table*}[t]
    \centering
    \caption{Performance Comparison of the investigated LLM UQ methods of Qwen3-1.7B in reasoning-augmented generation across three evaluated benchmarks.}
    \label{qwen3_1_7b_answer_reasoning_results}
    \resizebox{0.7\textwidth}{!}{
    \begin{tabular}{llccccccc}
        \toprule
        \multirow{2}{*}{\textbf{Datasets}}
        & \multirow{2}{*}{\textbf{Methods}}
        & \multicolumn{3}{c}{\textbf{Ranking}}
        & \multicolumn{3}{c}{\textbf{Calibration}}
        & \textbf{Coverage} \\
        \cmidrule(lr){3-5} \cmidrule(lr){6-8} \cmidrule(lr){9-9}
        & & \textbf{AUROC} & \textbf{AUPRC} & \textbf{AUARC}
        & \textbf{ECE} & \textbf{MCE} & \textbf{BS}
        & \textbf{AURC} \\
        \midrule
        \multirow{21}{*}{\textbf{TriviaQA}}
        & MaxSeqProb~\citep{Vashurin2025LM-Polygraph}
        & 69.20$_{\pm 1.32}$ & 49.29$_{\pm 2.30}$ & 43.51$_{\pm 1.76}$
        & 28.02$_{\pm 1.20}$ & 57.11$_{\pm 4.29}$ & 27.93$_{\pm 1.17}$
        & 56.42$_{\pm 1.76}$ \\

        & MeanTokEnt~\citep{fomicheva2020PPL}
        & 67.20$_{\pm 1.49}$ & 46.69$_{\pm 2.27}$ & 41.99$_{\pm 1.73}$
        & 39.21$_{\pm 1.16}$ & 44.31$_{\pm 2.04}$ & 34.74$_{\pm 0.59}$
        & 57.95$_{\pm 1.73}$ \\

        & MaxTokProb~\citep{fomicheva2020PPL}
        & 46.51$_{\pm 1.30}$ & 28.09$_{\pm 1.28}$ & 27.88$_{\pm 1.65}$
        & 20.39$_{\pm 1.21}$ & 20.39$_{\pm 1.21}$ & 25.00$_{\pm 0.00}$
        & 72.05$_{\pm 1.65}$ \\

        & PPL~\citep{fomicheva2020PPL}
        & 66.43$_{\pm 1.47}$ & 46.50$_{\pm 2.22}$ & 41.77$_{\pm 1.70}$
        & 47.45$_{\pm 1.17}$ & 51.77$_{\pm 1.69}$ & 42.07$_{\pm 0.74}$
        & 58.16$_{\pm 1.70}$ \\

        & BSDetector~\citep{chen2024BSDETECTOR}
        & 65.46$_{\pm 1.54}$ & 46.07$_{\pm 2.46}$ & 41.38$_{\pm 1.80}$
        & 23.09$_{\pm 1.18}$ & 61.45$_{\pm 6.28}$ & 26.96$_{\pm 0.94}$
        & 58.55$_{\pm 1.80}$ \\

        & SelfCheckGPT~\citep{manakul2023selfcheckgpt}
        & 81.97$_{\pm 1.19}$ & 65.90$_{\pm 2.42}$ & 52.24$_{\pm 1.74}$
        & 43.08$_{\pm 1.03}$ & 59.57$_{\pm 10.12}$ & 34.62$_{\pm 0.77}$
        & 47.69$_{\pm 1.74}$ \\

        & NumSet~\citep{lin2024numset/eigv/deg/ecc}
        & 84.58$_{\pm 0.98}$ & 65.08$_{\pm 2.12}$ & 53.20$_{\pm 1.77}$
        & 34.68$_{\pm 1.01}$ & 54.29$_{\pm 2.47}$ & 27.52$_{\pm 0.74}$
        & 46.73$_{\pm 1.77}$ \\

        & EigV~\citep{lin2024numset/eigv/deg/ecc}
        & 84.35$_{\pm 1.06}$ & 66.48$_{\pm 2.20}$ & 53.16$_{\pm 1.81}$
        & 13.43$_{\pm 0.96}$ & 51.32$_{\pm 24.68}$ & 16.52$_{\pm 0.66}$
        & 46.77$_{\pm 1.81}$ \\

        & Deg~\citep{lin2024numset/eigv/deg/ecc}
        & 84.35$_{\pm 1.06}$ & 66.48$_{\pm 2.20}$ & 53.16$_{\pm 1.81}$
        & 13.43$_{\pm 0.96}$ & 51.32$_{\pm 24.68}$ & 16.52$_{\pm 0.66}$
        & 46.77$_{\pm 1.81}$ \\

        & Ecc~\citep{lin2024numset/eigv/deg/ecc}
        & 19.26$_{\pm 1.01}$ & 23.11$_{\pm 0.99}$ & 11.88$_{\pm 0.86}$
        & 35.22$_{\pm 1.26}$ & 93.05$_{\pm 0.95}$ & 39.79$_{\pm 0.56}$
        & 88.05$_{\pm 0.86}$ \\

        & ConU~\citep{wang2024ConU}
        & 83.39$_{\pm 1.13}$ & 65.98$_{\pm 2.31}$ & 52.40$_{\pm 1.76}$
        & 24.31$_{\pm 0.99}$ & 43.90$_{\pm 4.16}$ & 21.06$_{\pm 0.68}$
        & 47.54$_{\pm 1.76}$ \\

        & Semantic Entropy~\citep{farquhar2024SemanticEntropy}
        & 80.56$_{\pm 1.36}$ & 64.72$_{\pm 2.30}$ & 51.67$_{\pm 1.77}$
        & 8.83$_{\pm 0.91}$ & 32.67$_{\pm 8.69}$ & 15.95$_{\pm 0.76}$
        & 48.26$_{\pm 1.77}$ \\

        & Semantic Density~\citep{Qiu2024SemanticDensity}
        & 83.95$_{\pm 1.08}$ & 67.83$_{\pm 2.33}$ & 53.31$_{\pm 1.80}$
        & 19.35$_{\pm 0.98}$ & 42.92$_{\pm 4.69}$ & 18.76$_{\pm 0.65}$
        & 46.63$_{\pm 1.80}$ \\

        & MARS~\citep{Bakman2024MARS}
        & 65.61$_{\pm 1.35}$ & 42.93$_{\pm 2.05}$ & 39.82$_{\pm 1.67}$
        & 47.96$_{\pm 1.20}$ & 78.90$_{\pm 23.58}$ & 42.66$_{\pm 0.77}$
        & 60.11$_{\pm 1.67}$ \\

        & Verbalized Probability~\citep{Lin2022Verbalized_Probability}
        & 57.25$_{\pm 1.65}$ & 37.75$_{\pm 2.34}$ & 35.73$_{\pm 1.92}$
        & 29.80$_{\pm 1.32}$ & 53.27$_{\pm 3.41}$ & 30.94$_{\pm 0.65}$
        & 64.20$_{\pm 1.92}$ \\

        & $P(\text{True})$~\citep{Kadavath2022Ptrue/PIK}
        & 63.49$_{\pm 1.67}$ & 42.61$_{\pm 2.19}$ & 39.58$_{\pm 1.74}$
        & 48.91$_{\pm 1.38}$ & 65.88$_{\pm 2.54}$ & 49.19$_{\pm 1.28}$
        & 60.36$_{\pm 1.74}$ \\

        & CoT-UQ-ME~\citep{Zhang2025CoT-UQ}
        & 47.96$_{\pm 1.68}$ & 28.62$_{\pm 1.72}$ & 28.10$_{\pm 1.73}$
        & 23.22$_{\pm 1.21}$ & 92.90$_{\pm 1.49}$ & 28.23$_{\pm 0.47}$
        & 71.83$_{\pm 1.73}$ \\

        & CoT-UQ-MI~\citep{Zhang2025CoT-UQ}
        & 50.07$_{\pm 1.61}$ & 30.98$_{\pm 1.89}$ & 30.29$_{\pm 1.76}$
        & 29.10$_{\pm 1.22}$ & 65.26$_{\pm 16.17}$ & 29.15$_{\pm 1.20}$
        & 69.65$_{\pm 1.76}$ \\

        & CoT-UQ-SAR~\citep{Zhang2025CoT-UQ}
        & 54.46$_{\pm 1.54}$ & 33.26$_{\pm 1.92}$ & 32.45$_{\pm 1.77}$
        & 13.33$_{\pm 1.21}$ & 43.79$_{\pm 15.46}$ & 22.66$_{\pm 0.79}$
        & 67.48$_{\pm 1.77}$ \\

        & CoT-UQ-SP~\citep{Zhang2025CoT-UQ}
        & 54.35$_{\pm 1.18}$ & 31.51$_{\pm 1.40}$ & 31.58$_{\pm 1.71}$
        & 58.23$_{\pm 1.31}$ & 68.92$_{\pm 3.23}$ & 57.34$_{\pm 1.16}$
        & 68.35$_{\pm 1.71}$ \\

        & Topo-UQ~\citep{Da2025Topo-uq}
        & 43.66$_{\pm 1.65}$ & 25.55$_{\pm 1.35}$ & 24.15$_{\pm 1.51}$
        & 45.66$_{\pm 1.30}$ & 81.27$_{\pm 6.02}$ & 44.66$_{\pm 0.78}$
        & 75.79$_{\pm 1.51}$ \\
        \midrule
        \midrule
        \multirow{21}{*}{\textbf{GSM8K}}
        & MaxSeqProb~\citep{Vashurin2025LM-Polygraph}
        & 70.77$_{\pm 1.66}$ & 83.43$_{\pm 1.50}$ & 82.10$_{\pm 1.44}$
        & 69.44$_{\pm 1.37}$ & 82.35$_{\pm 4.57}$ & 68.90$_{\pm 1.36}$
        & 17.83$_{\pm 1.44}$ \\

        & MeanTokEnt~\citep{fomicheva2020PPL}
        & 65.68$_{\pm 1.79}$ & 80.30$_{\pm 1.63}$ & 79.41$_{\pm 1.54}$
        & 3.97$_{\pm 0.95}$ & 26.60$_{\pm 12.64}$ & 19.35$_{\pm 0.66}$
        & 20.51$_{\pm 1.54}$ \\

        & MaxTokProb~\citep{fomicheva2020PPL}
        & 50.13$_{\pm 0.13}$ & 70.53$_{\pm 1.37}$ & 70.63$_{\pm 1.85}$
        & 20.48$_{\pm 1.37}$ & 20.48$_{\pm 1.37}$ & 25.00$_{\pm 0.00}$
        & 29.29$_{\pm 1.85}$ \\

        & PPL~\citep{fomicheva2020PPL}
        & 64.35$_{\pm 1.78}$ & 79.37$_{\pm 1.65}$ & 78.58$_{\pm 1.58}$
        & 5.29$_{\pm 0.98}$ & 34.76$_{\pm 22.89}$ & 19.69$_{\pm 0.72}$
        & 21.34$_{\pm 1.58}$ \\

        & BSDetector~\citep{chen2024BSDETECTOR}
        & 92.67$_{\pm 0.82}$ & 95.85$_{\pm 0.62}$ & 92.00$_{\pm 0.83}$
        & 14.09$_{\pm 1.02}$ & 46.64$_{\pm 2.77}$ & 13.32$_{\pm 0.64}$
        & 7.92$_{\pm 0.83}$ \\

        & SelfCheckGPT~\citep{manakul2023selfcheckgpt}
        & 90.66$_{\pm 0.94}$ & 95.02$_{\pm 0.78}$ & 91.32$_{\pm 0.86}$
        & 9.32$_{\pm 0.91}$ & 39.90$_{\pm 14.51}$ & 12.80$_{\pm 0.59}$
        & 8.61$_{\pm 0.86}$ \\

        & NumSet~\citep{lin2024numset/eigv/deg/ecc}
        & 87.09$_{\pm 1.00}$ & 92.54$_{\pm 0.73}$ & 90.27$_{\pm 0.91}$
        & 14.68$_{\pm 1.13}$ & 38.29$_{\pm 4.94}$ & 16.32$_{\pm 0.79}$
        & 9.65$_{\pm 0.91}$ \\

        & EigV~\citep{lin2024numset/eigv/deg/ecc}
        & 88.45$_{\pm 0.99}$ & 94.17$_{\pm 0.68}$ & 90.75$_{\pm 0.89}$
        & 4.45$_{\pm 0.68}$ & 14.98$_{\pm 4.34}$ & 12.12$_{\pm 0.59}$
        & 9.17$_{\pm 0.89}$ \\

        & Deg~\citep{lin2024numset/eigv/deg/ecc}
        & 88.39$_{\pm 0.99}$ & 93.91$_{\pm 0.68}$ & 90.72$_{\pm 0.88}$
        & 4.45$_{\pm 0.68}$ & 14.98$_{\pm 4.34}$ & 12.12$_{\pm 0.59}$
        & 9.20$_{\pm 0.88}$ \\

        & Ecc~\citep{lin2024numset/eigv/deg/ecc}
        & 22.74$_{\pm 1.36}$ & 62.13$_{\pm 1.40}$ & 49.84$_{\pm 1.93}$
        & 65.59$_{\pm 1.40}$ & 99.13$_{\pm 4.28}$ & 67.03$_{\pm 1.24}$
        & 50.08$_{\pm 1.93}$ \\

        & ConU~\citep{wang2024ConU}
        & 88.57$_{\pm 0.98}$ & 94.57$_{\pm 0.68}$ & 90.93$_{\pm 0.83}$
        & 8.03$_{\pm 1.01}$ & 24.13$_{\pm 3.88}$ & 13.10$_{\pm 0.62}$
        & 8.99$_{\pm 0.83}$ \\

        & Semantic Entropy~\citep{farquhar2024SemanticEntropy}
        & 74.16$_{\pm 1.47}$ & 85.54$_{\pm 1.24}$ & 83.91$_{\pm 1.26}$
        & 22.24$_{\pm 1.24}$ & 45.51$_{\pm 10.35}$ & 23.69$_{\pm 0.87}$
        & 16.01$_{\pm 1.26}$ \\

        & Semantic Density~\citep{Qiu2024SemanticDensity}
        & 75.64$_{\pm 1.48}$ & 85.97$_{\pm 1.23}$ & 84.25$_{\pm 1.27}$
        & 10.79$_{\pm 1.11}$ & 28.07$_{\pm 7.54}$ & 18.19$_{\pm 0.71}$
        & 15.67$_{\pm 1.27}$ \\

        & MARS~\citep{Bakman2024MARS}
        & 66.46$_{\pm 1.72}$ & 81.06$_{\pm 1.59}$ & 80.04$_{\pm 1.53}$
        & 4.06$_{\pm 1.02}$ & 34.59$_{\pm 20.09}$ & 19.34$_{\pm 0.67}$
        & 19.89$_{\pm 1.53}$ \\

        & Verbalized Probability~\citep{Lin2022Verbalized_Probability}
        & 57.46$_{\pm 1.71}$ & 75.85$_{\pm 1.73}$ & 75.37$_{\pm 1.67}$
        & 26.56$_{\pm 1.24}$ & 63.55$_{\pm 11.88}$ & 27.70$_{\pm 1.20}$
        & 24.56$_{\pm 1.67}$ \\

        & $P(\text{True})$~\citep{Kadavath2022Ptrue/PIK}
        & 69.09$_{\pm 1.82}$ & 80.95$_{\pm 1.69}$ & 80.01$_{\pm 1.60}$
        & 27.70$_{\pm 1.34}$ & 84.40$_{\pm 3.17}$ & 27.53$_{\pm 1.33}$
        & 19.92$_{\pm 1.60}$ \\

        & CoT-UQ-ME~\citep{Zhang2025CoT-UQ}
        & 56.86$_{\pm 1.97}$ & 76.17$_{\pm 1.86}$ & 75.63$_{\pm 1.82}$
        & 16.32$_{\pm 1.55}$ & 83.74$_{\pm 13.68}$ & 23.16$_{\pm 1.12}$
        & 24.26$_{\pm 1.82}$ \\

        & CoT-UQ-MI~\citep{Zhang2025CoT-UQ}
        & 55.87$_{\pm 2.09}$ & 75.67$_{\pm 1.95}$ & 75.16$_{\pm 1.90}$
        & 17.17$_{\pm 1.41}$ & 59.72$_{\pm 11.19}$ & 24.72$_{\pm 0.92}$
        & 24.74$_{\pm 1.90}$ \\

        & CoT-UQ-SAR~\citep{Zhang2025CoT-UQ}
        & 56.64$_{\pm 1.97}$ & 76.23$_{\pm 1.86}$ & 75.68$_{\pm 1.82}$
        & 16.49$_{\pm 1.54}$ & 83.94$_{\pm 13.08}$ & 23.27$_{\pm 1.11}$
        & 24.21$_{\pm 1.82}$ \\

        & CoT-UQ-SP~\citep{Zhang2025CoT-UQ}
        & 50.57$_{\pm 2.02}$ & 71.62$_{\pm 1.98}$ & 71.71$_{\pm 2.08}$
        & 43.48$_{\pm 1.58}$ & 66.75$_{\pm 2.44}$ & 44.06$_{\pm 1.13}$
        & 28.18$_{\pm 2.08}$ \\

        & Topo-UQ~\citep{Da2025Topo-uq}
        & 45.80$_{\pm 1.79}$ & 69.93$_{\pm 1.76}$ & 69.57$_{\pm 1.78}$
        & 22.56$_{\pm 1.33}$ & 70.88$_{\pm 6.01}$ & 26.61$_{\pm 1.17}$
        & 30.35$_{\pm 1.78}$ \\
        \midrule
        \midrule
        \multirow{21}{*}{\textbf{HumanEval}}
        & MaxSeqProb~\citep{Vashurin2025LM-Polygraph}
        & 57.23$_{\pm 4.92}$ & 63.24$_{\pm 5.88}$ & 61.02$_{\pm 5.84}$
        & 58.99$_{\pm 3.97}$ & 86.48$_{\pm 19.36}$ & 58.89$_{\pm 3.97}$
        & 38.36$_{\pm 5.84}$ \\

        & MeanTokEnt~\citep{fomicheva2020PPL}
        & 49.00$_{\pm 4.47}$ & 56.84$_{\pm 5.01}$ & 54.16$_{\pm 5.29}$
        & 22.31$_{\pm 3.84}$ & 67.06$_{\pm 14.33}$ & 29.20$_{\pm 2.05}$
        & 45.22$_{\pm 5.29}$ \\

        & MaxTokProb~\citep{fomicheva2020PPL}
        & 47.97$_{\pm 1.01}$ & 57.42$_{\pm 4.05}$ & 56.73$_{\pm 5.46}$
        & 8.48$_{\pm 3.79}$ & 8.48$_{\pm 3.79}$ & 25.00$_{\pm 0.00}$
        & 42.65$_{\pm 5.46}$ \\

        & PPL~\citep{fomicheva2020PPL}
        & 48.07$_{\pm 4.40}$ & 57.08$_{\pm 5.11}$ & 54.73$_{\pm 5.34}$
        & 23.99$_{\pm 3.71}$ & 90.48$_{\pm 15.56}$ & 29.30$_{\pm 1.95}$
        & 44.64$_{\pm 5.34}$ \\

        & BSDetector~\citep{chen2024BSDETECTOR}
        & 58.17$_{\pm 4.97}$ & 66.31$_{\pm 5.31}$ & 63.87$_{\pm 5.25}$
        & 25.03$_{\pm 3.44}$ & 58.78$_{\pm 9.65}$ & 30.84$_{\pm 2.35}$
        & 35.51$_{\pm 5.25}$ \\

        & SelfCheckGPT~\citep{manakul2023selfcheckgpt}
        & 63.39$_{\pm 4.08}$ & 73.60$_{\pm 4.75}$ & 70.03$_{\pm 4.62}$
        & 25.80$_{\pm 3.76}$ & 78.57$_{\pm 25.13}$ & 29.72$_{\pm 2.71}$
        & 29.34$_{\pm 4.62}$ \\

        & NumSet~\citep{lin2024numset/eigv/deg/ecc}
        & 61.72$_{\pm 4.66}$ & 69.50$_{\pm 4.75}$ & 68.10$_{\pm 4.75}$
        & 22.88$_{\pm 3.44}$ & 56.28$_{\pm 9.91}$ & 28.93$_{\pm 2.47}$
        & 31.28$_{\pm 4.75}$ \\

        & EigV~\citep{lin2024numset/eigv/deg/ecc}
        & 60.27$_{\pm 4.73}$ & 68.66$_{\pm 4.86}$ & 66.76$_{\pm 4.81}$
        & 35.04$_{\pm 3.77}$ & 54.22$_{\pm 6.36}$ & 37.02$_{\pm 2.85}$
        & 32.62$_{\pm 4.81}$ \\

        & Deg~\citep{lin2024numset/eigv/deg/ecc}
        & 60.27$_{\pm 4.73}$ & 68.66$_{\pm 4.86}$ & 66.76$_{\pm 4.81}$
        & 35.04$_{\pm 3.77}$ & 54.22$_{\pm 6.36}$ & 37.02$_{\pm 2.85}$
        & 32.62$_{\pm 4.81}$ \\

        & Ecc~\citep{lin2024numset/eigv/deg/ecc}
        & 37.17$_{\pm 4.65}$ & 52.75$_{\pm 4.83}$ & 50.45$_{\pm 5.55}$
        & 40.83$_{\pm 4.14}$ & 69.62$_{\pm 5.72}$ & 43.90$_{\pm 3.11}$
        & 48.93$_{\pm 5.55}$ \\

        & ConU~\citep{wang2024ConU}
        & 59.65$_{\pm 4.72}$ & 68.51$_{\pm 4.96}$ & 65.64$_{\pm 4.86}$
        & 28.13$_{\pm 3.48}$ & 56.63$_{\pm 8.29}$ & 32.22$_{\pm 2.42}$
        & 33.74$_{\pm 4.86}$ \\

        & Semantic Entropy~\citep{farquhar2024SemanticEntropy}
        & 50.13$_{\pm 4.69}$ & 59.86$_{\pm 5.46}$ & 57.52$_{\pm 5.56}$
        & 41.66$_{\pm 3.92}$ & 87.23$_{\pm 19.57}$ & 42.56$_{\pm 2.86}$
        & 41.86$_{\pm 5.56}$ \\

        & Semantic Density~\citep{Qiu2024SemanticDensity}
        & 48.55$_{\pm 4.63}$ & 60.44$_{\pm 5.51}$ & 58.19$_{\pm 5.76}$
        & 16.48$_{\pm 3.18}$ & 90.35$_{\pm 13.70}$ & 27.22$_{\pm 1.24}$
        & 41.19$_{\pm 5.76}$ \\

        & MARS~\citep{Bakman2024MARS}
        & 56.35$_{\pm 4.41}$ & 63.42$_{\pm 5.10}$ & 61.29$_{\pm 5.14}$
        & 17.55$_{\pm 3.12}$ & 46.00$_{\pm 16.14}$ & 27.00$_{\pm 1.91}$
        & 38.09$_{\pm 5.14}$ \\

        & Verbalized Probability~\citep{Lin2022Verbalized_Probability}
        & 49.94$_{\pm 4.68}$ & 61.73$_{\pm 5.60}$ & 59.44$_{\pm 5.52}$
        & 24.54$_{\pm 3.75}$ & 93.57$_{\pm 9.43}$ & 30.88$_{\pm 2.18}$
        & 39.94$_{\pm 5.52}$ \\

        & $P(\text{True})$~\citep{Kadavath2022Ptrue/PIK}
        & 58.50$_{\pm 4.95}$ & 68.51$_{\pm 5.50}$ & 65.79$_{\pm 5.27}$
        & 44.23$_{\pm 3.99}$ & 76.44$_{\pm 6.41}$ & 43.35$_{\pm 3.97}$
        & 33.59$_{\pm 5.27}$ \\

        & CoT-UQ-ME~\citep{Zhang2025CoT-UQ}
        & 39.68$_{\pm 4.38}$ & 53.26$_{\pm 4.76}$ & 50.49$_{\pm 5.22}$
        & 28.75$_{\pm 3.66}$ & 84.95$_{\pm 10.12}$ & 33.72$_{\pm 2.41}$
        & 48.88$_{\pm 5.22}$ \\

        & CoT-UQ-MI~\citep{Zhang2025CoT-UQ}
        & 62.84$_{\pm 4.60}$ & 74.54$_{\pm 4.40}$ & 70.60$_{\pm 4.37}$
        & 54.62$_{\pm 3.91}$ & 84.14$_{\pm 5.11}$ & 53.28$_{\pm 3.84}$
        & 28.78$_{\pm 4.37}$ \\

        & CoT-UQ-SAR~\citep{Zhang2025CoT-UQ}
        & 47.14$_{\pm 5.21}$ & 58.31$_{\pm 5.43}$ & 56.03$_{\pm 5.72}$
        & 38.65$_{\pm 4.05}$ & 59.38$_{\pm 7.34}$ & 40.52$_{\pm 2.82}$
        & 43.35$_{\pm 5.72}$ \\

        & CoT-UQ-SP~\citep{Zhang2025CoT-UQ}
        & 58.12$_{\pm 3.99}$ & 63.27$_{\pm 4.62}$ & 62.67$_{\pm 5.31}$
        & 36.66$_{\pm 3.94}$ & 84.42$_{\pm 14.49}$ & 37.62$_{\pm 3.82}$
        & 36.71$_{\pm 5.31}$ \\

        & Topo-UQ~\citep{Da2025Topo-uq}
        & 49.44$_{\pm 4.65}$ & 61.42$_{\pm 4.95}$ & 59.28$_{\pm 4.99}$
        & 36.05$_{\pm 3.78}$ & 87.75$_{\pm 16.53}$ & 36.78$_{\pm 3.24}$
        & 40.10$_{\pm 4.99}$ \\
        \bottomrule
    \end{tabular}
    }
\end{table*}
\FloatBarrier

\begin{table*}[t]
    \centering
    \caption{Performance Comparison of the investigated LLM UQ methods of Qwen3-4B in reasoning-augmented generation across three evaluated benchmarks.}
    \label{qwen3_4b_answer_reasoning_results}
    \resizebox{0.7\textwidth}{!}{
    \begin{tabular}{llccccccc}
        \toprule
        \multirow{2}{*}{\textbf{Datasets}}
        & \multirow{2}{*}{\textbf{Methods}}
        & \multicolumn{3}{c}{\textbf{Ranking}}
        & \multicolumn{3}{c}{\textbf{Calibration}}
        & \textbf{Coverage} \\
        \cmidrule(lr){3-5} \cmidrule(lr){6-8} \cmidrule(lr){9-9}
        & & \textbf{AUROC} & \textbf{AUPRC} & \textbf{AUARC}
        & \textbf{ECE} & \textbf{MCE} & \textbf{BS}
        & \textbf{AURC} \\
        \midrule
        \multirow{21}{*}{\textbf{TriviaQA}}
        & MaxSeqProb~\citep{Vashurin2025LM-Polygraph}
        & 65.42$_{\pm 1.30}$ & 60.17$_{\pm 1.95}$ & 57.44$_{\pm 1.70}$
        & 44.12$_{\pm 1.32}$ & 68.64$_{\pm 9.01}$ & 43.83$_{\pm 1.30}$
        & 42.49$_{\pm 1.70}$ \\

        & MeanTokEnt~\citep{fomicheva2020PPL}
        & 64.75$_{\pm 1.36}$ & 59.77$_{\pm 1.94}$ & 57.10$_{\pm 1.69}$
        & 17.00$_{\pm 1.23}$ & 25.31$_{\pm 3.30}$ & 26.01$_{\pm 0.48}$
        & 42.83$_{\pm 1.69}$ \\

        & MaxTokProb~\citep{fomicheva2020PPL}
        & 47.87$_{\pm 1.00}$ & 44.11$_{\pm 1.43}$ & 44.22$_{\pm 1.96}$
        & 4.78$_{\pm 1.32}$ & 4.78$_{\pm 1.32}$ & 25.00$_{\pm 0.00}$
        & 55.71$_{\pm 1.96}$ \\

        & PPL~\citep{fomicheva2020PPL}
        & 63.83$_{\pm 1.38}$ & 58.52$_{\pm 2.01}$ & 56.20$_{\pm 1.75}$
        & 30.59$_{\pm 1.29}$ & 35.73$_{\pm 2.24}$ & 32.81$_{\pm 0.75}$
        & 43.73$_{\pm 1.75}$ \\

        & BSDetector~\citep{chen2024BSDETECTOR}
        & 77.27$_{\pm 1.05}$ & 72.29$_{\pm 1.75}$ & 65.72$_{\pm 1.54}$
        & 29.04$_{\pm 1.09}$ & 51.94$_{\pm 8.03}$ & 29.82$_{\pm 1.01}$
        & 34.22$_{\pm 1.54}$ \\

        & SelfCheckGPT~\citep{manakul2023selfcheckgpt}
        & 85.14$_{\pm 1.05}$ & 80.82$_{\pm 1.69}$ & 70.89$_{\pm 1.48}$
        & 36.33$_{\pm 1.13}$ & 60.01$_{\pm 2.82}$ & 31.11$_{\pm 0.85}$
        & 29.05$_{\pm 1.48}$ \\

        & NumSet~\citep{lin2024numset/eigv/deg/ecc}
        & 87.81$_{\pm 0.85}$ & 81.24$_{\pm 1.36}$ & 72.16$_{\pm 1.39}$
        & 30.83$_{\pm 1.06}$ & 55.32$_{\pm 2.47}$ & 25.62$_{\pm 0.76}$
        & 27.77$_{\pm 1.39}$ \\

        & EigV~\citep{lin2024numset/eigv/deg/ecc}
        & 88.17$_{\pm 0.88}$ & 82.71$_{\pm 1.39}$ & 72.39$_{\pm 1.45}$
        & 13.13$_{\pm 0.94}$ & 41.97$_{\pm 22.73}$ & 15.26$_{\pm 0.64}$
        & 27.55$_{\pm 1.45}$ \\

        & Deg~\citep{lin2024numset/eigv/deg/ecc}
        & 88.17$_{\pm 0.88}$ & 82.71$_{\pm 1.39}$ & 72.39$_{\pm 1.45}$
        & 13.13$_{\pm 0.94}$ & 41.97$_{\pm 22.73}$ & 15.26$_{\pm 0.64}$
        & 27.55$_{\pm 1.45}$ \\

        & Ecc~\citep{lin2024numset/eigv/deg/ecc}
        & 19.02$_{\pm 0.96}$ & 38.75$_{\pm 1.19}$ & 21.58$_{\pm 1.29}$
        & 35.93$_{\pm 1.10}$ & 100.00$_{\pm 0.00}$ & 33.13$_{\pm 0.37}$
        & 78.35$_{\pm 1.29}$ \\

        & ConU~\citep{wang2024ConU}
        & 87.30$_{\pm 0.90}$ & 81.61$_{\pm 1.50}$ & 71.35$_{\pm 1.40}$
        & 21.60$_{\pm 1.00}$ & 40.89$_{\pm 3.63}$ & 19.11$_{\pm 0.69}$
        & 28.58$_{\pm 1.40}$ \\

        & Semantic Entropy~\citep{farquhar2024SemanticEntropy}
        & 86.20$_{\pm 0.98}$ & 81.96$_{\pm 1.44}$ & 71.55$_{\pm 1.44}$
        & 8.21$_{\pm 0.94}$ & 22.08$_{\pm 6.03}$ & 14.62$_{\pm 0.74}$
        & 28.38$_{\pm 1.44}$ \\

        & Semantic Density~\citep{Qiu2024SemanticDensity}
        & 87.73$_{\pm 0.89}$ & 82.74$_{\pm 1.46}$ & 72.08$_{\pm 1.44}$
        & 17.38$_{\pm 0.99}$ & 38.80$_{\pm 4.71}$ & 17.24$_{\pm 0.68}$
        & 27.85$_{\pm 1.44}$ \\

        & MARS~\citep{Bakman2024MARS}
        & 64.88$_{\pm 1.48}$ & 58.61$_{\pm 2.22}$ & 56.33$_{\pm 1.92}$
        & 35.75$_{\pm 1.27}$ & 77.38$_{\pm 28.62}$ & 36.01$_{\pm 0.88}$
        & 43.61$_{\pm 1.92}$ \\

        & Verbalized Probability~\citep{Lin2022Verbalized_Probability}
        & 74.08$_{\pm 1.09}$ & 67.70$_{\pm 1.89}$ & 62.85$_{\pm 1.63}$
        & 29.00$_{\pm 1.29}$ & 47.76$_{\pm 3.74}$ & 30.18$_{\pm 0.72}$
        & 37.08$_{\pm 1.63}$ \\
        
        & $P(\text{True})$~\citep{Kadavath2022Ptrue/PIK}
        & 69.50$_{\pm 1.30}$ & 62.70$_{\pm 2.04}$ & 59.61$_{\pm 1.76}$
        & 44.81$_{\pm 1.31}$ & 74.53$_{\pm 5.23}$ & 44.37$_{\pm 1.28}$
        & 40.32$_{\pm 1.76}$ \\

        & CoT-UQ-ME~\citep{Zhang2025CoT-UQ}
        & 52.54$_{\pm 1.36}$ & 48.05$_{\pm 1.73}$ & 47.54$_{\pm 1.67}$
        & 13.39$_{\pm 1.17}$ & 45.18$_{\pm 8.12}$ & 27.44$_{\pm 0.43}$
        & 52.40$_{\pm 1.67}$ \\

        & CoT-UQ-MI~\citep{Zhang2025CoT-UQ}
        & 56.00$_{\pm 1.51}$ & 50.25$_{\pm 1.93}$ & 49.48$_{\pm 1.82}$
        & 44.70$_{\pm 1.32}$ & 65.29$_{\pm 7.19}$ & 44.68$_{\pm 1.31}$
        & 50.45$_{\pm 1.82}$ \\

        & CoT-UQ-SAR~\citep{Zhang2025CoT-UQ}
        & 53.65$_{\pm 1.54}$ & 50.34$_{\pm 2.11}$ & 49.47$_{\pm 1.95}$
        & 29.16$_{\pm 1.32}$ & 86.13$_{\pm 21.60}$ & 33.54$_{\pm 0.91}$
        & 50.46$_{\pm 1.95}$ \\

        & CoT-UQ-SP~\citep{Zhang2025CoT-UQ}
        & 52.33$_{\pm 1.33}$ & 46.40$_{\pm 1.59}$ & 46.27$_{\pm 1.88}$
        & 26.25$_{\pm 1.37}$ & 64.14$_{\pm 8.14}$ & 35.98$_{\pm 0.87}$
        & 53.67$_{\pm 1.88}$ \\

        & Topo-UQ~\citep{Da2025Topo-uq}
        & 44.58$_{\pm 1.44}$ & 40.25$_{\pm 1.52}$ & 38.80$_{\pm 1.71}$
        & 29.43$_{\pm 1.32}$ & 75.11$_{\pm 13.14}$ & 35.42$_{\pm 0.76}$
        & 61.13$_{\pm 1.71}$ \\
        \midrule
        \midrule
        \multirow{21}{*}{\textbf{GSM8K}}
        & MaxSeqProb~\citep{Vashurin2025LM-Polygraph}
        & 66.41$_{\pm 1.95}$ & 89.03$_{\pm 1.13}$ & 88.58$_{\pm 1.12}$
        & 82.41$_{\pm 1.01}$ & 94.82$_{\pm 7.68}$ & 82.13$_{\pm 1.01}$
        & 11.35$_{\pm 1.12}$ \\

        & MeanTokEnt~\citep{fomicheva2020PPL}
        & 60.89$_{\pm 2.22}$ & 85.52$_{\pm 1.41}$ & 85.25$_{\pm 1.43}$
        & 11.37$_{\pm 0.95}$ & 25.96$_{\pm 3.86}$ & 14.81$_{\pm 0.50}$
        & 14.67$_{\pm 1.43}$ \\

        & MaxTokProb~\citep{fomicheva2020PPL}
        & 51.48$_{\pm 0.59}$ & 83.12$_{\pm 1.00}$ & 83.25$_{\pm 1.39}$
        & 17.23$_{\pm 1.00}$ & 17.24$_{\pm 1.00}$ & 17.23$_{\pm 1.00}$
        & 16.67$_{\pm 1.39}$ \\

        & PPL~\citep{fomicheva2020PPL}
        & 60.09$_{\pm 2.29}$ & 85.96$_{\pm 1.32}$ & 85.74$_{\pm 1.31}$
        & 5.72$_{\pm 0.94}$ & 35.02$_{\pm 15.32}$ & 13.93$_{\pm 0.62}$
        & 14.19$_{\pm 1.31}$ \\

        & BSDetector~\citep{chen2024BSDETECTOR}
        & 93.03$_{\pm 1.04}$ & 97.77$_{\pm 0.43}$ & 96.44$_{\pm 0.51}$
        & 5.80$_{\pm 0.65}$ & 44.15$_{\pm 6.20}$ & 7.41$_{\pm 0.47}$
        & 3.48$_{\pm 0.51}$ \\

        & SelfCheckGPT~\citep{manakul2023selfcheckgpt}
        & 90.31$_{\pm 1.08}$ & 97.35$_{\pm 0.40}$ & 96.07$_{\pm 0.43}$
        & 4.85$_{\pm 0.70}$ & 24.22$_{\pm 4.34}$ & 8.50$_{\pm 0.43}$
        & 3.85$_{\pm 0.43}$ \\

        & NumSet~\citep{lin2024numset/eigv/deg/ecc}
        & 86.01$_{\pm 1.12}$ & 95.56$_{\pm 0.47}$ & 95.12$_{\pm 0.57}$
        & 6.88$_{\pm 0.84}$ & 28.50$_{\pm 9.62}$ & 11.00$_{\pm 0.59}$
        & 4.80$_{\pm 0.57}$ \\

        & EigV~\citep{lin2024numset/eigv/deg/ecc}
        & 87.59$_{\pm 1.13}$ & 96.56$_{\pm 0.45}$ & 95.44$_{\pm 0.55}$
        & 5.09$_{\pm 0.78}$ & 28.60$_{\pm 8.37}$ & 10.27$_{\pm 0.51}$
        & 4.48$_{\pm 0.55}$ \\

        & Deg~\citep{lin2024numset/eigv/deg/ecc}
        & 87.21$_{\pm 1.15}$ & 96.28$_{\pm 0.46}$ & 95.36$_{\pm 0.56}$
        & 5.09$_{\pm 0.78}$ & 28.60$_{\pm 8.37}$ & 10.27$_{\pm 0.51}$
        & 4.56$_{\pm 0.56}$ \\

        & Ecc~\citep{lin2024numset/eigv/deg/ecc}
        & 23.22$_{\pm 1.67}$ & 76.06$_{\pm 1.24}$ & 67.54$_{\pm 1.93}$
        & 76.36$_{\pm 1.15}$ & 97.52$_{\pm 3.27}$ & 76.70$_{\pm 0.94}$
        & 32.38$_{\pm 1.93}$ \\

        & ConU~\citep{wang2024ConU}
        & 87.30$_{\pm 1.21}$ & 96.54$_{\pm 0.52}$ & 95.36$_{\pm 0.57}$
        & 4.42$_{\pm 0.73}$ & 24.46$_{\pm 6.77}$ & 10.06$_{\pm 0.55}$
        & 4.57$_{\pm 0.57}$ \\

        & Semantic Entropy~\citep{farquhar2024SemanticEntropy}
        & 73.46$_{\pm 1.65}$ & 92.15$_{\pm 0.80}$ & 91.51$_{\pm 0.85}$
        & 28.76$_{\pm 1.09}$ & 57.01$_{\pm 3.16}$ & 25.64$_{\pm 0.86}$
        & 8.42$_{\pm 0.85}$ \\

        & Semantic Density~\citep{Qiu2024SemanticDensity}
        & 75.16$_{\pm 1.64}$ & 92.46$_{\pm 0.79}$ & 91.78$_{\pm 0.81}$
        & 15.29$_{\pm 0.99}$ & 49.12$_{\pm 6.94}$ & 16.30$_{\pm 0.61}$
        & 8.14$_{\pm 0.81}$ \\

        & MARS~\citep{Bakman2024MARS}
        & 59.45$_{\pm 2.27}$ & 85.43$_{\pm 1.38}$ & 85.21$_{\pm 1.38}$
        & 11.53$_{\pm 0.88}$ & 44.83$_{\pm 22.47}$ & 15.36$_{\pm 0.52}$
        & 14.72$_{\pm 1.38}$ \\

        & Verbalized Probability~\citep{Lin2022Verbalized_Probability}
        & 72.68$_{\pm 1.89}$ & 91.61$_{\pm 1.01}$ & 90.99$_{\pm 0.98}$
        & 6.59$_{\pm 0.95}$ & 58.90$_{\pm 21.39}$ & 13.14$_{\pm 0.69}$
        & 8.94$_{\pm 0.98}$ \\

        & $P(\text{True})$~\citep{Kadavath2022Ptrue/PIK}
        & 71.49$_{\pm 1.82}$ & 91.41$_{\pm 0.99}$ & 90.80$_{\pm 0.96}$
        & 37.68$_{\pm 1.27}$ & 76.49$_{\pm 5.13}$ & 37.53$_{\pm 1.26}$
        & 9.12$_{\pm 0.96}$ \\

        & CoT-UQ-ME~\citep{Zhang2025CoT-UQ}
        & 55.72$_{\pm 2.16}$ & 87.79$_{\pm 1.22}$ & 87.54$_{\pm 1.23}$
        & 7.64$_{\pm 0.90}$ & 89.10$_{\pm 15.28}$ & 13.62$_{\pm 0.75}$
        & 12.38$_{\pm 1.23}$ \\

        & CoT-UQ-MI~\citep{Zhang2025CoT-UQ}
        & 57.04$_{\pm 2.20}$ & 88.21$_{\pm 1.20}$ & 87.93$_{\pm 1.21}$
        & 21.46$_{\pm 1.27}$ & 82.67$_{\pm 10.58}$ & 20.32$_{\pm 0.61}$
        & 11.99$_{\pm 1.21}$ \\

        & CoT-UQ-SAR~\citep{Zhang2025CoT-UQ}
        & 55.82$_{\pm 2.19}$ & 87.76$_{\pm 1.24}$ & 87.50$_{\pm 1.25}$
        & 7.43$_{\pm 0.90}$ & 89.19$_{\pm 15.49}$ & 13.43$_{\pm 0.76}$
        & 12.42$_{\pm 1.25}$ \\

        & CoT-UQ-SP~\citep{Zhang2025CoT-UQ}
        & 51.35$_{\pm 2.26}$ & 85.19$_{\pm 1.38}$ & 84.99$_{\pm 1.46}$
        & 57.11$_{\pm 1.12}$ & 78.11$_{\pm 1.72}$ & 50.59$_{\pm 0.91}$
        & 14.93$_{\pm 1.46}$ \\

        & Topo-UQ~\citep{Da2025Topo-uq}
        & 61.95$_{\pm 2.16}$ & 88.17$_{\pm 1.06}$ & 87.81$_{\pm 1.05}$
        & 9.72$_{\pm 0.85}$ & 45.01$_{\pm 16.99}$ & 14.46$_{\pm 0.78}$
        & 12.11$_{\pm 1.05}$ \\
        \midrule
        \midrule
        \multirow{21}{*}{\textbf{HumanEval}}
        & MaxSeqProb~\citep{Vashurin2025LM-Polygraph}
        & 42.40$_{\pm 4.64}$ & 61.42$_{\pm 4.89}$ & 58.68$_{\pm 5.45}$
        & 66.20$_{\pm 3.45}$ & 89.85$_{\pm 13.46}$ & 65.76$_{\pm 3.37}$
        & 40.70$_{\pm 5.45}$ \\

        & MeanTokEnt~\citep{fomicheva2020PPL}
        & 56.64$_{\pm 4.74}$ & 72.83$_{\pm 4.75}$ & 70.90$_{\pm 4.72}$
        & 15.54$_{\pm 3.35}$ & 84.03$_{\pm 8.77}$ & 24.69$_{\pm 1.83}$
        & 28.47$_{\pm 4.72}$ \\

        & MaxTokProb~\citep{fomicheva2020PPL}
        & 51.39$_{\pm 1.92}$ & 67.11$_{\pm 3.65}$ & 66.34$_{\pm 5.21}$
        & 16.47$_{\pm 3.55}$ & 16.47$_{\pm 3.55}$ & 25.00$_{\pm 0.00}$
        & 33.04$_{\pm 5.21}$ \\

        & PPL~\citep{fomicheva2020PPL}
        & 53.99$_{\pm 4.48}$ & 70.51$_{\pm 4.66}$ & 68.76$_{\pm 4.68}$
        & 16.74$_{\pm 3.13}$ & 84.86$_{\pm 8.29}$ & 25.50$_{\pm 2.14}$
        & 30.62$_{\pm 4.68}$ \\

        & BSDetector~\citep{chen2024BSDETECTOR}
        & 45.31$_{\pm 4.99}$ & 63.09$_{\pm 5.31}$ & 60.56$_{\pm 5.86}$
        & 49.48$_{\pm 3.54}$ & 71.98$_{\pm 6.33}$ & 48.78$_{\pm 2.91}$
        & 38.82$_{\pm 5.86}$ \\

        & SelfCheckGPT~\citep{manakul2023selfcheckgpt}
        & 53.41$_{\pm 5.04}$ & 68.36$_{\pm 5.17}$ & 66.26$_{\pm 5.34}$
        & 22.19$_{\pm 3.45}$ & 36.89$_{\pm 11.26}$ & 26.68$_{\pm 2.68}$
        & 33.12$_{\pm 5.34}$ \\

        & NumSet~\citep{lin2024numset/eigv/deg/ecc}
        & 45.72$_{\pm 5.21}$ & 63.12$_{\pm 4.92}$ & 60.90$_{\pm 5.76}$
        & 33.79$_{\pm 3.66}$ & 71.53$_{\pm 7.30}$ & 35.78$_{\pm 2.56}$
        & 38.48$_{\pm 5.76}$ \\

        & EigV~\citep{lin2024numset/eigv/deg/ecc}
        & 44.94$_{\pm 5.13}$ & 62.97$_{\pm 5.18}$ & 61.04$_{\pm 5.80}$
        & 47.03$_{\pm 3.62}$ & 70.46$_{\pm 6.29}$ & 47.48$_{\pm 2.65}$
        & 38.34$_{\pm 5.80}$ \\

        & Deg~\citep{lin2024numset/eigv/deg/ecc}
        & 44.94$_{\pm 5.13}$ & 62.97$_{\pm 5.18}$ & 61.04$_{\pm 5.80}$
        & 47.03$_{\pm 3.62}$ & 70.46$_{\pm 6.29}$ & 47.48$_{\pm 2.65}$
        & 38.34$_{\pm 5.80}$ \\

        & Ecc~\citep{lin2024numset/eigv/deg/ecc}
        & 53.49$_{\pm 5.35}$ & 66.66$_{\pm 4.37}$ & 64.40$_{\pm 4.97}$
        & 40.26$_{\pm 3.74}$ & 65.31$_{\pm 8.03}$ & 39.44$_{\pm 2.85}$
        & 34.98$_{\pm 4.97}$ \\

        & ConU~\citep{wang2024ConU}
        & 44.47$_{\pm 5.11}$ & 63.22$_{\pm 5.37}$ & 61.43$_{\pm 5.72}$
        & 39.51$_{\pm 3.53}$ & 74.31$_{\pm 6.76}$ & 40.50$_{\pm 2.36}$
        & 37.95$_{\pm 5.72}$ \\

        & Semantic Entropy~\citep{farquhar2024SemanticEntropy}
        & 41.10$_{\pm 4.77}$ & 63.45$_{\pm 5.03}$ & 61.35$_{\pm 5.42}$
        & 52.47$_{\pm 3.70}$ & 77.79$_{\pm 8.62}$ & 52.14$_{\pm 2.95}$
        & 38.03$_{\pm 5.42}$ \\

        & Semantic Density~\citep{Qiu2024SemanticDensity}
        & 52.65$_{\pm 5.03}$ & 70.23$_{\pm 4.55}$ & 68.42$_{\pm 4.56}$
        & 20.09$_{\pm 3.25}$ & 83.90$_{\pm 24.54}$ & 26.81$_{\pm 1.10}$
        & 30.96$_{\pm 4.56}$ \\

        & MARS~\citep{Bakman2024MARS}
        & 51.57$_{\pm 5.10}$ & 68.59$_{\pm 5.40}$ & 66.71$_{\pm 5.53}$
        & 20.23$_{\pm 3.26}$ & 90.64$_{\pm 15.54}$ & 26.84$_{\pm 2.43}$
        & 32.67$_{\pm 5.53}$ \\

        & Verbalized Probability~\citep{Lin2022Verbalized_Probability}
        & 54.41$_{\pm 5.12}$ & 65.68$_{\pm 5.22}$ & 62.38$_{\pm 5.79}$
        & 19.74$_{\pm 3.30}$ & 64.75$_{\pm 13.76}$ & 25.91$_{\pm 2.35}$
        & 37.00$_{\pm 5.79}$ \\

        & $P(\text{True})$~\citep{Kadavath2022Ptrue/PIK}
        & 54.69$_{\pm 5.19}$ & 66.30$_{\pm 5.23}$ & 63.27$_{\pm 5.74}$
        & 59.70$_{\pm 3.54}$ & 66.92$_{\pm 3.97}$ & 59.18$_{\pm 3.56}$
        & 36.11$_{\pm 5.74}$ \\

        & CoT-UQ-ME~\citep{Zhang2025CoT-UQ}
        & 62.36$_{\pm 5.25}$ & 72.03$_{\pm 4.48}$ & 69.82$_{\pm 4.59}$
        & 14.44$_{\pm 2.77}$ & 38.26$_{\pm 8.93}$ & 22.30$_{\pm 1.97}$
        & 29.56$_{\pm 4.59}$ \\

        & CoT-UQ-MI~\citep{Zhang2025CoT-UQ}
        & 53.14$_{\pm 5.46}$ & 68.38$_{\pm 5.11}$ & 66.52$_{\pm 5.31}$
        & 64.63$_{\pm 3.55}$ & 70.25$_{\pm 7.20}$ & 64.01$_{\pm 3.53}$
        & 32.85$_{\pm 5.31}$ \\

        & CoT-UQ-SAR~\citep{Zhang2025CoT-UQ}
        & 55.94$_{\pm 4.72}$ & 69.71$_{\pm 4.76}$ & 67.82$_{\pm 4.94}$
        & 38.86$_{\pm 3.54}$ & 54.26$_{\pm 6.57}$ & 37.93$_{\pm 1.98}$
        & 31.56$_{\pm 4.94}$ \\

        & CoT-UQ-SP~\citep{Zhang2025CoT-UQ}
        & 53.62$_{\pm 4.38}$ & 68.38$_{\pm 4.44}$ & 67.99$_{\pm 5.37}$
        & 29.73$_{\pm 3.91}$ & 62.31$_{\pm 12.77}$ & 32.04$_{\pm 3.48}$
        & 31.39$_{\pm 5.37}$ \\

        & Topo-UQ~\citep{Da2025Topo-uq}
        & 48.04$_{\pm 4.74}$ & 67.29$_{\pm 4.50}$ & 65.39$_{\pm 4.66}$
        & 23.72$_{\pm 3.33}$ & 75.66$_{\pm 16.19}$ & 27.94$_{\pm 2.39}$
        & 33.98$_{\pm 4.66}$ \\
        \bottomrule
    \end{tabular}
    }
\end{table*}
\FloatBarrier

\begin{table*}[t]
    \centering
    \caption{Performance Comparison of the investigated LLM UQ methods of Llama-3.2-1B in reasoning-augmented generation across three evaluated benchmarks.}
    \label{Llama-3.2-1B_answer_reasoning_results}
    \resizebox{0.7\textwidth}{!}{
    \begin{tabular}{llccccccc}
        \toprule
        \multirow{2}{*}{\textbf{Datasets}}
        & \multirow{2}{*}{\textbf{Methods}}
        & \multicolumn{3}{c}{\textbf{Ranking}}
        & \multicolumn{3}{c}{\textbf{Calibration}}
        & \textbf{Coverage} \\
        \cmidrule(lr){3-5} \cmidrule(lr){6-8} \cmidrule(lr){9-9}
        & & \textbf{AUROC} & \textbf{AUPRC} & \textbf{AUARC}
        & \textbf{ECE} & \textbf{MCE} & \textbf{BS}
        & \textbf{AURC} \\
        \midrule
        \multirow{21}{*}{\textbf{TriviaQA}}
        & MaxSeqProb~\citep{Vashurin2025LM-Polygraph}
        & 66.71$_{\pm 1.46}$ & 44.13$_{\pm 2.49}$ & 39.31$_{\pm 1.91}$
        & 27.30$_{\pm 1.23}$ & 75.95$_{\pm 2.49}$ & 27.18$_{\pm 1.22}$
        & 60.62$_{\pm 1.91}$ \\

        & MeanTokEnt~\citep{fomicheva2020PPL}
        & 62.30$_{\pm 1.46}$ & 35.81$_{\pm 2.24}$ & 34.24$_{\pm 1.93}$
        & 28.78$_{\pm 1.25}$ & 64.56$_{\pm 9.80}$ & 28.29$_{\pm 0.50}$
        & 65.69$_{\pm 1.93}$ \\

        & MaxTokProb~\citep{fomicheva2020PPL}
        & 53.83$_{\pm 1.68}$ & 30.22$_{\pm 1.90}$ & 29.58$_{\pm 1.80}$
        & 64.30$_{\pm 1.22}$ & 66.58$_{\pm 1.45}$ & 61.45$_{\pm 1.07}$
        & 70.35$_{\pm 1.80}$ \\

        & PPL~\citep{fomicheva2020PPL}
        & 64.06$_{\pm 1.45}$ & 38.78$_{\pm 2.40}$ & 36.25$_{\pm 1.95}$
        & 51.07$_{\pm 1.21}$ & 55.59$_{\pm 1.75}$ & 45.42$_{\pm 0.85}$
        & 63.68$_{\pm 1.95}$ \\

        & BSDetector~\citep{chen2024BSDETECTOR}
        & 54.10$_{\pm 1.64}$ & 30.85$_{\pm 1.93}$ & 30.17$_{\pm 1.82}$
        & 58.26$_{\pm 1.22}$ & 61.96$_{\pm 2.24}$ & 54.00$_{\pm 0.92}$
        & 69.76$_{\pm 1.82}$ \\

        & SelfCheckGPT~\citep{manakul2023selfcheckgpt}
        & 75.26$_{\pm 1.56}$ & 56.03$_{\pm 2.46}$ & 45.60$_{\pm 1.79}$
        & 43.14$_{\pm 1.12}$ & 53.21$_{\pm 2.39}$ & 36.32$_{\pm 0.80}$
        & 54.33$_{\pm 1.79}$ \\

        & NumSet~\citep{lin2024numset/eigv/deg/ecc}
        & 80.92$_{\pm 1.22}$ & 58.78$_{\pm 2.35}$ & 48.74$_{\pm 1.78}$
        & 25.02$_{\pm 1.08}$ & 36.59$_{\pm 2.36}$ & 21.96$_{\pm 0.61}$
        & 51.19$_{\pm 1.78}$ \\

        & EigV~\citep{lin2024numset/eigv/deg/ecc}
        & 81.24$_{\pm 1.22}$ & 61.13$_{\pm 2.41}$ & 49.03$_{\pm 1.75}$
        & 4.36$_{\pm 0.80}$ & 22.37$_{\pm 4.51}$ & 15.20$_{\pm 0.65}$
        & 50.90$_{\pm 1.75}$ \\

        & Deg~\citep{lin2024numset/eigv/deg/ecc}
        & 81.24$_{\pm 1.22}$ & 61.13$_{\pm 2.41}$ & 49.03$_{\pm 1.75}$
        & 4.36$_{\pm 0.80}$ & 22.37$_{\pm 4.51}$ & 15.20$_{\pm 0.65}$
        & 50.90$_{\pm 1.75}$ \\

        & Ecc~\citep{lin2024numset/eigv/deg/ecc}
        & 22.00$_{\pm 1.24}$ & 19.43$_{\pm 0.95}$ & 11.92$_{\pm 0.93}$
        & 48.64$_{\pm 1.34}$ & 93.69$_{\pm 1.08}$ & 48.29$_{\pm 0.75}$
        & 88.01$_{\pm 0.93}$ \\

        & ConU~\citep{wang2024ConU}
        & 80.35$_{\pm 1.24}$ & 61.14$_{\pm 2.56}$ & 48.48$_{\pm 1.79}$
        & 15.94$_{\pm 1.05}$ & 29.39$_{\pm 3.75}$ & 18.07$_{\pm 0.58}$
        & 51.45$_{\pm 1.79}$ \\

        & Semantic Entropy~\citep{farquhar2024SemanticEntropy}
        & 77.61$_{\pm 1.39}$ & 58.72$_{\pm 2.55}$ & 47.19$_{\pm 1.84}$
        & 8.77$_{\pm 1.08}$ & 23.27$_{\pm 3.78}$ & 16.66$_{\pm 0.78}$
        & 52.74$_{\pm 1.84}$ \\

        & Semantic Density~\citep{Qiu2024SemanticDensity}
        & 79.32$_{\pm 1.36}$ & 60.98$_{\pm 2.57}$ & 48.25$_{\pm 1.79}$
        & 13.28$_{\pm 1.12}$ & 34.44$_{\pm 5.24}$ & 17.53$_{\pm 0.60}$
        & 51.69$_{\pm 1.79}$ \\

        & MARS~\citep{Bakman2024MARS}
        & 61.21$_{\pm 1.62}$ & 35.40$_{\pm 2.10}$ & 33.89$_{\pm 1.81}$
        & 43.76$_{\pm 1.22}$ & 55.97$_{\pm 3.95}$ & 39.19$_{\pm 0.70}$
        & 66.05$_{\pm 1.81}$ \\

        & Verbalized Probability~\citep{Lin2022Verbalized_Probability}
        & 42.59$_{\pm 1.60}$ & 24.40$_{\pm 1.49}$ & 23.65$_{\pm 1.57}$
        & 24.04$_{\pm 1.27}$ & 80.95$_{\pm 8.92}$ & 27.84$_{\pm 0.32}$
        & 76.28$_{\pm 1.57}$ \\

        & $P(\text{True})$~\citep{Kadavath2022Ptrue/PIK}
        & 53.65$_{\pm 1.78}$ & 30.52$_{\pm 1.70}$ & 30.34$_{\pm 1.68}$
        & 27.91$_{\pm 1.26}$ & 64.02$_{\pm 14.59}$ & 28.87$_{\pm 0.41}$
        & 69.59$_{\pm 1.68}$ \\

        & CoT-UQ-ME~\citep{Zhang2025CoT-UQ}
        & 49.75$_{\pm 1.70}$ & 26.48$_{\pm 1.53}$ & 25.82$_{\pm 1.59}$
        & 17.06$_{\pm 1.23}$ & 94.65$_{\pm 2.41}$ & 25.12$_{\pm 0.51}$
        & 74.11$_{\pm 1.59}$ \\

        & CoT-UQ-MI~\citep{Zhang2025CoT-UQ}
        & 52.43$_{\pm 1.61}$ & 29.91$_{\pm 1.76}$ & 29.25$_{\pm 1.66}$
        & 25.63$_{\pm 1.24}$ & 86.05$_{\pm 21.17}$ & 26.49$_{\pm 1.19}$
        & 70.68$_{\pm 1.66}$ \\

        & CoT-UQ-SAR~\citep{Zhang2025CoT-UQ}
        & 53.07$_{\pm 1.55}$ & 29.68$_{\pm 1.79}$ & 29.03$_{\pm 1.68}$
        & 8.29$_{\pm 1.10}$ & 66.18$_{\pm 22.55}$ & 21.16$_{\pm 0.58}$
        & 70.90$_{\pm 1.68}$ \\

        & CoT-UQ-SP~\citep{Zhang2025CoT-UQ}
        & 50.34$_{\pm 0.66}$ & 27.81$_{\pm 1.28}$ & 28.07$_{\pm 1.71}$
        & 24.23$_{\pm 1.25}$ & 72.71$_{\pm 5.13}$ & 26.61$_{\pm 0.30}$
        & 71.87$_{\pm 1.71}$ \\

        & Topo-UQ~\citep{Da2025Topo-uq}
        & 30.21$_{\pm 1.40}$ & 19.23$_{\pm 0.98}$ & 15.23$_{\pm 1.06}$
        & 53.05$_{\pm 1.27}$ & 87.84$_{\pm 5.41}$ & 52.63$_{\pm 0.84}$
        & 84.70$_{\pm 1.06}$ \\
        \midrule
        \midrule
        \multirow{21}{*}{\textbf{GSM8K}}
        & MaxSeqProb~\citep{Vashurin2025LM-Polygraph}
        & 59.09$_{\pm 2.64}$ & 8.17$_{\pm 1.17}$ & 7.45$_{\pm 0.98}$
        & 6.59$_{\pm 0.70}$ & 88.87$_{\pm 19.71}$ & 6.53$_{\pm 0.70}$
        & 92.47$_{\pm 0.98}$ \\

        & MeanTokEnt~\citep{fomicheva2020PPL}
        & 53.99$_{\pm 3.09}$ & 7.80$_{\pm 1.22}$ & 7.09$_{\pm 1.00}$
        & 43.36$_{\pm 0.80}$ & 96.01$_{\pm 2.55}$ & 26.60$_{\pm 0.38}$
        & 92.84$_{\pm 1.00}$ \\

        & MaxTokProb~\citep{fomicheva2020PPL}
        & 53.48$_{\pm 2.66}$ & 7.14$_{\pm 1.04}$ & 6.53$_{\pm 0.90}$
        & 81.29$_{\pm 0.76}$ & 86.83$_{\pm 1.04}$ & 72.71$_{\pm 0.69}$
        & 93.40$_{\pm 0.90}$ \\

        & PPL~\citep{fomicheva2020PPL}
        & 53.77$_{\pm 2.83}$ & 7.37$_{\pm 1.08}$ & 6.81$_{\pm 0.93}$
        & 63.55$_{\pm 0.82}$ & 88.60$_{\pm 3.40}$ & 48.17$_{\pm 0.57}$
        & 93.12$_{\pm 0.93}$ \\

        & BSDetector~\citep{chen2024BSDETECTOR}
        & 71.48$_{\pm 3.07}$ & 14.85$_{\pm 2.28}$ & 10.94$_{\pm 1.30}$
        & 17.08$_{\pm 0.65}$ & 92.21$_{\pm 10.84}$ & 9.03$_{\pm 0.42}$
        & 88.98$_{\pm 1.30}$ \\

        & SelfCheckGPT~\citep{manakul2023selfcheckgpt}
        & 57.68$_{\pm 3.32}$ & 10.05$_{\pm 1.77}$ & 8.35$_{\pm 1.16}$
        & 33.41$_{\pm 0.74}$ & 93.79$_{\pm 8.99}$ & 18.77$_{\pm 0.38}$
        & 91.57$_{\pm 1.16}$ \\

        & NumSet~\citep{lin2024numset/eigv/deg/ecc}
        & 67.31$_{\pm 3.03}$ & 12.82$_{\pm 2.17}$ & 10.53$_{\pm 1.32}$
        & 17.39$_{\pm 0.64}$ & 84.81$_{\pm 21.44}$ & 10.45$_{\pm 0.46}$
        & 89.39$_{\pm 1.32}$ \\

        & EigV~\citep{lin2024numset/eigv/deg/ecc}
        & 67.02$_{\pm 3.09}$ & 13.62$_{\pm 2.38}$ & 10.43$_{\pm 1.30}$
        & 2.11$_{\pm 0.58}$ & 84.67$_{\pm 22.27}$ & 6.04$_{\pm 0.59}$
        & 89.49$_{\pm 1.30}$ \\

        & Deg~\citep{lin2024numset/eigv/deg/ecc}
        & 66.97$_{\pm 3.08}$ & 13.47$_{\pm 2.35}$ & 10.43$_{\pm 1.29}$
        & 2.11$_{\pm 0.58}$ & 84.67$_{\pm 22.27}$ & 6.04$_{\pm 0.59}$
        & 89.49$_{\pm 1.29}$ \\

        & Ecc~\citep{lin2024numset/eigv/deg/ecc}
        & 33.25$_{\pm 2.98}$ & 4.74$_{\pm 0.57}$ & 3.79$_{\pm 0.66}$
        & 59.99$_{\pm 1.09}$ & 97.45$_{\pm 1.03}$ & 48.81$_{\pm 0.87}$
        & 96.13$_{\pm 0.66}$ \\

        & ConU~\citep{wang2024ConU}
        & 65.85$_{\pm 3.06}$ & 13.31$_{\pm 2.34}$ & 10.04$_{\pm 1.24}$
        & 10.28$_{\pm 0.65}$ & 91.20$_{\pm 14.10}$ & 7.80$_{\pm 0.51}$
        & 89.88$_{\pm 1.24}$ \\

        & Semantic Entropy~\citep{farquhar2024SemanticEntropy}
        & 60.93$_{\pm 3.24}$ & 9.55$_{\pm 1.53}$ & 8.30$_{\pm 1.14}$
        & 5.92$_{\pm 0.67}$ & 98.05$_{\pm 1.53}$ & 7.50$_{\pm 0.56}$
        & 91.62$_{\pm 1.14}$ \\

        & Semantic Density~\citep{Qiu2024SemanticDensity}
        & 62.61$_{\pm 3.33}$ & 10.56$_{\pm 1.76}$ & 8.90$_{\pm 1.23}$
        & 5.88$_{\pm 0.69}$ & 98.05$_{\pm 4.41}$ & 7.34$_{\pm 0.55}$
        & 91.02$_{\pm 1.23}$ \\

        & MARS~\citep{Bakman2024MARS}
        & 57.76$_{\pm 2.82}$ & 8.30$_{\pm 1.30}$ & 7.28$_{\pm 0.98}$
        & 62.34$_{\pm 0.80}$ & 89.12$_{\pm 2.39}$ & 46.38$_{\pm 0.56}$
        & 92.64$_{\pm 0.98}$ \\

        & Verbalized Probability~\citep{Lin2022Verbalized_Probability}
        & 48.86$_{\pm 3.02}$ & 6.63$_{\pm 1.06}$ & 6.03$_{\pm 0.92}$
        & 28.09$_{\pm 0.72}$ & 96.06$_{\pm 1.01}$ & 14.31$_{\pm 0.29}$
        & 93.89$_{\pm 0.92}$ \\

        & $P(\text{True})$~\citep{Kadavath2022Ptrue/PIK}
        & 51.60$_{\pm 3.21}$ & 7.88$_{\pm 1.61}$ & 6.50$_{\pm 1.01}$
        & 58.46$_{\pm 0.77}$ & 87.64$_{\pm 4.24}$ & 42.50$_{\pm 0.52}$
        & 93.43$_{\pm 1.01}$ \\

        & CoT-UQ-ME~\citep{Zhang2025CoT-UQ}
        & 48.30$_{\pm 2.90}$ & 6.56$_{\pm 0.97}$ & 5.97$_{\pm 0.87}$
        & 4.11$_{\pm 0.68}$ & 91.81$_{\pm 12.19}$ & 6.50$_{\pm 0.64}$
        & 93.96$_{\pm 0.87}$ \\

        & CoT-UQ-MI~\citep{Zhang2025CoT-UQ}
        & 48.09$_{\pm 2.94}$ & 6.72$_{\pm 0.99}$ & 6.13$_{\pm 0.88}$
        & 6.56$_{\pm 0.73}$ & 83.70$_{\pm 22.23}$ & 6.56$_{\pm 0.71}$
        & 93.79$_{\pm 0.88}$ \\

        & CoT-UQ-SAR~\citep{Zhang2025CoT-UQ}
        & 48.75$_{\pm 2.69}$ & 6.19$_{\pm 0.83}$ & 5.62$_{\pm 0.81}$
        & 9.46$_{\pm 0.71}$ & 84.35$_{\pm 20.57}$ & 7.48$_{\pm 0.50}$
        & 94.30$_{\pm 0.81}$ \\

        & CoT-UQ-SP~\citep{Zhang2025CoT-UQ}
        & 50.12$_{\pm 1.24}$ & 6.40$_{\pm 0.72}$ & 6.34$_{\pm 1.01}$
        & 91.07$_{\pm 0.80}$ & 93.53$_{\pm 0.72}$ & 90.47$_{\pm 0.81}$
        & 93.59$_{\pm 1.01}$ \\

        & Topo-UQ~\citep{Da2025Topo-uq}
        & 48.11$_{\pm 3.14}$ & 6.74$_{\pm 1.02}$ & 6.18$_{\pm 0.91}$
        & 89.34$_{\pm 0.73}$ & 90.86$_{\pm 0.72}$ & 86.23$_{\pm 0.69}$
        & 93.74$_{\pm 0.91}$ \\
        \midrule
        \midrule
        \multirow{21}{*}{\textbf{HumanEval}}
        & MaxSeqProb~\citep{Vashurin2025LM-Polygraph}
        & 64.02$_{\pm 5.00}$ & 41.97$_{\pm 6.78}$ & 36.95$_{\pm 5.40}$
        & 29.59$_{\pm 3.48}$ & 76.94$_{\pm 33.28}$ & 29.59$_{\pm 3.48}$
        & 62.43$_{\pm 5.40}$ \\

        & MeanTokEnt~\citep{fomicheva2020PPL}
        & 46.19$_{\pm 4.75}$ & 27.63$_{\pm 4.04}$ & 24.60$_{\pm 4.07}$
        & 24.43$_{\pm 3.21}$ & 95.19$_{\pm 3.42}$ & 28.17$_{\pm 1.40}$
        & 74.77$_{\pm 4.07}$ \\

        & MaxTokProb~\citep{fomicheva2020PPL}
        & 44.36$_{\pm 5.20}$ & 26.25$_{\pm 3.71}$ & 22.81$_{\pm 3.78}$
        & 58.75$_{\pm 3.54}$ & 95.25$_{\pm 9.45}$ & 56.26$_{\pm 3.09}$
        & 76.57$_{\pm 3.78}$ \\

        & PPL~\citep{fomicheva2020PPL}
        & 48.53$_{\pm 4.87}$ & 29.45$_{\pm 4.45}$ & 26.35$_{\pm 4.34}$
        & 38.97$_{\pm 3.40}$ & 82.42$_{\pm 13.71}$ & 37.39$_{\pm 1.67}$
        & 73.03$_{\pm 4.34}$ \\

        & BSDetector~\citep{chen2024BSDETECTOR}
        & 50.21$_{\pm 5.05}$ & 36.41$_{\pm 6.16}$ & 31.28$_{\pm 5.44}$
        & 20.68$_{\pm 3.33}$ & 47.66$_{\pm 22.13}$ & 24.36$_{\pm 2.81}$
        & 68.10$_{\pm 5.44}$ \\

        & SelfCheckGPT~\citep{manakul2023selfcheckgpt}
        & 45.26$_{\pm 5.33}$ & 31.04$_{\pm 5.44}$ & 27.45$_{\pm 5.02}$
        & 44.69$_{\pm 3.69}$ & 89.62$_{\pm 14.32}$ & 42.50$_{\pm 2.23}$
        & 71.93$_{\pm 5.02}$ \\

        & NumSet~\citep{lin2024numset/eigv/deg/ecc}
        & 54.16$_{\pm 3.57}$ & 34.52$_{\pm 5.19}$ & 33.72$_{\pm 5.46}$
        & 25.95$_{\pm 3.33}$ & 44.15$_{\pm 17.17}$ & 26.73$_{\pm 3.19}$
        & 65.66$_{\pm 5.46}$ \\

        & EigV~\citep{lin2024numset/eigv/deg/ecc}
        & 54.14$_{\pm 3.56}$ & 34.52$_{\pm 5.19}$ & 33.52$_{\pm 5.43}$
        & 27.70$_{\pm 3.38}$ & 27.98$_{\pm 3.42}$ & 27.73$_{\pm 3.32}$
        & 65.85$_{\pm 5.43}$ \\

        & Deg~\citep{lin2024numset/eigv/deg/ecc}
        & 54.14$_{\pm 3.56}$ & 34.52$_{\pm 5.19}$ & 33.52$_{\pm 5.43}$
        & 27.70$_{\pm 3.38}$ & 27.98$_{\pm 3.42}$ & 27.73$_{\pm 3.32}$
        & 65.85$_{\pm 5.43}$ \\

        & Ecc~\citep{lin2024numset/eigv/deg/ecc}
        & 45.82$_{\pm 3.57}$ & 27.36$_{\pm 3.48}$ & 26.91$_{\pm 4.60}$
        & 68.92$_{\pm 3.28}$ & 89.77$_{\pm 13.35}$ & 68.27$_{\pm 3.18}$
        & 72.47$_{\pm 4.60}$ \\

        & ConU~\citep{wang2024ConU}
        & 48.08$_{\pm 5.32}$ & 33.25$_{\pm 6.01}$ & 29.45$_{\pm 5.30}$
        & 23.98$_{\pm 3.36}$ & 40.72$_{\pm 9.90}$ & 25.96$_{\pm 3.07}$
        & 69.93$_{\pm 5.30}$ \\

        & Semantic Entropy~\citep{farquhar2024SemanticEntropy}
        & 40.53$_{\pm 4.66}$ & 25.37$_{\pm 3.79}$ & 22.06$_{\pm 3.99}$
        & 21.93$_{\pm 3.45}$ & 94.64$_{\pm 7.19}$ & 27.05$_{\pm 2.43}$
        & 77.32$_{\pm 3.99}$ \\

        & Semantic Density~\citep{Qiu2024SemanticDensity}
        & 53.18$_{\pm 5.00}$ & 37.59$_{\pm 6.45}$ & 32.98$_{\pm 5.30}$
        & 34.70$_{\pm 3.47}$ & 59.74$_{\pm 11.16}$ & 33.07$_{\pm 1.51}$
        & 66.40$_{\pm 5.30}$ \\

        & MARS~\citep{Bakman2024MARS}
        & 57.27$_{\pm 5.05}$ & 36.07$_{\pm 5.92}$ & 32.51$_{\pm 5.17}$
        & 27.57$_{\pm 3.19}$ & 75.95$_{\pm 15.51}$ & 28.79$_{\pm 1.42}$
        & 66.87$_{\pm 5.17}$ \\

        & Verbalized Probability~\citep{Lin2022Verbalized_Probability}
        & 52.70$_{\pm 5.31}$ & 35.24$_{\pm 6.00}$ & 31.32$_{\pm 5.12}$
        & 21.14$_{\pm 3.18}$ & 72.02$_{\pm 18.41}$ & 25.71$_{\pm 1.29}$
        & 68.06$_{\pm 5.12}$ \\

        & $P(\text{True})$~\citep{Kadavath2022Ptrue/PIK}
        & 34.44$_{\pm 4.44}$ & 24.13$_{\pm 3.37}$ & 19.69$_{\pm 3.70}$
        & 37.31$_{\pm 3.58}$ & 95.61$_{\pm 7.03}$ & 37.90$_{\pm 1.76}$
        & 79.69$_{\pm 3.70}$ \\

        & CoT-UQ-ME~\citep{Zhang2025CoT-UQ}
        & 46.11$_{\pm 5.06}$ & 26.89$_{\pm 3.82}$ & 23.27$_{\pm 3.87}$
        & 36.84$_{\pm 3.63}$ & 93.88$_{\pm 2.93}$ & 36.62$_{\pm 2.03}$
        & 76.11$_{\pm 3.87}$ \\

        & CoT-UQ-MI~\citep{Zhang2025CoT-UQ}
        & 48.54$_{\pm 4.54}$ & 30.94$_{\pm 4.74}$ & 28.13$_{\pm 4.56}$
        & 25.49$_{\pm 3.38}$ & 79.14$_{\pm 28.05}$ & 27.31$_{\pm 3.15}$
        & 71.25$_{\pm 4.56}$ \\

        & CoT-UQ-SAR~\citep{Zhang2025CoT-UQ}
        & 44.79$_{\pm 5.41}$ & 28.38$_{\pm 4.53}$ & 25.44$_{\pm 4.40}$
        & 24.15$_{\pm 3.32}$ & 93.42$_{\pm 5.02}$ & 27.84$_{\pm 1.73}$
        & 73.94$_{\pm 4.40}$ \\

        & CoT-UQ-SP~\citep{Zhang2025CoT-UQ}
        & 45.70$_{\pm 3.71}$ & 27.85$_{\pm 3.43}$ & 24.73$_{\pm 4.23}$
        & 33.24$_{\pm 4.06}$ & 81.50$_{\pm 8.36}$ & 37.49$_{\pm 2.32}$
        & 74.65$_{\pm 4.23}$ \\

        & Topo-UQ~\citep{Da2025Topo-uq}
        & 45.72$_{\pm 5.26}$ & 30.22$_{\pm 4.89}$ & 27.11$_{\pm 4.61}$
        & 48.50$_{\pm 3.46}$ & 85.91$_{\pm 10.77}$ & 44.49$_{\pm 2.13}$
        & 72.27$_{\pm 4.61}$ \\
        \bottomrule
    \end{tabular}
    }
\end{table*}
\FloatBarrier

\begin{table*}[t]
    \centering
    \caption{Performance Comparison of the investigated LLM UQ methods of Llama-3.2-3B in reasoning-augmented generation across three evaluated benchmarks.}
    \label{Llama-3.2-3B_answer_reasoning_results}
    \resizebox{0.7\textwidth}{!}{
    \begin{tabular}{llccccccc}
        \toprule
        \multirow{2}{*}{\textbf{Datasets}}
        & \multirow{2}{*}{\textbf{Methods}}
        & \multicolumn{3}{c}{\textbf{Ranking}}
        & \multicolumn{3}{c}{\textbf{Calibration}}
        & \textbf{Coverage} \\
        \cmidrule(lr){3-5} \cmidrule(lr){6-8} \cmidrule(lr){9-9}
        & & \textbf{AUROC} & \textbf{AUPRC} & \textbf{AUARC}
        & \textbf{ECE} & \textbf{MCE} & \textbf{BS}
        & \textbf{AURC} \\
        \midrule
        \multirow{21}{*}{\textbf{TriviaQA}}
        & MaxSeqProb~\citep{Vashurin2025LM-Polygraph}
        & 60.42$_{\pm 1.41}$ & 56.69$_{\pm 1.88}$ & 55.43$_{\pm 1.72}$
        & 47.58$_{\pm 1.33}$ & 47.66$_{\pm 1.33}$ & 47.54$_{\pm 1.33}$
        & 44.51$_{\pm 1.72}$ \\

        & MeanTokEnt~\citep{fomicheva2020PPL}
        & 58.64$_{\pm 1.37}$ & 53.77$_{\pm 1.95}$ & 52.99$_{\pm 1.85}$
        & 15.81$_{\pm 1.27}$ & 61.41$_{\pm 26.35}$ & 27.41$_{\pm 0.53}$
        & 46.94$_{\pm 1.85}$ \\

        & MaxTokProb~\citep{fomicheva2020PPL}
        & 53.28$_{\pm 1.33}$ & 49.65$_{\pm 1.67}$ & 49.22$_{\pm 1.65}$
        & 38.78$_{\pm 1.32}$ & 79.05$_{\pm 26.60}$ & 40.08$_{\pm 0.98}$
        & 50.71$_{\pm 1.65}$ \\

        & PPL~\citep{fomicheva2020PPL}
        & 60.16$_{\pm 1.34}$ & 55.84$_{\pm 1.94}$ & 54.78$_{\pm 1.78}$
        & 31.30$_{\pm 1.31}$ & 63.99$_{\pm 26.07}$ & 34.07$_{\pm 0.83}$
        & 45.15$_{\pm 1.78}$ \\

        & BSDetector~\citep{chen2024BSDETECTOR}
        & 52.88$_{\pm 1.51}$ & 50.22$_{\pm 1.84}$ & 49.78$_{\pm 1.80}$
        & 38.40$_{\pm 1.33}$ & 78.72$_{\pm 25.95}$ & 39.59$_{\pm 0.96}$
        & 50.15$_{\pm 1.80}$ \\

        & SelfCheckGPT~\citep{manakul2023selfcheckgpt}
        & 79.73$_{\pm 1.18}$ & 76.23$_{\pm 1.65}$ & 69.42$_{\pm 1.42}$
        & 30.20$_{\pm 1.13}$ & 42.34$_{\pm 2.64}$ & 28.09$_{\pm 0.78}$
        & 30.51$_{\pm 1.42}$ \\

        & NumSet~\citep{lin2024numset/eigv/deg/ecc}
        & 84.37$_{\pm 0.92}$ & 80.52$_{\pm 1.32}$ & 73.49$_{\pm 1.34}$
        & 18.47$_{\pm 1.05}$ & 33.83$_{\pm 3.21}$ & 20.10$_{\pm 0.57}$
        & 26.44$_{\pm 1.34}$ \\

        & EigV~\citep{lin2024numset/eigv/deg/ecc}
        & 85.55$_{\pm 0.92}$ & 83.32$_{\pm 1.29}$ & 74.25$_{\pm 1.29}$
        & 4.66$_{\pm 0.93}$ & 23.43$_{\pm 7.11}$ & 15.54$_{\pm 0.54}$
        & 25.68$_{\pm 1.29}$ \\

        & Deg~\citep{lin2024numset/eigv/deg/ecc}
        & 85.55$_{\pm 0.92}$ & 83.32$_{\pm 1.29}$ & 74.25$_{\pm 1.29}$
        & 4.66$_{\pm 0.93}$ & 23.43$_{\pm 7.11}$ & 15.54$_{\pm 0.54}$
        & 25.68$_{\pm 1.29}$ \\

        & Ecc~\citep{lin2024numset/eigv/deg/ecc}
        & 21.74$_{\pm 1.09}$ & 36.53$_{\pm 1.18}$ & 25.72$_{\pm 1.28}$
        & 25.75$_{\pm 1.19}$ & 92.19$_{\pm 3.86}$ & 36.12$_{\pm 0.68}$
        & 74.21$_{\pm 1.28}$ \\

        & ConU~\citep{wang2024ConU}
        & 85.22$_{\pm 0.95}$ & 83.47$_{\pm 1.43}$ & 73.68$_{\pm 1.34}$
        & 10.04$_{\pm 1.05}$ & 18.57$_{\pm 2.79}$ & 16.63$_{\pm 0.51}$
        & 26.26$_{\pm 1.34}$ \\

        & Semantic Entropy~\citep{farquhar2024SemanticEntropy}
        & 82.68$_{\pm 1.08}$ & 82.43$_{\pm 1.37}$ & 72.94$_{\pm 1.31}$
        & 15.63$_{\pm 1.11}$ & 30.06$_{\pm 2.63}$ & 19.61$_{\pm 0.73}$
        & 26.99$_{\pm 1.31}$ \\

        & Semantic Density~\citep{Qiu2024SemanticDensity}
        & 83.78$_{\pm 1.05}$ & 83.24$_{\pm 1.32}$ & 73.48$_{\pm 1.29}$
        & 8.34$_{\pm 1.01}$ & 17.18$_{\pm 3.03}$ & 17.12$_{\pm 0.56}$
        & 26.46$_{\pm 1.29}$ \\

        & MARS~\citep{Bakman2024MARS}
        & 58.45$_{\pm 1.37}$ & 54.37$_{\pm 1.87}$ & 53.49$_{\pm 1.77}$
        & 24.17$_{\pm 1.23}$ & 58.15$_{\pm 20.65}$ & 30.61$_{\pm 0.64}$
        & 46.45$_{\pm 1.77}$ \\

        & Verbalized Probability~\citep{Lin2022Verbalized_Probability}
        & 54.85$_{\pm 1.44}$ & 51.77$_{\pm 1.91}$ & 51.19$_{\pm 1.80}$
        & 8.54$_{\pm 1.19}$ & 28.39$_{\pm 9.76}$ & 25.71$_{\pm 0.39}$
        & 48.74$_{\pm 1.80}$ \\

        & $P(\text{True})$~\citep{Kadavath2022Ptrue/PIK}
        & 49.31$_{\pm 1.43}$ & 48.17$_{\pm 1.73}$ & 47.90$_{\pm 1.77}$
        & 13.97$_{\pm 1.20}$ & 70.06$_{\pm 14.94}$ & 27.78$_{\pm 0.40}$
        & 52.03$_{\pm 1.77}$ \\

        & CoT-UQ-ME~\citep{Zhang2025CoT-UQ}
        & 44.48$_{\pm 1.43}$ & 43.21$_{\pm 1.68}$ & 42.37$_{\pm 1.83}$
        & 17.64$_{\pm 1.21}$ & 93.13$_{\pm 11.04}$ & 29.18$_{\pm 0.42}$
        & 57.57$_{\pm 1.83}$ \\

        & CoT-UQ-MI~\citep{Zhang2025CoT-UQ}
        & 47.17$_{\pm 1.52}$ & 46.43$_{\pm 1.89}$ & 45.97$_{\pm 1.92}$
        & 45.28$_{\pm 1.32}$ & 51.66$_{\pm 5.48}$ & 45.57$_{\pm 1.26}$
        & 53.96$_{\pm 1.92}$ \\

        & CoT-UQ-SAR~\citep{Zhang2025CoT-UQ}
        & 44.00$_{\pm 1.48}$ & 42.87$_{\pm 1.69}$ & 41.93$_{\pm 1.85}$
        & 25.93$_{\pm 1.29}$ & 94.15$_{\pm 8.22}$ & 32.95$_{\pm 0.71}$
        & 58.00$_{\pm 1.85}$ \\

        & CoT-UQ-SP~\citep{Zhang2025CoT-UQ}
        & 49.84$_{\pm 0.43}$ & 47.50$_{\pm 1.34}$ & 47.52$_{\pm 1.85}$
        & 51.66$_{\pm 1.32}$ & 60.05$_{\pm 12.86}$ & 51.83$_{\pm 1.31}$
        & 52.42$_{\pm 1.85}$ \\

        & Topo-UQ~\citep{Da2025Topo-uq}
        & 34.95$_{\pm 1.45}$ & 37.42$_{\pm 1.40}$ & 34.14$_{\pm 1.66}$
        & 36.65$_{\pm 1.40}$ & 65.04$_{\pm 2.08}$ & 41.37$_{\pm 1.02}$
        & 65.79$_{\pm 1.66}$ \\
        \midrule
        \midrule
        \multirow{21}{*}{\textbf{GSM8K}}
        & MaxSeqProb~\citep{Vashurin2025LM-Polygraph}
        & 70.33$_{\pm 1.84}$ & 42.25$_{\pm 2.82}$ & 37.21$_{\pm 1.97}$
        & 24.93$_{\pm 1.24}$ & 43.78$_{\pm 12.40}$ & 24.89$_{\pm 1.23}$
        & 62.71$_{\pm 1.97}$ \\

        & MeanTokEnt~\citep{fomicheva2020PPL}
        & 62.32$_{\pm 1.99}$ & 35.19$_{\pm 2.39}$ & 32.67$_{\pm 1.84}$
        & 30.03$_{\pm 1.21}$ & 70.49$_{\pm 16.10}$ & 27.54$_{\pm 0.44}$
        & 67.26$_{\pm 1.84}$ \\

        & MaxTokProb~\citep{fomicheva2020PPL}
        & 56.46$_{\pm 1.89}$ & 28.32$_{\pm 1.87}$ & 27.44$_{\pm 1.74}$
        & 65.66$_{\pm 1.22}$ & 67.47$_{\pm 1.39}$ & 61.78$_{\pm 1.03}$
        & 72.48$_{\pm 1.74}$ \\

        & PPL~\citep{fomicheva2020PPL}
        & 61.36$_{\pm 2.01}$ & 34.92$_{\pm 2.46}$ & 32.42$_{\pm 1.88}$
        & 48.26$_{\pm 1.20}$ & 57.18$_{\pm 9.80}$ & 41.94$_{\pm 0.68}$
        & 67.50$_{\pm 1.88}$ \\

        & BSDetector~\citep{chen2024BSDETECTOR}
        & 89.00$_{\pm 1.10}$ & 80.46$_{\pm 1.86}$ & 53.00$_{\pm 1.75}$
        & 12.95$_{\pm 0.78}$ & 22.83$_{\pm 4.05}$ & 11.34$_{\pm 0.39}$
        & 46.92$_{\pm 1.75}$ \\

        & SelfCheckGPT~\citep{manakul2023selfcheckgpt}
        & 82.76$_{\pm 1.37}$ & 72.37$_{\pm 2.21}$ & 49.63$_{\pm 1.82}$
        & 20.23$_{\pm 0.96}$ & 28.86$_{\pm 2.13}$ & 16.59$_{\pm 0.34}$
        & 50.29$_{\pm 1.82}$ \\

        & NumSet~\citep{lin2024numset/eigv/deg/ecc}
        & 81.42$_{\pm 1.38}$ & 60.56$_{\pm 2.59}$ & 47.07$_{\pm 1.90}$
        & 12.13$_{\pm 0.98}$ & 23.74$_{\pm 2.99}$ & 15.38$_{\pm 0.51}$
        & 52.86$_{\pm 1.90}$ \\

        & EigV~\citep{lin2024numset/eigv/deg/ecc}
        & 82.18$_{\pm 1.37}$ & 65.96$_{\pm 2.59}$ & 48.01$_{\pm 1.91}$
        & 8.28$_{\pm 1.00}$ & 35.70$_{\pm 9.58}$ & 14.01$_{\pm 0.71}$
        & 51.92$_{\pm 1.91}$ \\

        & Deg~\citep{lin2024numset/eigv/deg/ecc}
        & 82.20$_{\pm 1.37}$ & 65.05$_{\pm 2.65}$ & 47.94$_{\pm 1.91}$
        & 8.28$_{\pm 1.00}$ & 35.70$_{\pm 9.58}$ & 14.01$_{\pm 0.71}$
        & 51.99$_{\pm 1.91}$ \\

        & Ecc~\citep{lin2024numset/eigv/deg/ecc}
        & 21.49$_{\pm 1.40}$ & 17.04$_{\pm 0.94}$ & 10.80$_{\pm 1.01}$
        & 46.84$_{\pm 1.32}$ & 96.33$_{\pm 1.57}$ & 46.67$_{\pm 0.74}$
        & 89.12$_{\pm 1.01}$ \\

        & ConU~\citep{wang2024ConU}
        & 82.19$_{\pm 1.38}$ & 67.00$_{\pm 2.68}$ & 48.02$_{\pm 1.93}$
        & 5.10$_{\pm 0.87}$ & 14.49$_{\pm 3.90}$ & 13.38$_{\pm 0.55}$
        & 51.90$_{\pm 1.93}$ \\

        & Semantic Entropy~\citep{farquhar2024SemanticEntropy}
        & 69.31$_{\pm 1.70}$ & 43.85$_{\pm 2.84}$ & 37.92$_{\pm 1.99}$
        & 12.23$_{\pm 1.13}$ & 59.48$_{\pm 8.96}$ & 18.57$_{\pm 0.87}$
        & 62.01$_{\pm 1.99}$ \\

        & Semantic Density~\citep{Qiu2024SemanticDensity}
        & 71.46$_{\pm 1.59}$ & 45.48$_{\pm 2.83}$ & 38.90$_{\pm 1.98}$
        & 8.54$_{\pm 1.10}$ & 39.79$_{\pm 11.36}$ & 17.42$_{\pm 0.73}$
        & 61.02$_{\pm 1.98}$ \\

        & MARS~\citep{Bakman2024MARS}
        & 65.15$_{\pm 1.94}$ & 38.08$_{\pm 2.83}$ & 34.53$_{\pm 2.04}$
        & 50.06$_{\pm 1.17}$ & 58.37$_{\pm 7.18}$ & 43.08$_{\pm 0.72}$
        & 65.39$_{\pm 2.04}$ \\

        & Verbalized Probability~\citep{Lin2022Verbalized_Probability}
        & 45.89$_{\pm 1.64}$ & 23.14$_{\pm 1.48}$ & 22.48$_{\pm 1.52}$
        & 4.43$_{\pm 1.05}$ & 96.94$_{\pm 4.01}$ & 19.67$_{\pm 0.64}$
        & 77.44$_{\pm 1.52}$ \\

        & $P(\text{True})$~\citep{Kadavath2022Ptrue/PIK}
        & 43.46$_{\pm 1.83}$ & 21.42$_{\pm 1.34}$ & 20.21$_{\pm 1.47}$
        & 33.94$_{\pm 1.33}$ & 94.49$_{\pm 0.96}$ & 32.92$_{\pm 0.48}$
        & 79.71$_{\pm 1.47}$ \\

        & CoT-UQ-ME~\citep{Zhang2025CoT-UQ}
        & 40.92$_{\pm 1.67}$ & 20.27$_{\pm 1.24}$ & 18.44$_{\pm 1.38}$
        & 23.77$_{\pm 1.25}$ & 90.07$_{\pm 14.32}$ & 24.69$_{\pm 1.20}$
        & 81.48$_{\pm 1.38}$ \\

        & CoT-UQ-MI~\citep{Zhang2025CoT-UQ}
        & 48.75$_{\pm 1.73}$ & 24.72$_{\pm 1.68}$ & 24.17$_{\pm 1.68}$
        & 25.11$_{\pm 1.24}$ & 91.89$_{\pm 16.44}$ & 25.12$_{\pm 1.24}$
        & 75.75$_{\pm 1.68}$ \\

        & CoT-UQ-SAR~\citep{Zhang2025CoT-UQ}
        & 45.62$_{\pm 1.82}$ & 21.94$_{\pm 1.40}$ & 20.75$_{\pm 1.55}$
        & 14.30$_{\pm 1.22}$ & 94.29$_{\pm 10.71}$ & 20.92$_{\pm 0.91}$
        & 79.17$_{\pm 1.55}$ \\

        & CoT-UQ-SP~\citep{Zhang2025CoT-UQ}
        & 50.45$_{\pm 0.59}$ & 25.20$_{\pm 1.26}$ & 25.21$_{\pm 1.76}$
        & 72.61$_{\pm 1.30}$ & 85.73$_{\pm 5.90}$ & 72.55$_{\pm 1.28}$
        & 74.72$_{\pm 1.76}$ \\

        & Topo-UQ~\citep{Da2025Topo-uq}
        & 40.00$_{\pm 1.98}$ & 22.55$_{\pm 1.74}$ & 21.25$_{\pm 1.74}$
        & 70.76$_{\pm 1.26}$ & 90.15$_{\pm 12.47}$ & 69.68$_{\pm 1.16}$
        & 78.68$_{\pm 1.74}$ \\
        \midrule
        \midrule
        \multirow{21}{*}{\textbf{HumanEval}}
        & MaxSeqProb~\citep{Vashurin2025LM-Polygraph}
        & 52.26$_{\pm 4.84}$ & 43.51$_{\pm 5.44}$ & 40.04$_{\pm 4.91}$
        & 37.50$_{\pm 3.59}$ & 88.31$_{\pm 16.55}$ & 37.20$_{\pm 3.57}$
        & 59.34$_{\pm 4.91}$ \\

        & MeanTokEnt~\citep{fomicheva2020PPL}
        & 45.67$_{\pm 4.74}$ & 36.25$_{\pm 4.50}$ & 33.31$_{\pm 4.56}$
        & 17.82$_{\pm 3.25}$ & 78.48$_{\pm 8.87}$ & 27.28$_{\pm 1.58}$
        & 66.07$_{\pm 4.56}$ \\

        & MaxTokProb~\citep{fomicheva2020PPL}
        & 53.51$_{\pm 4.76}$ & 41.34$_{\pm 5.36}$ & 38.39$_{\pm 5.03}$
        & 42.63$_{\pm 3.93}$ & 78.00$_{\pm 18.07}$ & 42.41$_{\pm 2.84}$
        & 60.99$_{\pm 5.03}$ \\

        & PPL~\citep{fomicheva2020PPL}
        & 43.98$_{\pm 4.75}$ & 35.91$_{\pm 4.50}$ & 32.96$_{\pm 4.59}$
        & 23.71$_{\pm 3.55}$ & 83.26$_{\pm 10.16}$ & 30.17$_{\pm 1.50}$
        & 66.42$_{\pm 4.59}$ \\

        & BSDetector~\citep{chen2024BSDETECTOR}
        & 51.64$_{\pm 5.07}$ & 46.67$_{\pm 5.43}$ & 41.88$_{\pm 4.90}$
        & 11.70$_{\pm 3.41}$ & 67.69$_{\pm 23.02}$ & 23.92$_{\pm 1.91}$
        & 57.50$_{\pm 4.90}$ \\

        & SelfCheckGPT~\citep{manakul2023selfcheckgpt}
        & 45.49$_{\pm 4.62}$ & 39.97$_{\pm 5.18}$ & 36.48$_{\pm 4.88}$
        & 41.32$_{\pm 3.65}$ & 88.02$_{\pm 14.92}$ & 42.04$_{\pm 2.25}$
        & 62.90$_{\pm 4.88}$ \\

        & NumSet~\citep{lin2024numset/eigv/deg/ecc}
        & 60.04$_{\pm 3.33}$ & 50.72$_{\pm 4.76}$ & 48.95$_{\pm 4.82}$
        & 29.64$_{\pm 3.42}$ & 39.17$_{\pm 8.61}$ & 29.97$_{\pm 3.33}$
        & 50.43$_{\pm 4.82}$ \\

        & EigV~\citep{lin2024numset/eigv/deg/ecc}
        & 60.07$_{\pm 3.33}$ & 51.15$_{\pm 4.73}$ & 49.05$_{\pm 4.85}$
        & 31.62$_{\pm 3.41}$ & 60.24$_{\pm 6.34}$ & 30.91$_{\pm 3.34}$
        & 50.33$_{\pm 4.85}$ \\

        & Deg~\citep{lin2024numset/eigv/deg/ecc}
        & 60.07$_{\pm 3.33}$ & 51.15$_{\pm 4.73}$ & 49.05$_{\pm 4.85}$
        & 31.62$_{\pm 3.41}$ & 60.24$_{\pm 6.34}$ & 30.91$_{\pm 3.34}$
        & 50.33$_{\pm 4.85}$ \\

        & Ecc~\citep{lin2024numset/eigv/deg/ecc}
        & 39.99$_{\pm 3.32}$ & 32.76$_{\pm 3.56}$ & 31.85$_{\pm 4.84}$
        & 64.02$_{\pm 3.72}$ & 94.43$_{\pm 7.39}$ & 63.29$_{\pm 3.42}$
        & 67.53$_{\pm 4.84}$ \\

        & ConU~\citep{wang2024ConU}
        & 55.34$_{\pm 4.86}$ & 52.70$_{\pm 5.63}$ & 45.88$_{\pm 4.87}$
        & 27.12$_{\pm 3.49}$ & 51.90$_{\pm 12.15}$ & 28.75$_{\pm 3.06}$
        & 53.49$_{\pm 4.87}$ \\

        & Semantic Entropy~\citep{farquhar2024SemanticEntropy}
        & 41.14$_{\pm 4.49}$ & 31.98$_{\pm 3.69}$ & 28.01$_{\pm 3.99}$
        & 30.57$_{\pm 3.68}$ & 92.22$_{\pm 13.03}$ & 33.27$_{\pm 2.93}$
        & 71.37$_{\pm 3.99}$ \\

        & Semantic Density~\citep{Qiu2024SemanticDensity}
        & 51.18$_{\pm 4.51}$ & 40.50$_{\pm 5.15}$ & 37.73$_{\pm 4.92}$
        & 29.67$_{\pm 3.57}$ & 63.58$_{\pm 11.95}$ & 32.81$_{\pm 1.59}$
        & 61.65$_{\pm 4.92}$ \\

        & MARS~\citep{Bakman2024MARS}
        & 52.05$_{\pm 4.53}$ & 43.07$_{\pm 5.35}$ & 39.67$_{\pm 4.91}$
        & 26.38$_{\pm 3.45}$ & 83.54$_{\pm 9.75}$ & 31.36$_{\pm 1.59}$
        & 59.71$_{\pm 4.91}$ \\

        & Verbalized Probability~\citep{Lin2022Verbalized_Probability}
        & 52.94$_{\pm 4.22}$ & 43.42$_{\pm 5.09}$ & 39.99$_{\pm 4.80}$
        & 25.59$_{\pm 3.67}$ & 73.93$_{\pm 16.30}$ & 30.44$_{\pm 1.55}$
        & 59.39$_{\pm 4.80}$ \\

        & $P(\text{True})$~\citep{Kadavath2022Ptrue/PIK}
        & 48.10$_{\pm 4.95}$ & 37.86$_{\pm 4.51}$ & 35.25$_{\pm 4.86}$
        & 23.47$_{\pm 3.79}$ & 74.44$_{\pm 17.04}$ & 30.11$_{\pm 1.71}$
        & 64.13$_{\pm 4.86}$ \\

        & CoT-UQ-ME~\citep{Zhang2025CoT-UQ}
        & 38.43$_{\pm 4.54}$ & 31.58$_{\pm 3.98}$ & 27.87$_{\pm 4.38}$
        & 35.04$_{\pm 3.85}$ & 84.27$_{\pm 9.16}$ & 38.54$_{\pm 2.07}$
        & 71.51$_{\pm 4.38}$ \\

        & CoT-UQ-MI~\citep{Zhang2025CoT-UQ}
        & 43.64$_{\pm 4.79}$ & 35.14$_{\pm 4.76}$ & 32.23$_{\pm 4.95}$
        & 28.39$_{\pm 3.58}$ & 81.44$_{\pm 25.15}$ & 32.09$_{\pm 2.86}$
        & 67.15$_{\pm 4.95}$ \\

        & CoT-UQ-SAR~\citep{Zhang2025CoT-UQ}
        & 42.03$_{\pm 4.81}$ & 33.55$_{\pm 4.30}$ & 30.41$_{\pm 4.62}$
        & 23.80$_{\pm 3.65}$ & 95.44$_{\pm 3.48}$ & 30.77$_{\pm 2.04}$
        & 68.97$_{\pm 4.62}$ \\

        & CoT-UQ-SP~\citep{Zhang2025CoT-UQ}
        & 38.90$_{\pm 3.81}$ & 34.21$_{\pm 3.57}$ & 30.53$_{\pm 4.47}$
        & 39.75$_{\pm 3.64}$ & 80.09$_{\pm 3.12}$ & 40.92$_{\pm 2.98}$
        & 68.85$_{\pm 4.47}$ \\

        & Topo-UQ~\citep{Da2025Topo-uq}
        & 41.31$_{\pm 4.82}$ & 37.62$_{\pm 4.83}$ & 33.70$_{\pm 4.88}$
        & 43.48$_{\pm 3.66}$ & 96.13$_{\pm 6.92}$ & 43.20$_{\pm 2.34}$
        & 65.68$_{\pm 4.88}$ \\
        \bottomrule
    \end{tabular}
    }
\end{table*}
\FloatBarrier

\clearpage
\begin{figure*}[!htb]
    \centering
    \includegraphics[width=\textwidth]{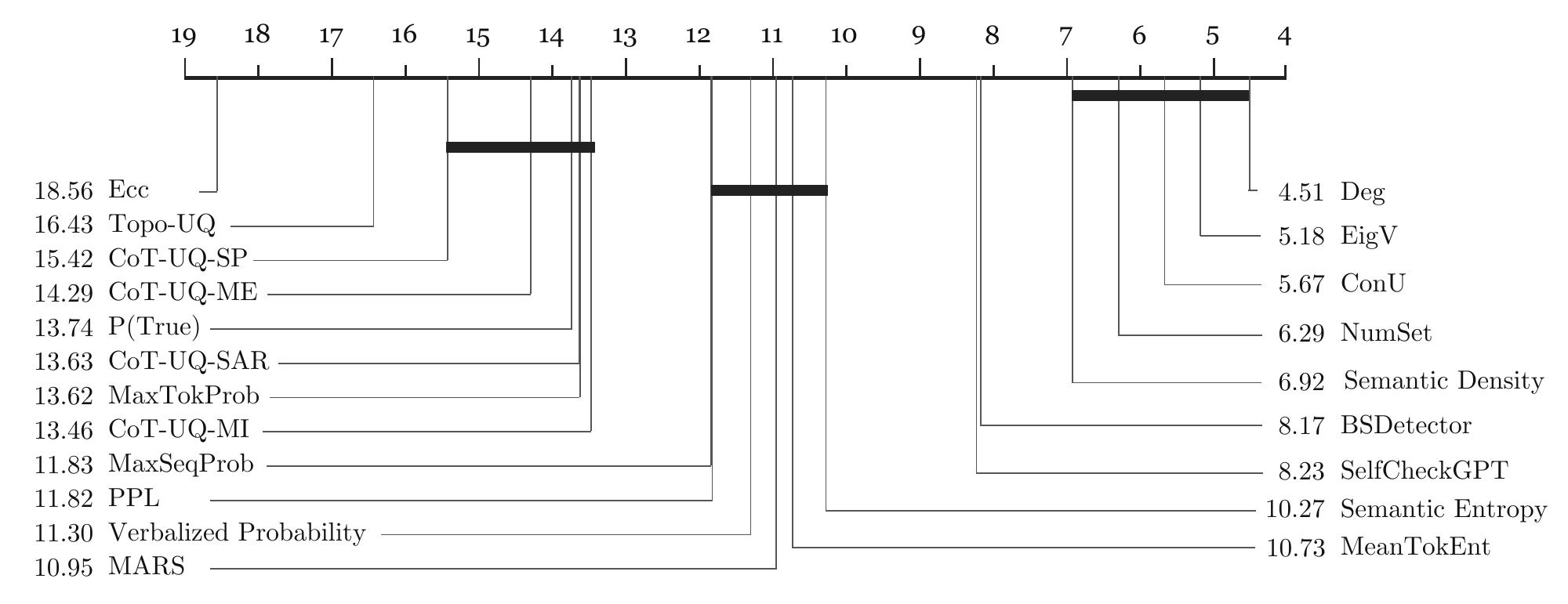}
    \caption{Overall evaluation ranking of the investigated LLM UQ methods in the reasoning-augmented generation setting.}
    \label{fig:reasoning_ranking}
\end{figure*}

Further, Figure~\ref{fig:reasoning_ranking} displays the overall evaluation ranks of the investigated LLM UQ methods in the reasoning-augmented generation setting, where the ranking scores are based on the average performance over 7 evaluation metrics across the Qwen3 and Llama families. There are four primary observations. First, many consensus-based methods, including Deg, EigV, ConU, and NumSet, fall into the first tier in terms of average rank; especially Deg and EigV achieve the best average ranks, clearly outperforming the other methods. Second, compared to the answer-only setting, the overall rankings of both Semantic Entropy and Semantic Density improve by 6 positions, representing substantially larger gains than those of other methods. We conjecture that disagreements among sampled responses reflect variations in both the final answers and reasoning steps, making semantic consistency and density more informative for uncertainty estimation. Third, reasoning-specific methods, including the four CoT-UQ variants and Topo-UQ, fall into the bottom tier in terms of overall rank. We conjecture that these reasoning-specific methods are more sensitive to task type, response format, sampling quality, and implementation details. Fourth, compared to the answer-only setting, the overall rankings of MaxSeqProb, PPL, MeanTokEnt, and MaxTokProb drop by 10, 2, 1, and 1 positions, respectively. These four token-level single-pass UQ methods fall into the second tier overall among all UQ methods, with MaxTokProb exhibiting particularly poor performance. A possible reason is that, under the reasoning-augmented setting, LLMs output reasoning chains with more variable lengths, expression styles, and local token-level fluctuations than in the answer-only setting; thus, methods depending on token-level signals may become more sensitive to noise. Figure~\ref{fig:reasoning_ranking_combined} in Appendix~\ref{app:ranking_answer_only} further visualizes the overall ranking plots for the ranking, calibration, and coverage evaluations in the reasoning-augmented setting.

\textbf{Overall, observations aforementioned in Subsections~\ref{subsec:answer_only_uq} and~\ref{subsec:reasoning_augmented_uq} answer Q2.}

\clearpage
\subsection{Relation between LLM Uncertainty and Scaling Law}  \label{subsec:scaling_law}
In this subsection, we investigate whether and to what extent the scaling law of LLMs affects their uncertainties, or equally, the effectiveness of the investigated UQ methods. This investigation is motivated by the seminal study of~\citet{kaplan2020scaling}, which empirically demonstrated that the test loss of language models decreases as the parameter scale increases, exhibiting a power-law relationship commonly referred to as the ``scaling law''. This finding has motivated subsequent research~\citep{Llama_Team2024llama,Qwen_Team2025qwen3} and industrial efforts~\citep{achiam2023gpt} to improve predictive precision of LLMs by scaling up parameter counts. Despite great progress in predictive precision, the natural yet critical question of whether LLM output uncertainty exhibits a similar scaling law remains underexplored. The answer to this question may significantly impact the credibility of next-generation LLMs, determining whether simply increasing model size is always beneficial.

Inspired by this recognition, we conduct experiments by quantifying the average uncertainty of LLMs with increasing scales against accuracy across the TriviaQA, GSM8K, and HumanEval datasets. The average uncertainty~\citep{ye2024benchmarking} is computed by first quantifying the output uncertainty of each instance using a specific UQ method, for which we select four representative approaches involving Semantic Entropy, EigV, NumSet, and BSDetector, and subsequently averaging these quantifications across all instances within a dataset. Therefore, this metric reflects the average output uncertainty of an LLM on a given dataset, with larger values indicating greater uncertainty.

\begin{figure*}[!htb]
    \centering
    \includegraphics[width=0.95\linewidth]{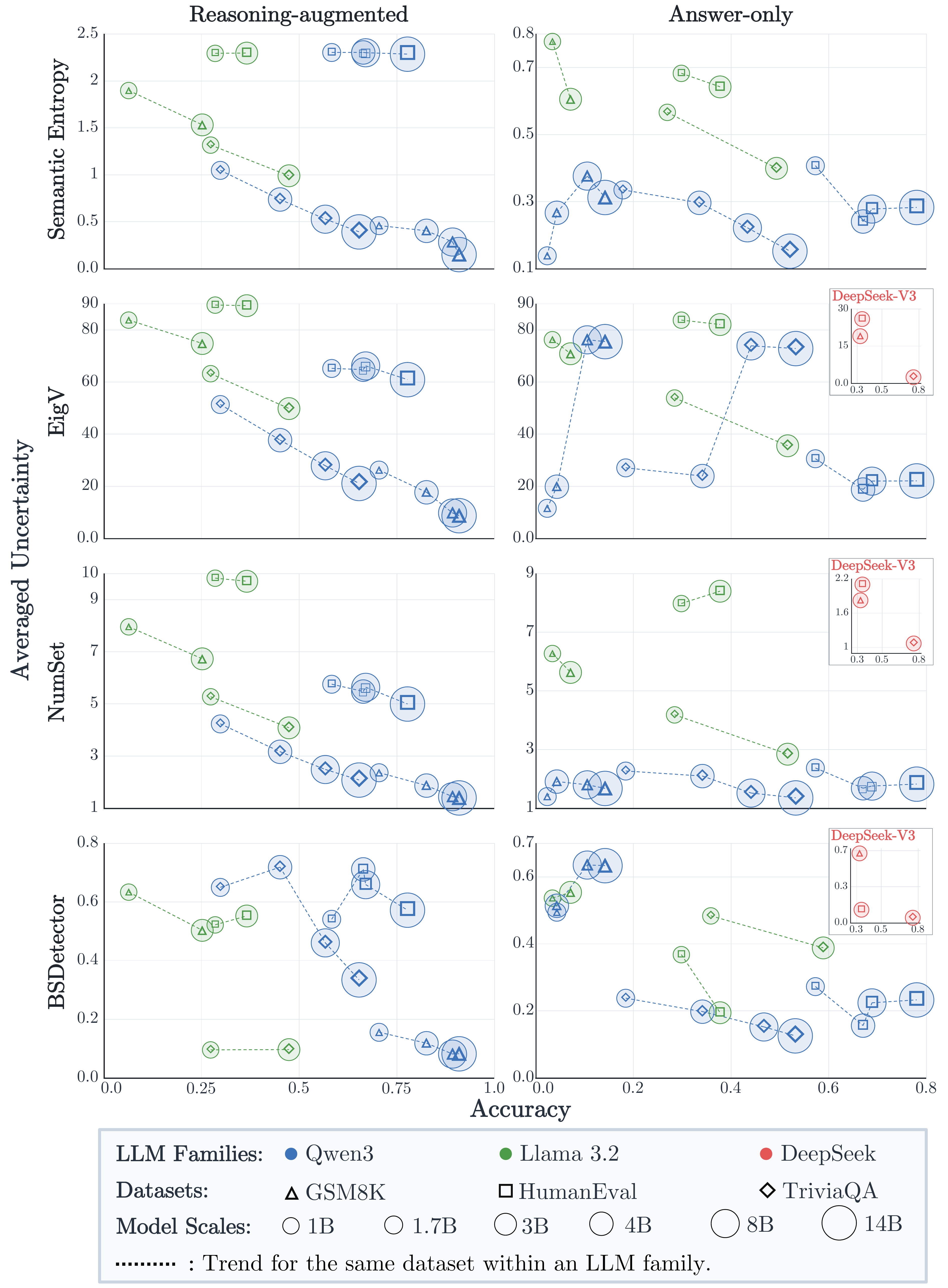}
    \caption{Plots of output uncertainty versus accuracy across parameter scales.}
    \label{fig:scaling_law}
\end{figure*}

Figure~\ref{fig:scaling_law} plots the curves of quantified uncertainty versus accuracy across parameter scales. The first and second columns correspond to the reasoning-augmented and answer-only settings, respectively, whereas the rows correspond to four representative UQ methods, namely Semantic Entropy, EigV, NumSet, and BSDetector. Within each subfigure, the x- and y-axes denote the accuracy and the average uncertainty, respectively. The dotted line illustrates the trend for the same LLM family, where a larger circle indicates a larger model scale. We observe that larger circles tend to cluster in the bottom-right region of each subfigure, whereas smaller circles are generally located in the top-left, resulting in a downward trend for the dotted line. This observation reveals two conclusions. First, LLMs with larger scales usually achieve higher predictive precision, which is consistent with the scaling law of model performance~\citep{kaplan2020scaling}. Second, larger-scale LLMs tend to produce outputs with lower uncertainty estimates. This insight reveals an empirical scaling law of LLM uncertainty, demonstrating that larger models tend to be better calibrated. This conjecture corroborates the findings in~\citet{Abe2022ensembles} regarding uncertainty calibration, demonstrating that larger models tend to be better calibrated. Both conclusions above may shed light on another potential benefit of scaling up LLMs, in which \textbf{LLMs with larger scales may become more credible from the perspective of uncertainty analysis, thus answering Q3.}

\section{Conclusions, Discussions, and Prospects}  \label{sec:discussions}
This work focuses on how to judge the predictive credibility of LLMs in reasoning-augmented and open-ended content generation. We proposed a granular uncertainty taxonomy that systematically attributes LLM uncertainty into input-level, parameter-level, token-level, and decoding-process sources and correspondingly categorized existing UQ methods into Bayesian, ensemble, consensus-based, and single-pass approaches. We further introduced a comprehensive evaluation framework covering diverse generation settings and metrics, with empirical investigations conducted on 21 typical UQ methods across three prominent LLM families, including Qwen3, Llama 3.2, and DeepSeek-V3, on benchmarks such as TriviaQA, GSM8K, and HumanEval. The results show that (1) the effectiveness of UQ methods is sensitive to task types and generation settings; (2) consensus-based methods, typed Deg and EigV, consistently outperform other UQ approaches; and (3) larger model scales correlate with lower uncertainty estimates, suggesting an empirical scaling law for LLM uncertainty.

\paragraph{About the Uncertainty Attribution and Inherent Stochasticity.} The integration of uncertainty attribution, quantification, and uncertainty-based decision-making constitutes the cornerstone of trustworthy LLMs. This integration forms a hierarchical pipeline where the identification of specific uncertainty origins fundamentally governs the credibility of downstream predictive outcomes. By decomposing uncertainty into a granular framework comprising input, parameter, token-level, and decoding uncertainties, we pinpoint that the LLM uncertainty is influenced by the stochasticity inside these uncertainty sources~\citep{piray2026not}, the model capacity~\citep{zhou2021over,Liang2026EMP}, and the compounding nature of cascading uncertainty across sequential generation steps~\citep{Da2025Topo-uq}. It is worth noting that the inherent stochasticity also persists as an irreducible limit in alignment with the fundamental constraints of PAC learning~\citep{valiant1984:PAC}. Thereby, while parameter-related and decoding-induced uncertainties may be mitigated through architectural refinement or improved calibration techniques, the residual uncertainty rooted in inherent data stochasticity establishes an unavoidable performance ceiling.

\paragraph{About the Uncertainty Quantification and Scaling Law.} To rigorously assess the efficacy of UQ methods in LLMs, we established a comprehensive evaluation framework that systematically examines representative UQ methods, such as Semantic Entropy, EigV, NumSet, and BSDetector, across diverse configurations, including both answer-only and reasoning-augmented generation settings, and evaluated them through metrics encompassing ranking performance, calibration, and coverage. Our empirical results, validated across prominent LLM families like Qwen3, Llama 3.2, and DeepSeek, identify methods such as Deg and ConU as superior for calibration, while Deg and EigV demonstrate exceptional performance in distinguishing correct from incorrect responses and ensuring reliable coverage behavior. Central to our findings is the observation of an empirical scaling law of uncertainty, which manifests as a consistent downward trend in quantified uncertainty as model parameter scales increase, as shown in Figure~\ref{fig:scaling_law}. This finding indicates that, beyond the well-documented improvements in predictive precision~\citep{kaplan2020scaling}, scaling up LLMs simultaneously diminishes output uncertainty and enhances model credibility. Collectively, these results reveal that larger-scale architectures may inherently possess a superior foundation for reliable and uncertainty-aware decision-making in complex tasks.

\paragraph{Future issues.} One important future issue is to reduce quantified uncertainty, thereby improving the certainty and credibility of model outputs. This is particularly important in safety-critical domains such as healthcare~\citep{nemani2023uncertainty} and autonomous driving~\citep{michelmore2018evaluating}. According to the framework of uncertainty sources proposed in Subsection~\ref{subsec:uncertainty_sources}, the key to reducing output uncertainty in LLMs is to reduce stochasticity in the input, parameters, token-level distributions, and decoding strategies, respectively. Therefore, from the perspective of probability computing, it suffices to make the corresponding distributions sharper. MOMA~\citep{Liang2026EMP} reduces stochasticity in model weights by making the weight distribution more concentrated at zero, thereby reducing output uncertainty through lower parameter uncertainty. Top-$k$ and top-$p$ decoding strategies~\citep{Holtzman2020topp} truncate the long tail of the token-level distribution, making the distribution sharper and thereby reducing token-level uncertainty. \cite{xiao2021hallucination} maintained several decoding sequences with the lowest uncertainty at each decoding step and finally selected the sequence with the lowest overall uncertainty, which reduced the decoding uncertainty. \citet{zhang2026uncertainty} collected the confidence calibration methods for mitigating the high LLM uncertainty. Despite these efforts, systematically reducing uncertainty in LLMs remains challenging, as output uncertainty is jointly induced by multiple sources rather than a single isolated factor. Therefore, developing unified uncertainty optimization methods that can simultaneously reduce the input, parameter, token-level, and decoding uncertainty is a promising direction for future research.

Another important future issue is to develop new uncertainty quantification and optimization methods for the LLM-driven agent models. These agent models have demonstrated remarkable performance in coding assistance~\citep{yang2024sweagent}, mathematical theorem proving~\citep{baba2025prover}, and automated scientific research~\citep{lu2026towards}. However, hallucination becomes more severe in LLM-driven agents because hallucinations from single-turn LLM generation can be amplified through multi-turn agentic loops~\citep{yu2026visual}. This makes such errors harder for humans to detect and intervene in, potentially leading to failures such as erroneous tool execution and incorrect decision-making. In high-stakes scenarios, these failures may further result in real-world damage, including property loss~\citep{shapira2026agents}. Uncertainty quantification and optimization provide practical methodological pathways for hallucination detection and mitigation in LLM-driven agents, respectively. Therefore, developing such methods represents a critical direction for future research.

\section*{Acknowledgements}
This research was supported by the Natural Science Foundation of China (62406138).

\appendix

\section{Evaluation Ranking}  \label{app:ranking_answer_only}
\begin{figure*}[t]
    \centering
    \begin{subfigure}[b]{\textwidth}
        \centering
        \includegraphics[width=\textwidth]{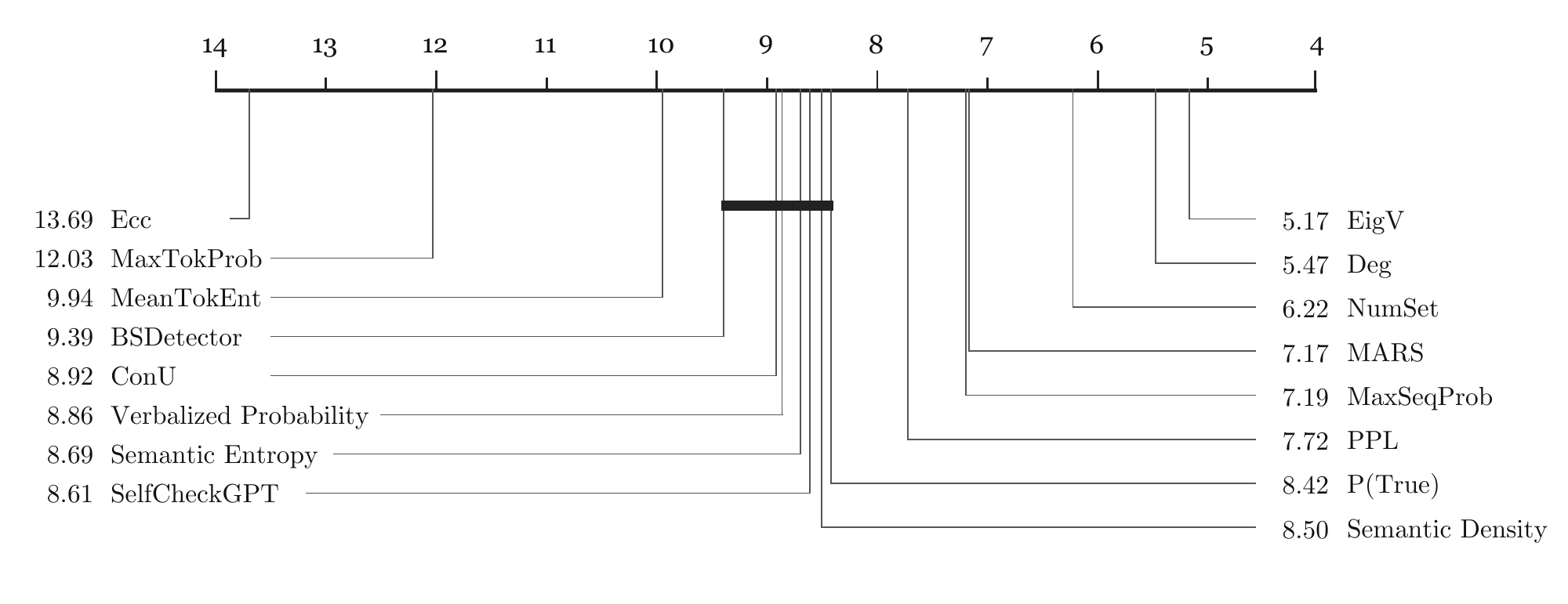}
        \caption{Ranking of ranking evaluation, averaged across AUROC, AUPRC, and AUARC.}
    \end{subfigure}
    \begin{subfigure}[b]{\textwidth}
        \centering
        \includegraphics[width=\textwidth]{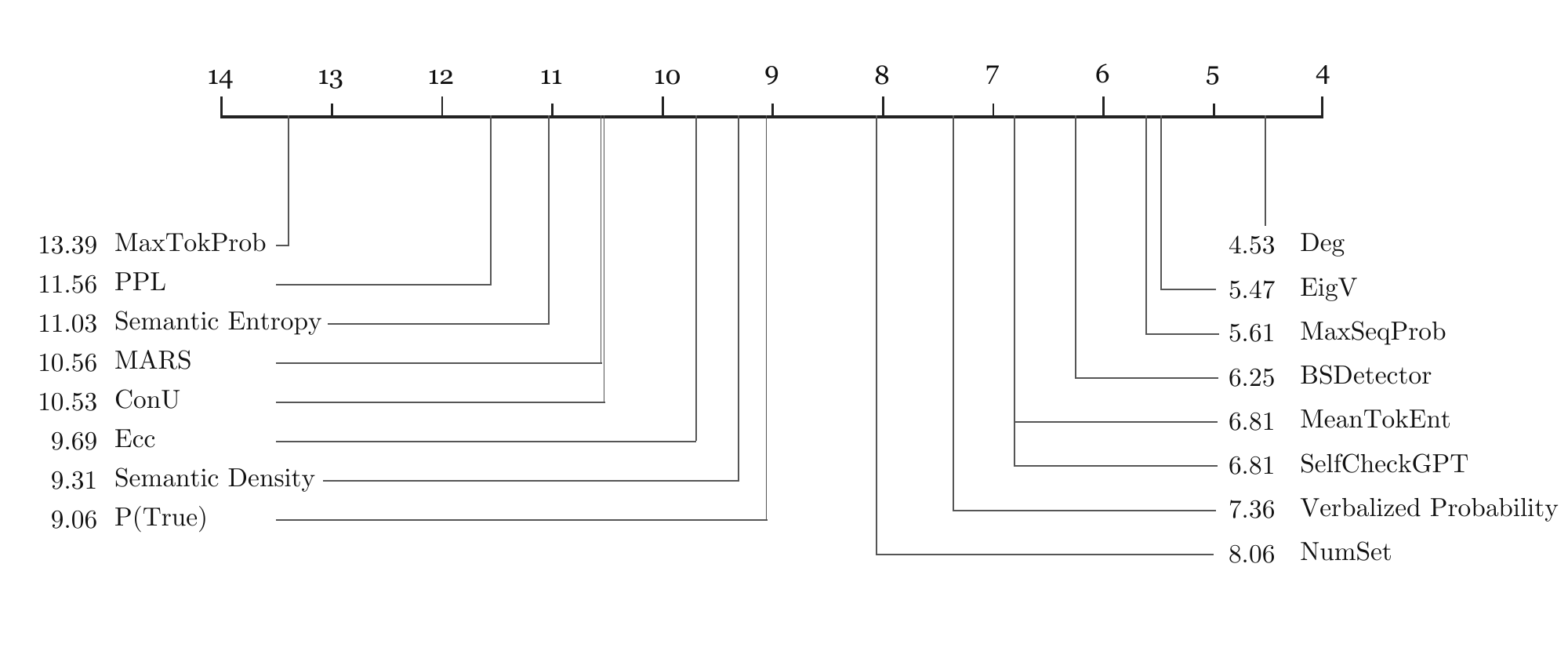}
        \caption{Ranking of calibration evaluation, averaged across ECE, MCE, and BS.}
    \end{subfigure}
    \begin{subfigure}[b]{\textwidth}
        \centering
        \includegraphics[width=\textwidth]{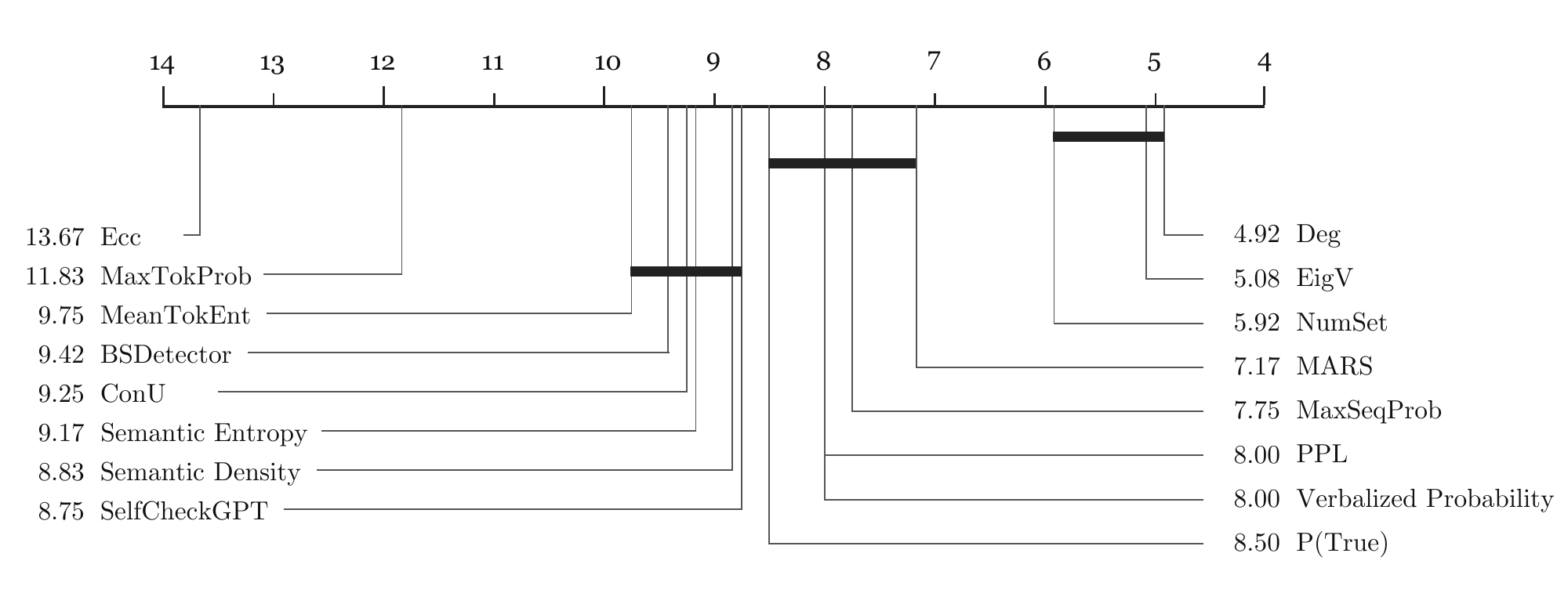}
        \caption{Ranking of AURC that corresponds to coverage evaluation.}
    \end{subfigure}
    \caption{Evaluation ranking of the investigated LLM UQ methods under the answer-only generation setting.}
    \label{fig:answer_only_ranking_combined}
\end{figure*}

\begin{figure*}[t]
    \centering
    \begin{subfigure}[b]{\textwidth}
        \centering
        \includegraphics[width=\textwidth]{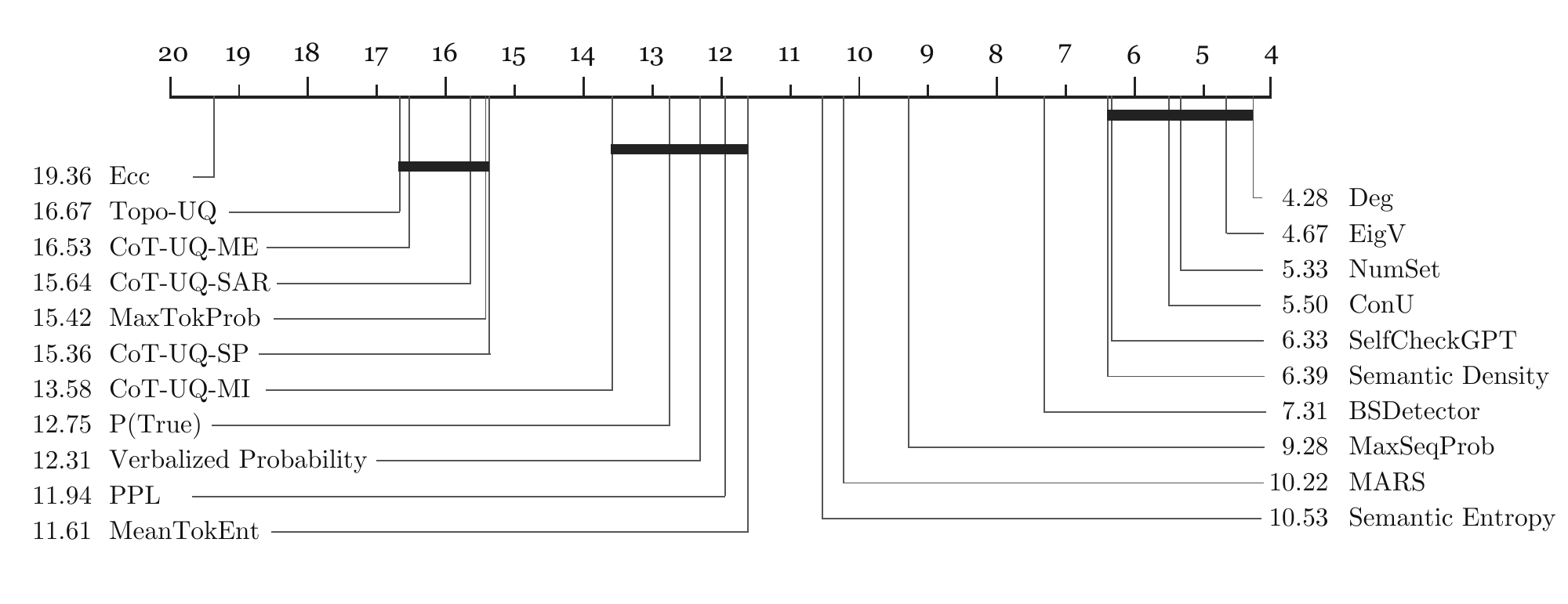}
        \caption{Ranking of ranking evaluation, averaged across AUROC, AUPRC, and AUARC.}
    \end{subfigure}
    \begin{subfigure}[b]{\textwidth}
        \centering
        \includegraphics[width=\textwidth]{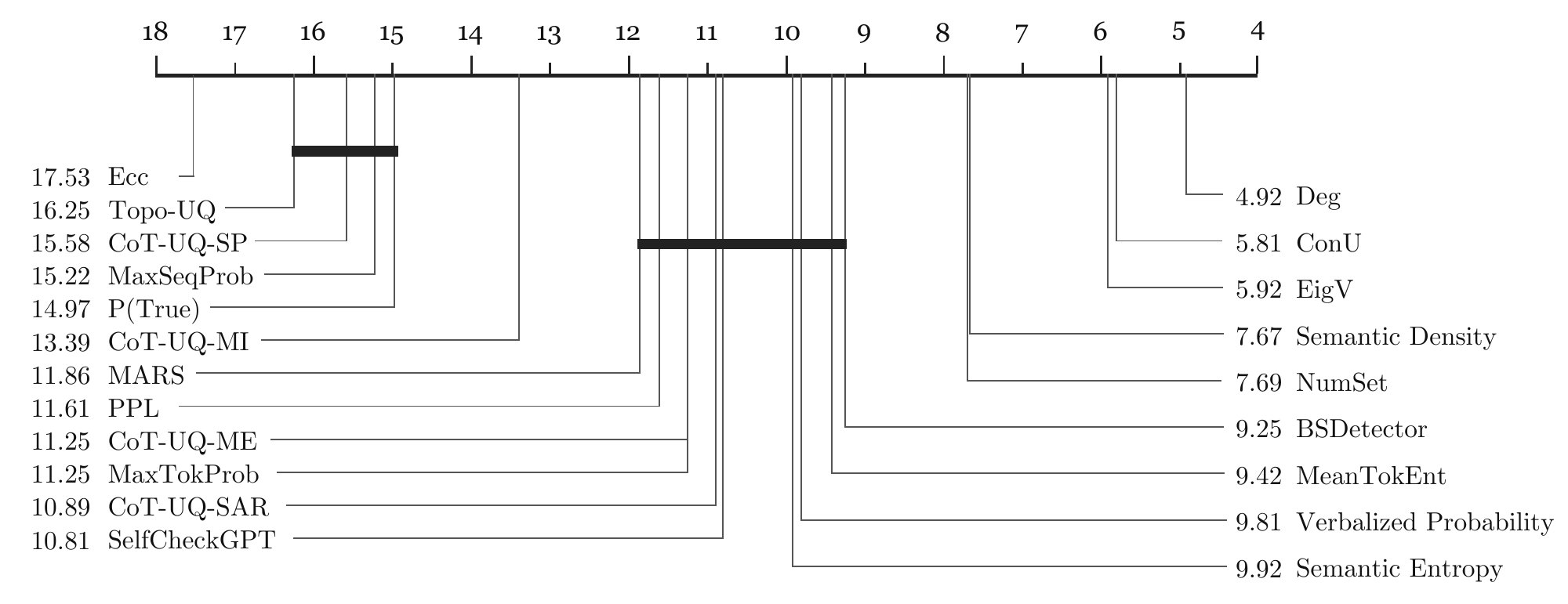}
        \caption{Ranking of calibration evaluation, averaged across ECE, MCE, and BS.}
    \end{subfigure}
    \begin{subfigure}[b]{\textwidth}
        \centering
        \includegraphics[width=\textwidth]{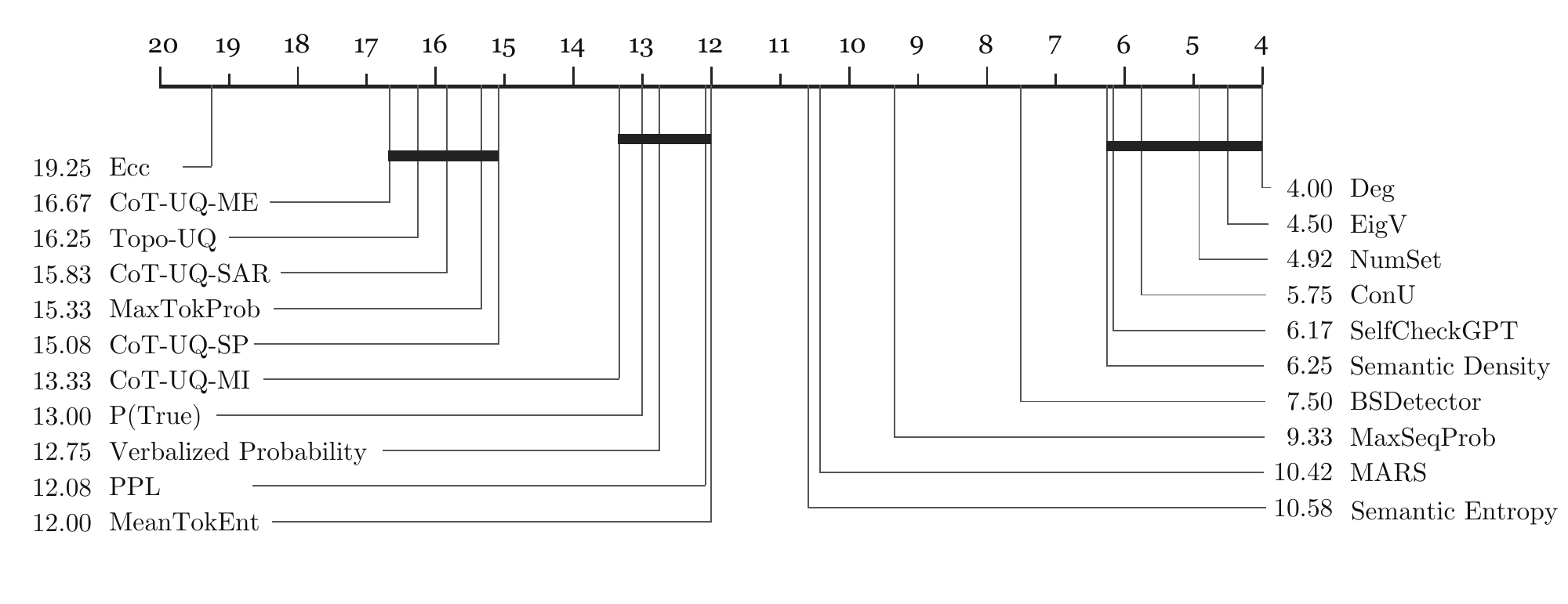}
        \caption{Ranking of AURC that corresponds to coverage evaluation.}
    \end{subfigure}
    \caption{Evaluation ranking of the investigated LLM UQ methods under the reasoning-augmented generation setting.}
    \label{fig:reasoning_ranking_combined}
\end{figure*}

Figure~\ref{fig:answer_only_ranking_combined} and Figure~\ref{fig:reasoning_ranking_combined} display the overall ranking plots for the ranking, calibration, and coverage evaluations in the answer-only and reasoning-augmented settings, respectively. It is observed from Figure~\ref{fig:answer_only_ranking_combined} that EigV, Deg, Numset, and MARS achieve top-tier performance in both ranking and coverage, whereas Deg, EigV, MaxSeqProb, and BSDetector lead in calibration in the answer-only setting. From Figure~\ref{fig:reasoning_ranking_combined}, we observe that Deg, EigV, NumSet, and ConU achieve top-tier performance in both ranking and coverage, whereas Deg, ConU, EigV, and Semantic Density lead in calibration.

\clearpage

\clearpage
\bibliographystyle{apalike}
\bibliography{LLM_UQ}

\end{document}